\documentclass{article}

\usepackage{graphicx}
\usepackage{cite} 
\usepackage{amsmath}

\newcommand{\widefigwidth}{5.0in}
\newcommand{\figwidth}{3.5in}
\newcommand{\mfigwidth}{3.0in}
\newcommand{\smallfigwidth}{2in}

\newcommand{\eq}[1]{Eq.~\ref{eq.#1}} 
\newcommand{\eqbare}[1]{\ref{eq.#1}} 
\newcommand{\fig}[1]{Fig.~\ref{fig.#1}}

\newcommand{\tbl}[1]{Table~\ref{table.#1}}

\newcommand{\sect}[1]{Section~\ref{sect.#1}}
\newcommand{\sectA}[1]{Appendix~\ref{sect.#1}}
\newcommand{\sectlabel}[1]{\label{sect.#1}}
\newcommand{\eqlabel}[1]{\label{eq.#1}}
\newcommand{\figlabel}[1]{\label{fig.#1}}
\newcommand{\tbllabel}[1]{\label{table.#1}}

\newcommand{\pressure}{p}
\newcommand{\power}{P}

\newcommand{\nHill}{n}  
\newcommand{\Phalf}{\pressure_{50}} 
\newcommand{\Oxygen}{\ensuremath{\mbox{O}_2}}
\newcommand{\CarbonDioxide}{\ensuremath{\mbox{CO}_2}}
\newcommand{\CoxygenMax}{\ensuremath{C_{\textnormal{\Oxygen}}^{\rm max}}}
\newcommand{\CoxygenLung}{\ensuremath{C_{\textnormal{\Oxygen}}^{\rm lung}}}
\newcommand{\Coxygen}{\ensuremath{C_{\textnormal{\Oxygen}}}}

\newcommand{\Hoxygen}{\ensuremath{H_{\textnormal{\Oxygen}}}} 
\newcommand{\Sequib}{\ensuremath{S_{\textnormal{equib}}}} 
\newcommand{\Doxygen}{\ensuremath{D_{\textnormal{\Oxygen}}}} 

\newcommand{\hematocrit}{\ensuremath{h}} 
\newcommand{\hematocritFull}{\ensuremath{h_{\textnormal{full}}}} 

\newcommand{\vCell}{\ensuremath{v_{\textnormal{cell}}}} 
\newcommand{\vPlasma}{\ensuremath{v_{\textnormal{plasma}}}} 
\newcommand{\alphaCell}{\ensuremath{\alpha_{\textnormal{cell}}}}
\newcommand{\alphaPlasma}{\ensuremath{\alpha_{\textnormal{plasma}}}}

\newcommand{\dVessel}{\ensuremath{d}} 
\newcommand{\vVessel}{\ensuremath{v}}  
\newcommand{\lVessel}{\ensuremath{\ell}} 
\newcommand{\tVessel}{\ensuremath{t}} 
\newcommand{\NVessel}{\ensuremath{N_{\textnormal{vessel}}}} 

\newcommand{\pTankMax}{\ensuremath{p_{\textnormal{max}}}} 

\newcommand{\reactionEnergy}{\ensuremath{e}} 

\newcommand{\PowerTissue}{\ensuremath{\power_{\textnormal{tissue}}}}
\newcommand{\PowerTissueMax}{\ensuremath{\power^{\textnormal{max}}_{\textnormal{tissue}}}}
\newcommand{\ChalfReactionTissue}{K_{\textnormal{tissue}}}
\newcommand{\radiusTissue}{\ensuremath{r_{\textnormal{tissue}}}} 

\newcommand{\radiusCapillary}{\ensuremath{r_{\textnormal{cap}}}} 

\newcommand{\density}{\ensuremath{\rho}}
\newcommand{\bloodVolume}{\ensuremath{V_{\textnormal{blood}}}}

\newcommand{\Rfrom}{\ensuremath{R_{\textnormal{from 2}}}}
\newcommand{\Rto}{\ensuremath{R_{\textnormal{to 1}}}}
\newcommand{\Rrobot}{\ensuremath{R_{\textnormal{robot}}}}
\newcommand{\Rtissue}{\ensuremath{R_{\textnormal{tissue}}}}

\newcommand{\rRobot}{\ensuremath{r_{\textnormal{robot}}}} 
\newcommand{\nDensityRobot}{\ensuremath{\nu_{\textnormal{robot}}}} 
\newcommand{\tRobot}{\ensuremath{\tau_{\textnormal{robot}}}} 
\newcommand{\rateRobot}{\ensuremath{\gamma_{\textnormal{robot}}}} 
\newcommand{\OxygenAbsorptionRate}{\ensuremath{J_{\mbox{\scriptsize robot}}}} 
\newcommand{\PowerRobot}{\ensuremath{\power_{\textnormal{robot}}}}
\newcommand{\fuelCellEfficiency}{\ensuremath{f_{\textnormal{robot}}}}

\newcommand{\PumpMaxFlux}{\ensuremath{F_{\mbox{\scriptsize pump}}}} 
\newcommand{\PumpOxygenAbsorptionRate}{\ensuremath{J_{\mbox{\scriptsize pump}}}}

\newcommand\Pec{\mbox{Pe}}  

\newcommand{\meter}{\mbox{m}}
\newcommand{\millimeter}{\mbox{mm}}
\newcommand{\micron}{\mbox{$\mu$m}}
\newcommand{\nanometer}{\mbox{nm}}
\newcommand{\liter}{\mbox{L}}
\newcommand{\milliliter}{\mbox{mL}}
\newcommand{\minute}{\mbox{min}}
\newcommand{\second}{\mbox{s}}
\newcommand{\millisecond}{\mbox{ms}}
\newcommand{\kilogram}{\mbox{kg}}
\newcommand{\pascal}{\mbox{Pa}}

\newcommand{\joule}{\mbox{J}}
\newcommand{\kilocalorie}{\mbox{kcal}}
\newcommand{\kilowatt}{\mbox{kW}}
\newcommand{\watt}{\mbox{W}}
\newcommand{\picowatt}{\mbox{pW}}

\newcommand{\molecule}{\mbox{molecule}}

\title{Chemical Power for Swarms of Microscopic Robots in Blood Vessels}
\author{Tad Hogg}\author{Tad Hogg\\
{\small Institute for Molecular Manufacturing}\\{\small Palo Alto, CA}
}

\begin{document}
\maketitle

\begin{abstract}

Microscopic robots in the bloodstream could obtain power from fuel cells using glucose and oxygen. Previous studies of small numbers of such robots operating near each other showed how robots compete with their neighbors for oxygen. However, proposed applications involve billions of such robots operating throughout the body. With such large numbers, the robots can have systemic effects on oxygen concentration. This paper evaluates these effects and their consequences for robot power generation, oxygen available to tissue and heating as such robots move with the blood. When robots consume oxygen as fast as it diffuses to their surfaces, available power decreases significantly as robots move from the lungs, through arteries to capillaries and veins. Tens of billions of robots can obtain hundreds of picowatts throughout the circuit, while a trillion robots significantly deplete oxygen in the veins. Robots can mitigate this depletion by limiting their oxygen consumption, either overall or in specific locations or situations. 

\end{abstract}

\section{Introduction}

Implanted medical devices offer a wide range of benefits. Smaller devices will enable additional high-precision applications~\cite{dong07,nelson10,sahoo07,schulz09}. 
For example, microscopic devices can act on specific cells based on logical combinations of surface properties~\cite{lajoie20}.
The development of autonomous microscopic robots~\cite{freitas99,jager00,morris01,nelson10} could enable more complex selection criteria based on the local environment of each robot. 
Autonomous robots determine when and how to use harvested energy for tasks such as locomotion, communication or drug release~\cite{brooks20}. These decisions can account for sensed data, information received from neighboring robots, and the history of such information saved in the robot's memory. Even relatively simple on-board control can provide a wide range of customized behaviors~\cite{webb20}. 
For example, nearby robots could release drugs simultaneously at a microscopic target location they identify with their sensors~\cite{wiesel-kapah16} to produce a large sudden increase in drug concentration at that location.

Realizing the potential of autonomous microscopic robots faces significant challenges for fabrication and operation.
One of these is power, which could come from extending approaches used for larger implanted devices~\cite{amar15,bazaka13,cook-chennault08}. 
For tasks of limited duration, an onboard fuel source might suffice. 
Otherwise, the robots need to obtain energy from their environment.
One demonstrated method to power microscopic devices is using external sources, such as magnetic fields~\cite{martel07,ceylan21} or ultrasound~\cite{ghanem20,wang07}.
Another approach is chemical power~\cite{freitas99}, which is available throughout the body so robots do not need to be close enough to the skin to receive adequate power from external sources. Chemical power is especially useful for long-term tasks where it may not be convenient or feasible to deliver power from outside the body or provide implanted batteries.
One possibility for chemical power is using oxygen in the bloodstream~\cite{barton04,davis07}, such as by reacting glucose with oxygen~\cite{an11,chaudhuri03,gogova10,rapoport12,zebda13}.
In this case, oxygen is the limiting chemical because its concentration is much lower than that of glucose~\cite{hogg10}. Thus this paper focuses on the availability of oxygen.

Of particular medical importance is the eventual development of autonomous robots a few microns in size. These are the largest robots that can travel throughout the entire circulatory system~\cite{freitas99}.
Micron-size robots are considerably smaller than sub-millimeter micromachines~\cite{brooks20} based on microelectromechanical systems (MEMS). On the other hand, they are much larger than nanoparticles used for drug delivery, which can only incorporate a few logic operations~\cite{douglas12}. 
Between these extremes, engineered biological cells~\cite{lim22} can combine sensing, communication and logic processing in micron-size devices.
Theoretical studies suggest nonbiological micron-size robots considered in this paper could have sufficient computation, sensing and communication to coordinate complex activities with single-cell resolution and a wider range of material properties and capabilities than biological cells~\cite{freitas99,hogg06a,li17,morris01,somasundar21}.

Microscopic robot applications can involve large numbers of robots. For instance, micron-size robots could provide customized medical diagnostics and treatments to individual cells throughout the body~\cite{jager00}. This task requires a correspondingly large number of robots to provide a reasonable treatment time, e.g., a few hours. Thus the scenarios discussed in this paper consist of billions of robots, with a total mass in the range of tens to hundreds of milligrams~\cite{freitas99}.

Interacting robots can create swarms with performance and robustness beyond that of individual robots~\cite{floreano21}, including a variety of coherent structures and behaviors~\cite{vicsek95}. For microscopic robots, external fields can induce these interactions~\cite{mohanty20} to form, for instance, specific shapes~\cite{xie19} and fluid motions~\cite{schuerle19,hernandez05}.
For more flexibility, interactions among autonomous robots can arise by exchanging information with neighbors~\cite{rubenstein14,bonabeau99}.

Previous studies of chemical power for microscopic robots focus on one or a few robots~\cite{freitas99,hogg10}. 
Such studies are sufficient to evaluate available power when the number of robots is too small to significantly affect oxygen concentration on large scales.
However, large numbers of robots could have systemic effects in addition to local effects on nearby cells and robots. In particular, robots powered by oxygen could significantly reduce oxygen concentration in the blood before it reaches capillaries where blood delivers oxygen to tissues.

\begin{figure}[t]
\centering 
\includegraphics[width=\figwidth]{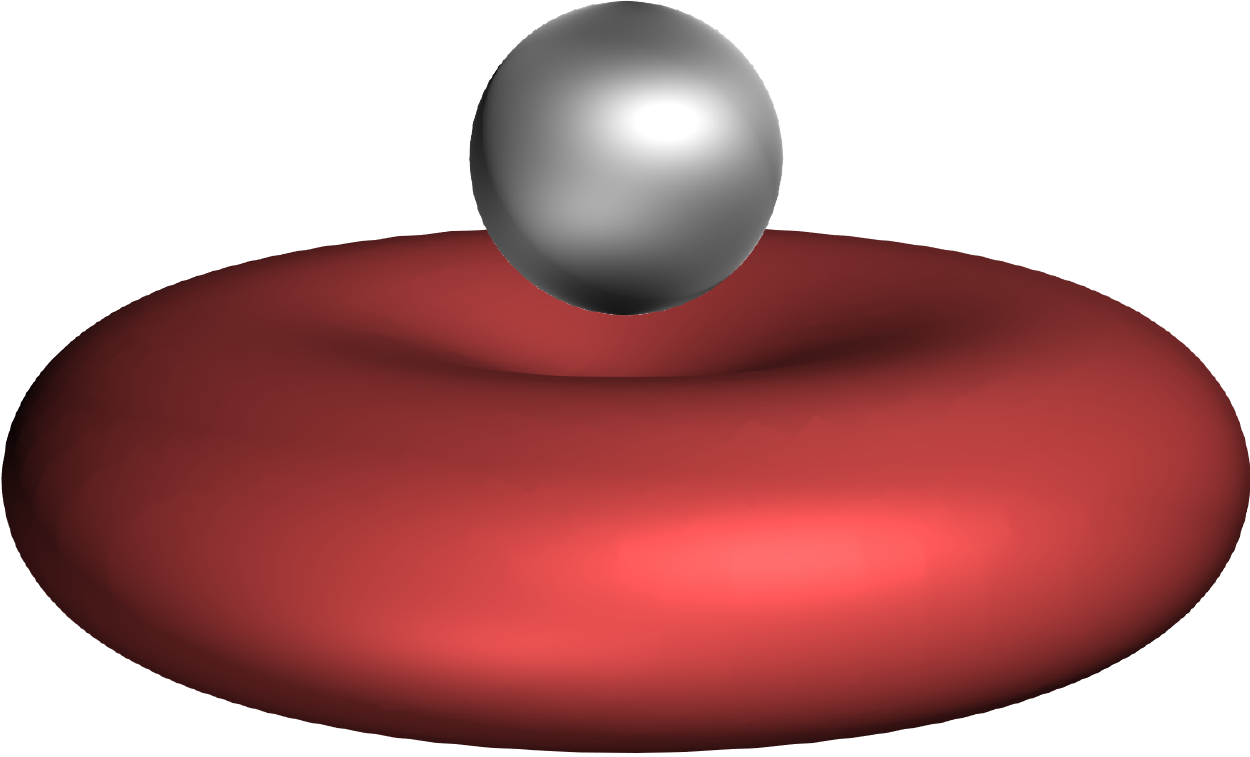}
\caption{Schematic illustration of a 2\,\micron-diameter spherical robot next to a red blood cell.}\figlabel{cell and robot}
\end{figure}

This paper investigates how large numbers of robots small enough to fit through capillaries and illustrated in \fig{cell and robot} affect oxygen concentration as they travel with the blood. Specifically, \sect{model} describes a model of robot transit and oxygen consumption. The model's estimate for robot power and oxygen concentration is described in \sect{power} for a range of robot numbers. In some cases, robots can significantly deplete oxygen in the blood. To avoid this, \sect{mitigation} presents several mitigation strategies robots could use. 
\sect{discussion} describes limitations and extensions to the model that could cover a wider variety of situations than considered here.
Finally, \sect{conclusion} summarizes the model results and its implications for using chemical power for microscopic robots.

\section{Determining the Power Available to Circulating Robots}\sectlabel{model}

The model developed here for robot oxygen consumption consists of two parts: the first evaluates how oxygen concentration changes during a typical circulation loop, and the second determines how much power robots produce from the available oxygen.
Most of the circulation through the body passes through a single capillary network between arteries and veins. We focus on this typical case. Other situations, such as portal flows where the blood flows through two sets of capillaries, require generalizing the model considered here.

\subsection{Robot Power Generating Capacity}

We consider robots using fuel cells that oxidize glucose. Oxygen is the rate-limiting chemical for this reaction since its concentration in the blood is much lower than that of glucose. Thus the maximum power is when robots use all oxygen reaching their surfaces. This is the case considered here, which corresponds to the ``high capacity robot with pump'' design of Ref.~\cite{hogg10}.

When the distance between neighboring robots is large compared to their size, a good approximation of robot oxygen collection is that of an isolated absorbing sphere. Such a robot in blood plasma with oxygen concentration $c$ absorbs oxygen at the rate~\cite{berg93}
\begin{equation}\eqlabel{robot absorption}
\OxygenAbsorptionRate = 4\pi \Doxygen \rRobot c
\end{equation}
where $\Doxygen$ is oxygen's diffusion coefficient in plasma and \rRobot\ is the robot radius.
Absorption can be close to this value even when only a few percent of the robot surface absorbs oxygen~\cite{berg93}.

The reaction energy of glucose oxidation and rate of oxygen absorption determine the power generated by the fuel cell.
Not all of that power is available for robot operations due to losses such as dissipation in wires connecting fuel cells to loads in the robot and internal losses in the fuel cell, e.g., due to membrane transport and incomplete catalysis. 
In addition, robots could use pumps to improve oxygen collection, in which case the power used by those pumps contributes a small amount the power system losses~\cite{freitas99}. 
We characterize these losses by the fuel cell efficiency, \fuelCellEfficiency: the fraction of the full reaction energy available to the robot. 
Thus the power available to the robot from glucose oxidation is
\begin{equation}\eqlabel{robot power}
\PowerRobot = \fuelCellEfficiency \frac{\reactionEnergy}{6}  \OxygenAbsorptionRate
\end{equation}
where \reactionEnergy\ is the energy released by oxidizing a glucose molecule and the factor of $1/6$ arises from the six oxygen molecules needed to oxidize each glucose molecule.

\subsection{Modeling Robot Oxygen Consumption}\sectlabel{oxygen}

A small number of robots in the bloodstream have no systemic effect on oxygen concentration. In that case, the normal concentrations found in the circulation determine the power available to the robots and nearby tissue, whether the robots are widely separated or form aggregates~\cite{hogg10}.
However, large numbers of robots may consume a significant fraction of the oxygen in the blood. 

Red blood cells carry most of the oxygen in blood. These cells release oxygen in partial compensation for that consumed by the robots. This replenishment depends on the concentration of red blood cells, i.e., the hematocrit. In small vessels, cells typically travel a bit faster than plasma, 
leading to a reduced hematocrit in these vessels~\cite{mcHedlishvili87}. 
In addition, tissue consumes oxygen from nearby capillaries.

\begin{figure}[t]
\centering 
\includegraphics[width=\widefigwidth]{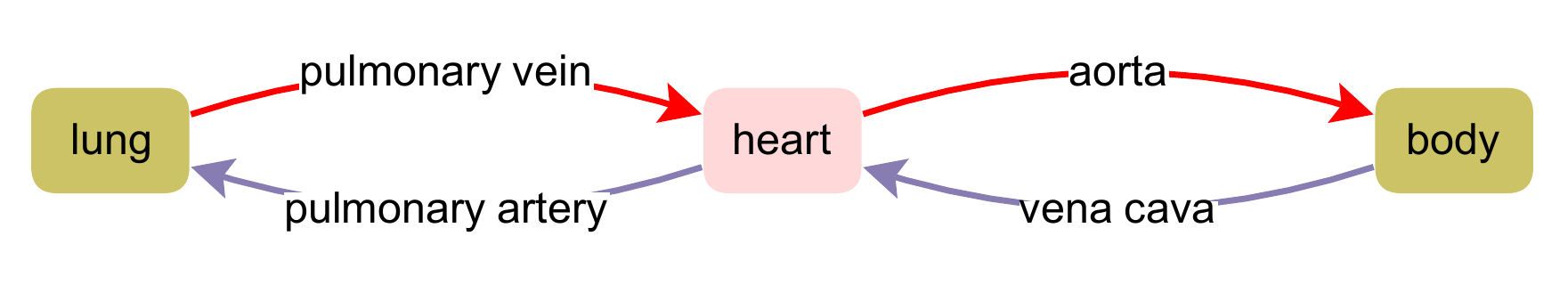}
\caption{Circulation from lungs to the rest of the body and back to the lungs.}\figlabel{circulation}
\end{figure}

Estimating systemic effects of robot oxygen consumption does not require precise vessel geometry of the many circulatory paths through the body. Instead, we aggregate vessels of each size and consider flow through an average aggregated structure for a single loop around the circulation shown in \fig{circulation}. This loop starts as fully oxygenated blood leaves a lung capillary and continues until the blood next reaches a lung capillary. A useful simplification for evaluating the typical power available to robots is to average over the vessel cross section, which gives a one-dimensional model of the blood flow.

A circulation loop takes about one minute. Evaluating behavior on this time scale averages over the short term variations in flow speeds, particularly in arteries, due to heart beats. We also consider a fixed, resting pose for the body and do not treat longer-term variations from changes in activity level or environmental factors.
This temporal averaging leads to a steady-state profile of oxygen concentration in a typical circulation loop.
\sectA{vessel structure} provides the details of this vessel model, which combines spatial aggregation and temporal averaging. 
This aggregated model contrasts with more detailed models of small portions of vessels required to evaluate the local effects of a small number of robots~\cite{hogg10}.

\begin{figure}
\centering 
\includegraphics[width=\mfigwidth]{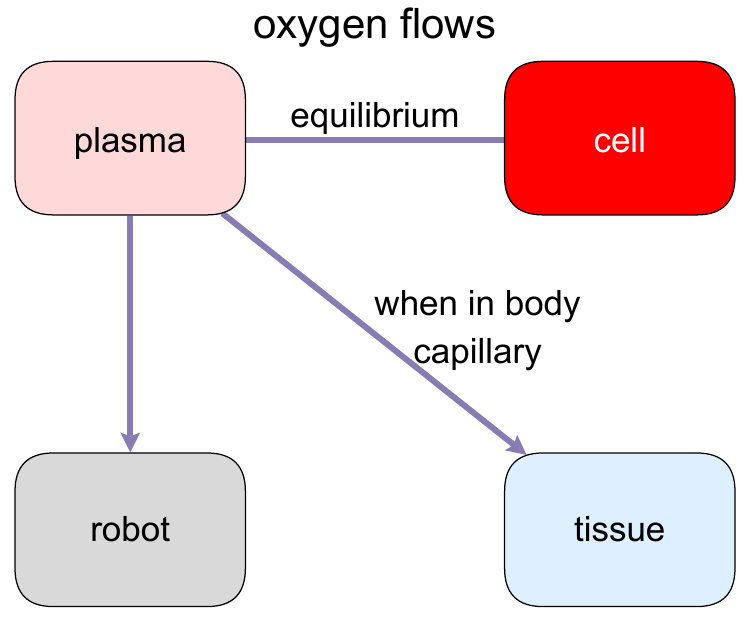}
\caption{Compartment model of oxygen flows after blood leaves the lungs.}\figlabel{oxygen flows}
\end{figure}

Consider a small volume of blood moving through the circulation, containing plasma, cells and robots. 
The oxygen concentration model consists of four compartments. Blood vessels contain three of these compartments: blood plasma, blood cells, and robots. The robots occupy a small fraction of the plasma volume. The fourth model compartment is the tissue around capillaries.   
Oxygen concentration in the moving volume of blood changes due to consumption by robots and tissue, and oxygen release by red blood cells. \fig{oxygen flows} illustrates these processes, with details given in \sectA{oxygen changes}.

\section{Robot Power and Oxygen Concentration}\sectlabel{power}

\begin{figure}
\centering 
\begin{tabular}{cc}
\includegraphics[width=\smallfigwidth]{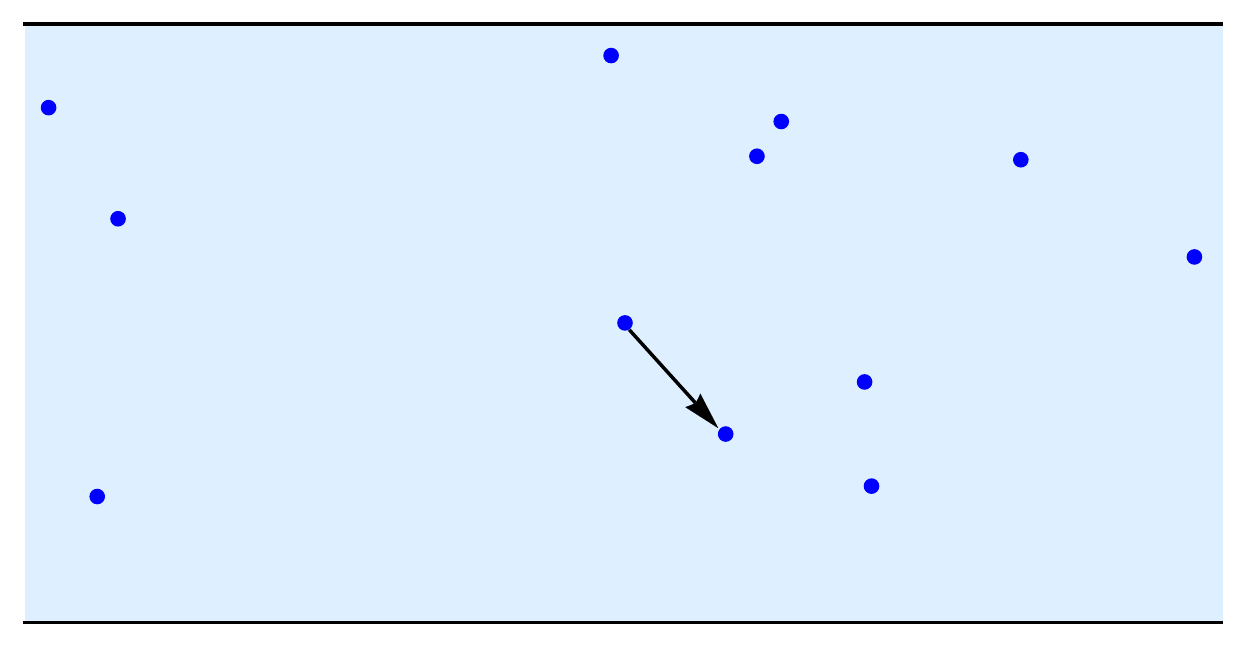}	&	\includegraphics[width=\smallfigwidth]{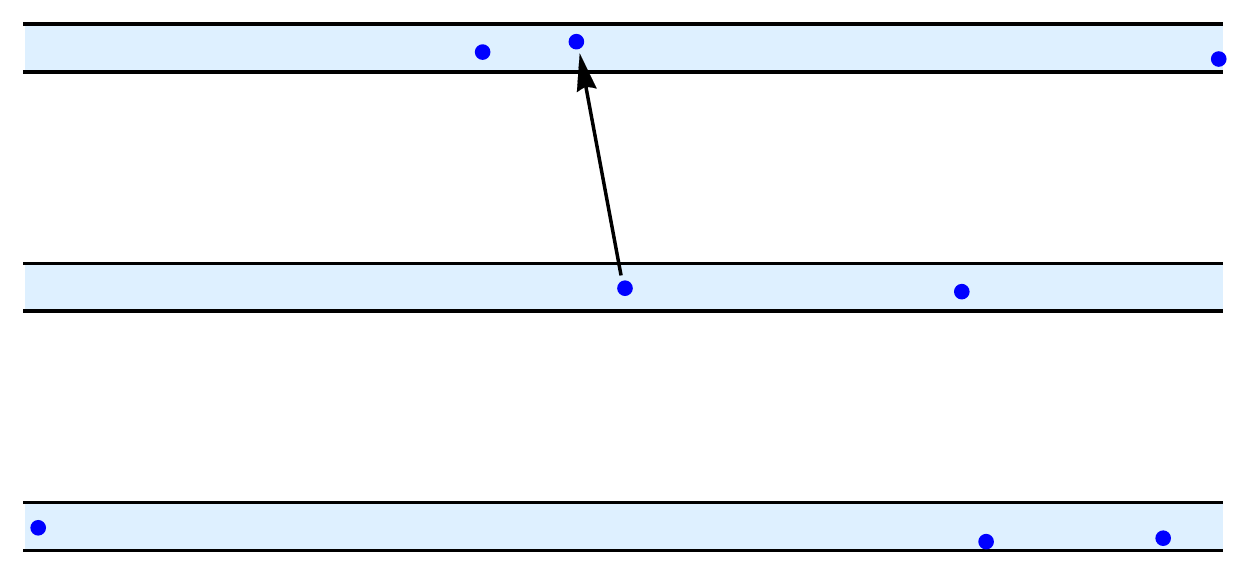} \\
(a) & (b) \\
\end{tabular}
\caption{Schematic illustration of typical spacing between robots, indicated as points, in vessels. The arrows indicates the neighbor closest to the robot in the center of each diagram. (a) In large vessels, the nearest robots are within the same vessel. (b) In small vessels, the nearest robot may be in a neighboring vessel.}\figlabel{spacing}
\end{figure}

\begin{table}
\centering
\begin{tabular}{ccc|ccc}
			&				& robot					& \multicolumn{3}{c}{robot spacing}\\
number		& nanocrit			& number density			& large vessels		& capilliaries		& body  \\ \hline
$10^{10}$		&  $8\times10^{-6}$	& $2\times 10^{12}/\meter^3$	& $80\,\micron$		& N/A			& $170\,\micron$	\\
$10^{11}$		& $8\times10^{-5}$	& $2\times 10^{13}/\meter^3$	& $40\,\micron$		& N/A			& $80\,\micron$		\\
$10^{12}$		& $8\times10^{-4}$	& $2\times 10^{14}/\meter^3$	& $20\,\micron$		& $100\,\micron$	& $40\,\micron$		\\
\end{tabular}
\caption{Properties of various numbers of robots in the circulation. Nanocrit is the fraction of the blood volume occupied by robots, in analogy with hematocrit, which is the fraction occupied by red blood cells. The robot spacing is typical distance between neighboring robots. The value for large vessels is the cube root of the average blood volume per robot. In capillaries, the spacing is from the blood volume in a $8\,\micron$-diameter vessel, up to a maximum capillary length of $1\,\millimeter$. The spacing in the body is the cube root of the average body volume per robot for a nominal $50\,\liter$ body volume. 
}\tbllabel{scenarios}
\end{table}

This section applies the model described in \sect{model} to determine robot power and oxygen concentration for the robot numbers given in \tbl{scenarios}, which are typical values for proposed applications~\cite{freitas99}.  The largest number of robots considered here, $10^{12}$, have a combined mass of several grams and volume of a few milliliters.
The table gives typical spacings between robots in large vessels, within capillaries and between nearby small vessels, as illustrated in \fig{spacing}. These spacings are much larger than the size of the robots so they do not directly compete with neighboring robots for oxygen, in contrast to situations where robots operate in close proximity~\cite{hogg10}. 

Micron-size robots have considerably smaller volume than red blood cells, so even the largest number of robots considered here occupy less than $0.1\%$ of the blood volume, compared to 45\% typically occupied by blood cells.
This small fraction of robots does not significantly affect blood rheology~\cite{freitas99}, allowing the model to use typical flow speeds in the absence of robots.

\subsection{Power when Robots Maximally Consume Oxygen}\sectlabel{typical circulation}

\begin{figure}
\centering 
\includegraphics[width=\figwidth]{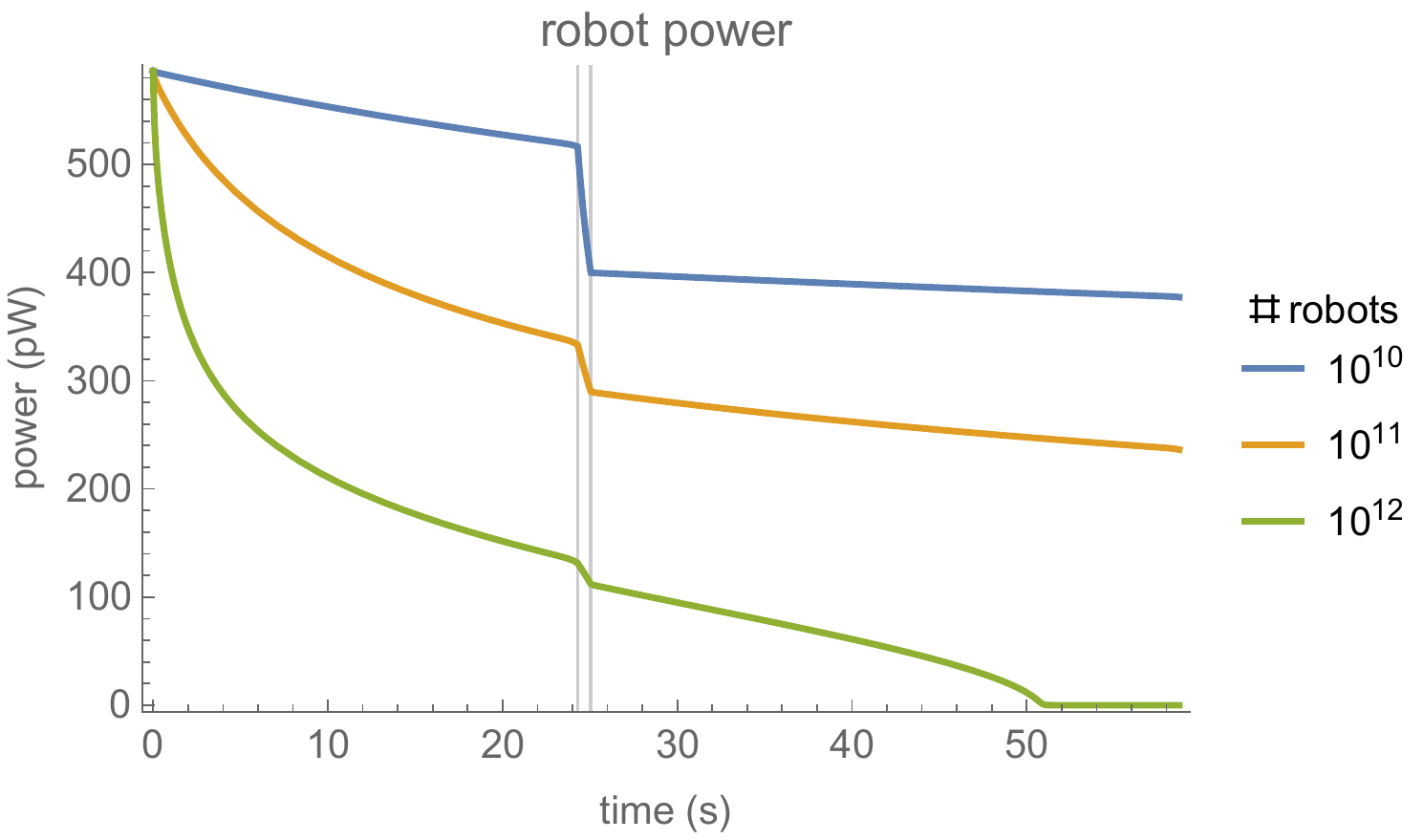} 
\caption{Power used by a robot during a circulation loop: starting when a robot leaves a lung capillary and ending just before it next enters a lung capillary. The light vertical lines indicate when the blood is in a capillary, where tissue consumes oxygen from the blood.}\figlabel{power}
\end{figure}

\fig{power} shows how robot power varies during a circulation loop when robots consume oxygen as fast as it diffuses to their surfaces, so robot power is given by \eq{robot power}.
Robot power decreases as the robot moves from the lung: hundreds of picowatts in the arteries, an abrupt reduction in the capillary where tissue competes with the robots for oxygen, and a continued gradual decrease as the robot travels through veins back to the lung.

The consequences of this variation in power during a circuit depend on how well it matches the power requirements of the robots' tasks. 
For instance, if the robots require some power to maintain their activity, the minimum power available during the circulation, i.e., just before returning to the lung, is the relevant model result. If this minimum is below the required power, either the task will need to use fewer robots or the robots will need sufficient onboard energy storage to support their activities until they return to the lung.

Some applications have flexible timing of robot power demand. Such tasks could include computation to evaluate sensor measurements, maintenance tasks such as checking robot functionality, and communicating information to other robots. 
In such cases, robots could wait until there is abundant power to perform these tasks, e.g., while passing through arteries.

In other cases, the robots might have highest power demand during the short time they pass through capillaries. For instance, robots might need to move to cells near the capillary that are emitting specific chemicals into the blood, thereby requiring the robots to measure, compute and propel themselves while in the capillaries~\cite{hogg06a}. In this case, the relevant value from \fig{power} is the power available while robots pass through capillaries.

\subsection{Consequences of Robot Oxygen Consumption}\sectlabel{consequences}

\begin{figure}
\centering 
\begin{tabular}{cc}
\includegraphics[width=2.25in]{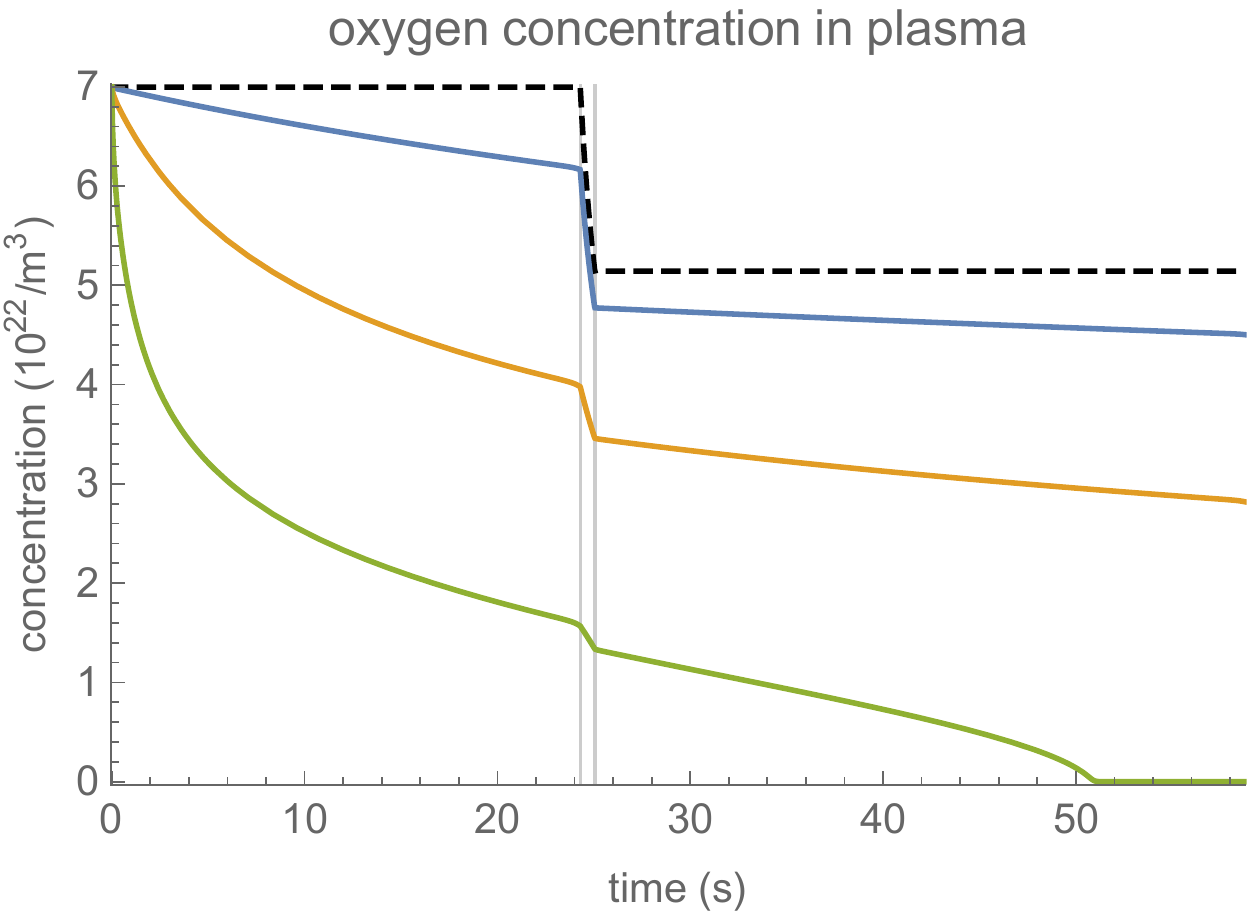}	&	\includegraphics[width=\mfigwidth]{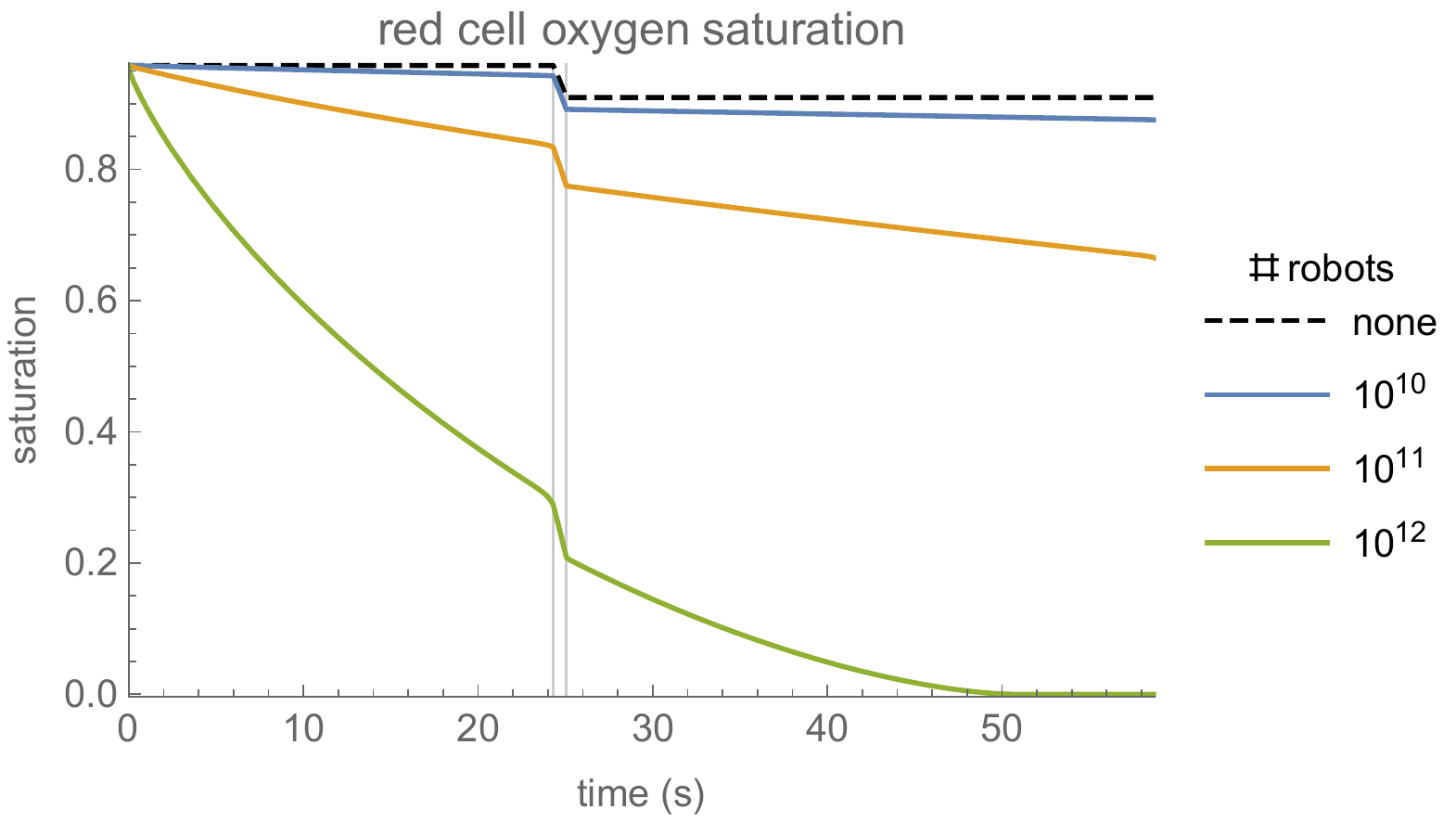} \\
(a) & (b) \\
\end{tabular}
\caption{(a) Oxygen concentration in plasma expressed in molecules per unit volume.  For its relation to units more commonly used in macroscopic settings, see \sectA{parameters}.
 (b) Oxygen saturation of red cells.
The light vertical lines indicate when the blood is in a capillary.}\figlabel{oxygen}
\end{figure}

For further insight into the consequences of robots using oxygen for power, \fig{oxygen} shows oxygen concentration in the plasma and red cell saturation. The figure compares these values with those without robots, in which case the only oxygen consumption occurs in tissue around the capillaries. 

Except for the largest number of robots considered here, the diffusion limit on the rate oxygen reaches the robot surface (i.e., \eq{robot absorption}) prevents the robots from fully depleting oxygen in the blood. That is, as is the case without robots, the blood returns to the lungs with much of the oxygen it originally took from them.
By contrast, a trillion robots consuming oxygen as fast as possible completely depletes oxygen by the end of the circuit.

\subsubsection{tissue}

\begin{figure}
\centering 
\includegraphics[width=\figwidth]{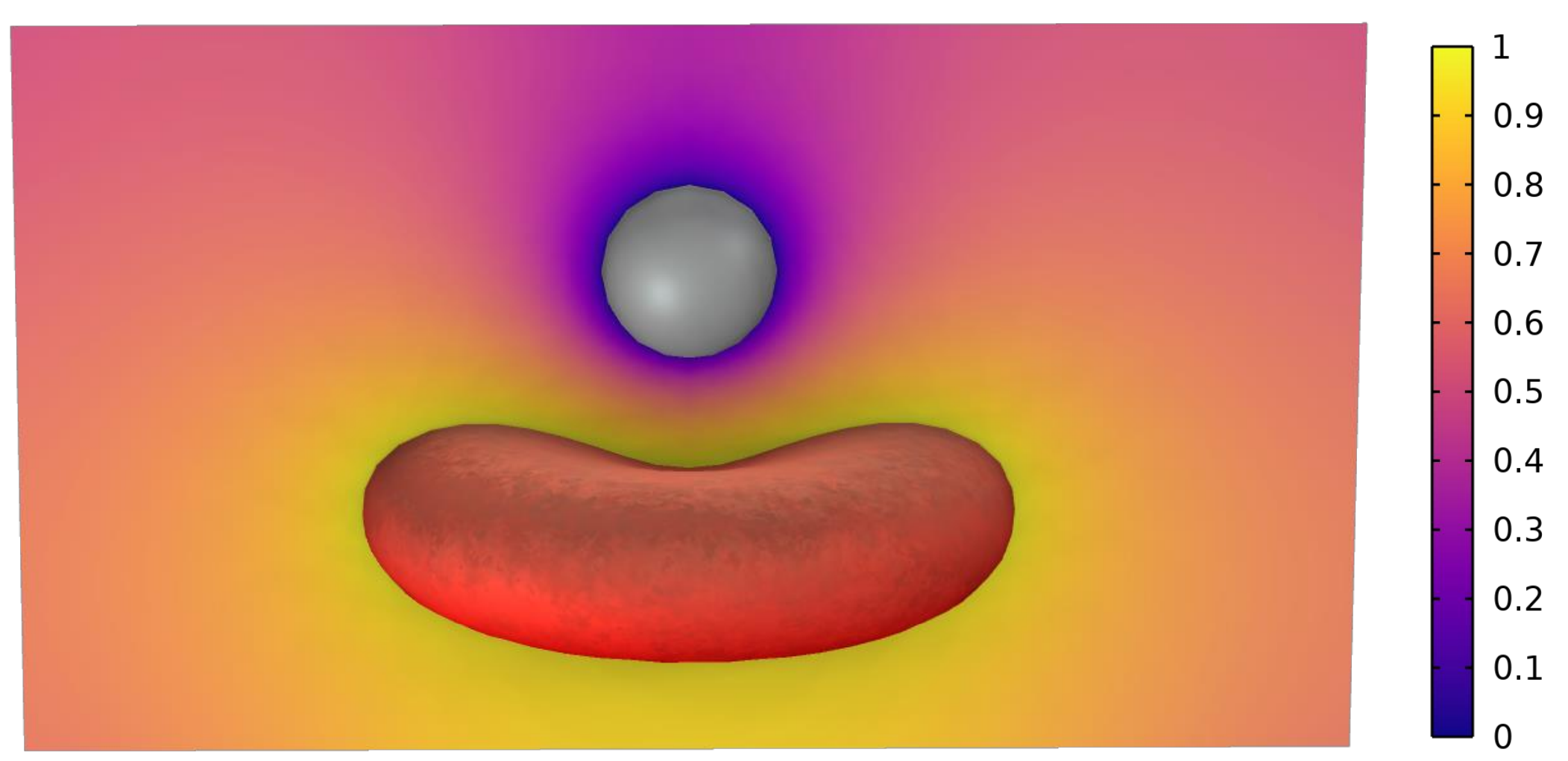} 
\caption{Relative oxygen concentration around an absorbing robot next to a blood cell, ranging from 1 at the cell surface to 0 at the robot.}\figlabel{cell robot concentration}
\end{figure}

\begin{figure}
\centering 
\includegraphics[width=\figwidth]{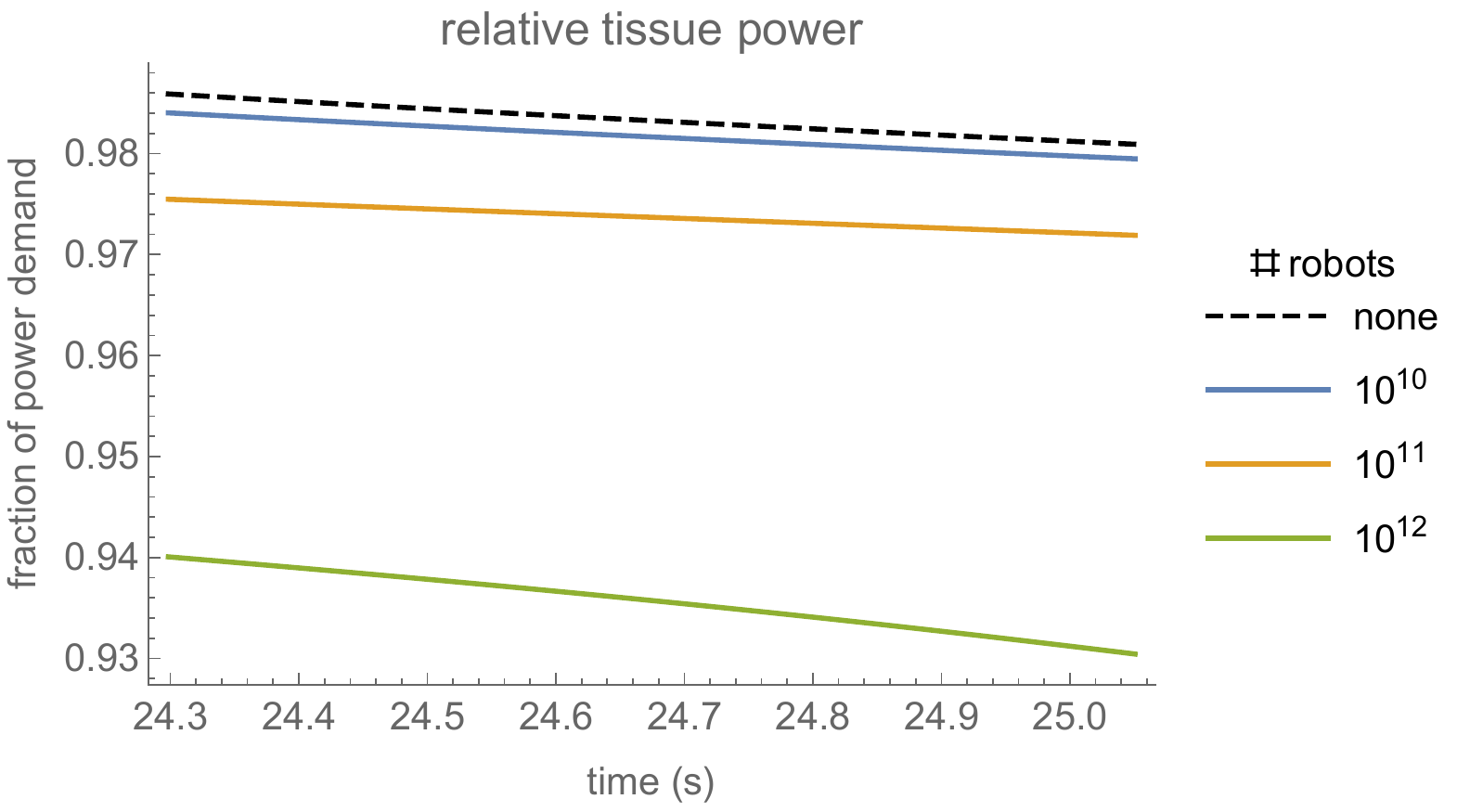} 
\caption{Tissue power relative to its maximum value with unlimited oxygen.}\figlabel{tissue}
\end{figure}

An important consequence of robots consuming oxygen is their effect on oxygen available to tissue. Robots consume oxygen in arteries bringing oxygen to capillaries and compete with tissue for oxygen in the capillaries. Much of the oxygen robots use comes from red cells because absorbing robots produce large concentration gradients at nearby red cells, as illustrated in \fig{cell robot concentration}.
This additional oxygen from red cells limits the decrease in concentration in plasma. Thus robot consumption leads to only modest reduction of oxygen in the capillaries, as shown in \fig{oxygen}a.

The extent to which this reduced concentration affects tissue depends on the minimum oxygen tissue requires to support its metabolism. Tissue demand leads to the abrupt reduction in concentration in capillaries seen in \fig{oxygen}a. 
Even with the largest number of robots considered here, the concentration in capillaries is sufficient to support resting tissue demand.
Specifically, cellular functions continue to operate normally until oxygen partial pressure is below $5\,\mbox{mmHg}$~\cite{ernsting63}, corresponding to concentration $0.4\times 10^{22}\, \molecule/\meter^3$. This is lower than the capillary concentrations seen in \fig{oxygen}a.

The conclusion of sufficient oxygen for tissue also follows from the tissue power generation described by \eq{tissue reaction rate}. With the tissue parameters in \tbl{parameters}, tissue power is nearly independent of oxygen concentration at the concentrations seen in \fig{oxygen}a. 
Quantitatively, \fig{tissue} shows the reduction in oxygen in capillaries due to robots has only a minor effect on tissue power in the case considered here, i.e., resting metabolic demand. 
Thus blood typically transports much more oxygen than required by tissue and so provides a buffer for any localized increase in tissue metabolic demands.

\subsubsection{vein walls}

\fig{oxygen} indicates the lowest oxygen concentrations occur toward the end of the circulation loop, as robots return to the heart and then to the lungs. Normally, blood returning to the lungs contains a significant portion of the oxygen originally collected in the lungs. Robots can make use of this oxygen without concern of reducing oxygen available to tissue since this blood has already passed through capillaries to deliver oxygen to tissue.

A caveat to this conclusion is that cells comprising the walls of blood vessels consume a portion of the oxygen carried in those vessels.
Specifically, small arteries and veins, and the inner portion of walls of larger vessels, receive nutrients that diffuse into the wall from the blood carried in the vessel~\cite{ritman07}. These cells form a small fraction of the body tissue so their oxygen use is not a significant contribution to total tissue oxygen consumption. Moreover the small diffusion distance of oxygen and the relatively large volume of the vessels compared to capillaries means consumption by the vessel walls does not significantly alter the concentration in the vessels during the time blood flows through them. Thus there is no need to include this oxygen consumption in the model.

However, the lowered oxygen concentration in veins due to the robots could be a safety issue.
For example, normal leg veins have oxygen concentrations of about $30\text{--}40\,\mbox{mmHg}$ and red cell saturation $50\text{--}70\%$~\cite{lim11}.  
As discussed in \sectA{parameters}, $30\,\mbox{mmHg}$ is $2.5\times 10^{22}/\meter^3$. Thus these ranges are a bit below the values toward the end of the circulation loop with $10^{11}$ robots.
This comparison suggests lowered concentration in veins will not be an issue when using that many robots, but could be important for larger numbers of robots.
Thus the oxygen requirements of vein walls may place more stringent constraints on robot oxygen consumption than the requirements of most tissue, which receives nutrients from capillaries. 
Quantifying this constraint requires determining the minimum concentration these cells can tolerate for various amounts of time. An initial assessment is to assume these cells have requirements similar to those of resting tissue discussed above. In that case, \fig{oxygen}a shows that $10^{12}$ robots could be detrimental to vessel walls over the last $15\,\second$ or so of the circulation.

\subsubsection{heating}\sectlabel{heating}

\begin{table}
\centering
\begin{tabular}{cccc}
number		& average power		& minimum power		& total dissipation \\ \hline
$10^{10}$		& $460\,\picowatt$		&  $380\,\picowatt$ 		&  $9\,\watt$ \\
$10^{11}$		& $325\,\picowatt$		& $240\,\picowatt$		& $65\,\watt$\\
$10^{12}$		& $120\,\picowatt$		& none				& $240\,\watt$\\
\end{tabular}
\caption{Power for robots during circulation: average over the circulation, minimum (just before returning to the lungs), and total dissipation from the oxygen consumed by all the robots.
}\tbllabel{power}
\end{table}

Robot power production adds heat to the body. This heating arises from the full reaction energy in the fuel cells, not just the fraction available for the robot's use. Thus the total dissipation from all the robots equals the average available power over a circulation, multiplied by the number of robots and divided by the fuel cell efficiency.
\tbl{power} gives these values for the numbers of robots in \tbl{scenarios}. 
For comparison, typical basal metabolism is $100\,\watt$ or $2000\,\kilocalorie/\mbox{day}$. 

The table shows $10^{10}$ robots, each consuming as much oxygen as it can, add less than $10\%$ to basal metabolism. On the other hand, $10^{12}$ robots add more than twice the basal amount, thereby adding significant heat to the body. This is well below the heat dissipation during exercise~\cite{freitas99}, but nevertheless may be a limiting factor during prolonged robot operation.

\subsubsection{oxygen concentration measurement}\sectlabel{conventional sensors}

The oxygen concentration profile produced by robots alters assumptions underlying some clinical measurements, particularly the relation between direct measurements and inferred properties based on causal relationships in the body~\cite{bartlett95}. 
For instance, normally oxygen is only removed from blood in capillaries so the arterial concentration is the same throughout the body. This is the basis for pulse oximetry, a common, noninvasive measurement of respiratory heath and oxygen supply to organs~\cite{nitzan14}.

Robots consume oxygen in arteries, leading to significant concentration reduction in the vessels between the lung and capillaries, as shown in \fig{oxygen}. Thus in the presence of robots, pulse oximetry will require modified calibration to account for variation in arterial oxygen depending on body location and the time since the blood left the lung. 
These systemic changes could also affect the interpretation of other large-scale functional oxygen measurement techniques, such as
photoacoustic imaging~\cite{cao17a}.
The model described here indicates how oxygen varies with location and could aid in calibrating and interpreting these measurements.

As compensation for their alteration of conventional clinical measurements, robots will be able to measure oxygen concentration throughout the body at micron length scales, thereby far surpassing the accuracy and quantity of external measurements. Interpreting such measurements will need to account for the systemic changes in concentration produced by the robots. In spite of this robot capability, recalibrated conventional sensors will remain useful checks on robot performance and allow comparison with established clinical practice. Moreover, receiving measurements from robots may be delayed due to communication limits, e.g., if robots must wait until they are near the skin to send information out of the body. Delays or reduced information could also occur if continual communication takes too much power away from the robots' main tasks. In such cases, the improved sensing capability of robots will not be as readily available as measurements from conventional sensors.

\subsection{Robots Operating Only in Capillaries}\sectlabel{capillaries}

\begin{figure}
\centering 
\begin{tabular}{cc}
\includegraphics[width=2.5in]{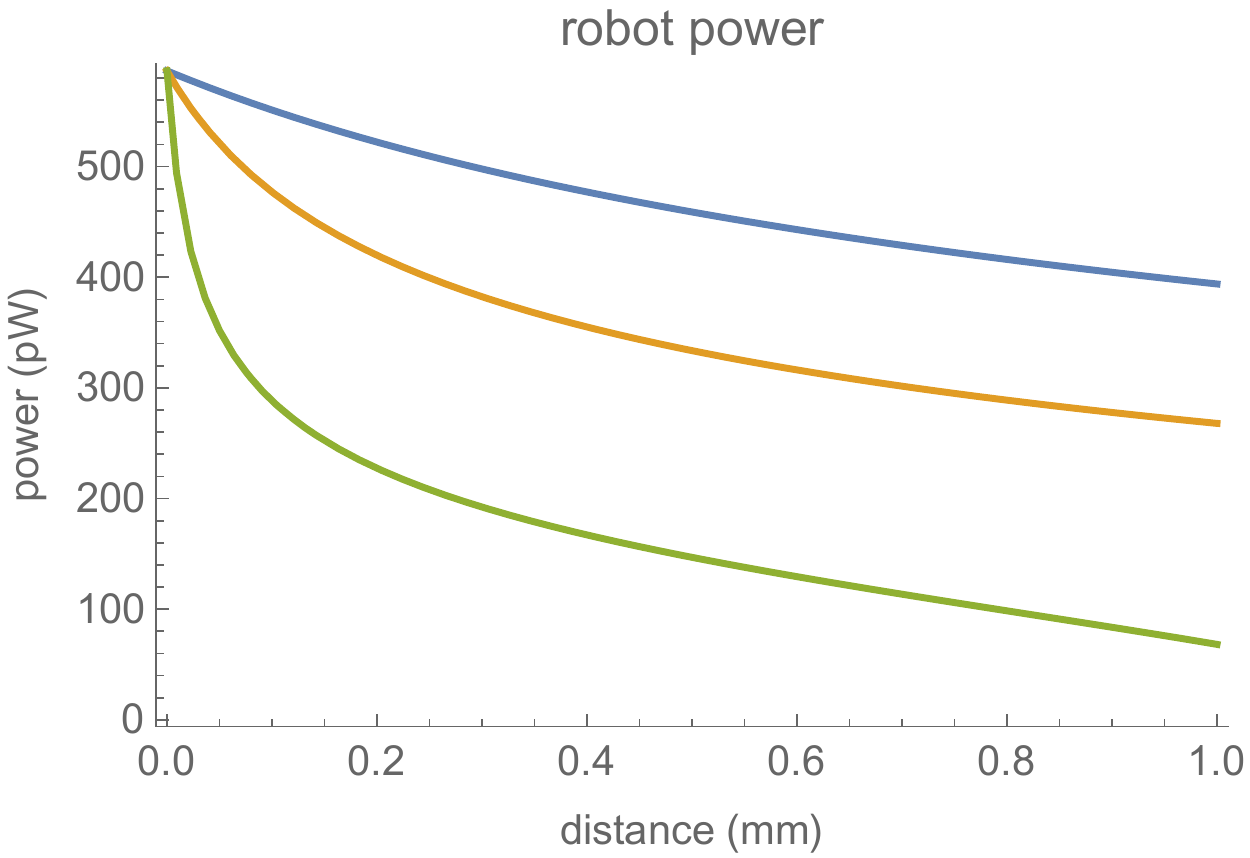}	&	\includegraphics[width=\mfigwidth]{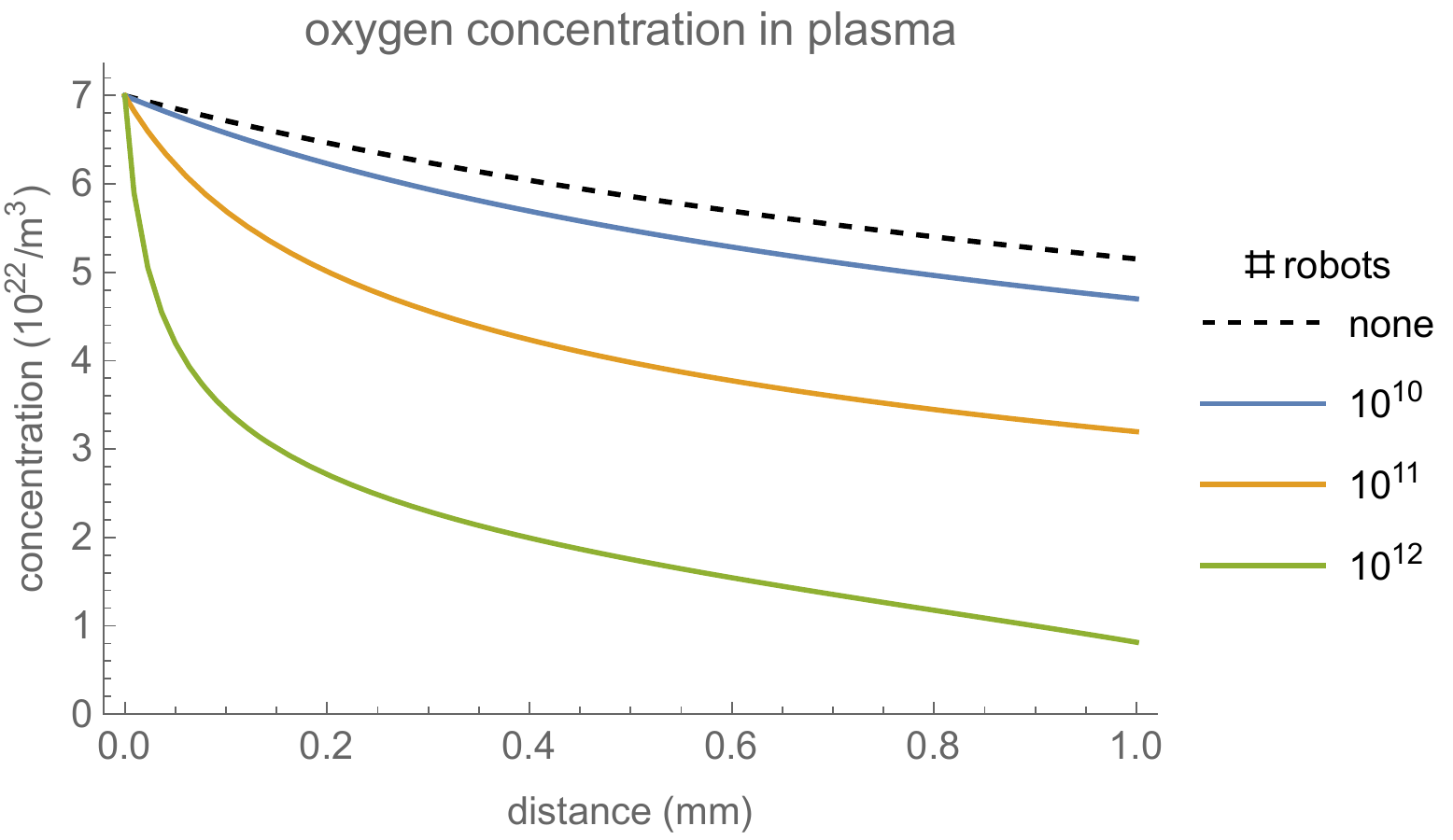} \\
(a) & (b) \\
\end{tabular}
\caption{Behavior with all robots in capillaries as a function of distance along a capillary, with arterial flow arriving from the left. (a) Robot power. (b) Oxygen concentration in plasma.}\figlabel{capillary}
\end{figure}

The focus of this paper is on robots generating power as they travel with the blood.
In some applications, robots perform most of their tasks in capillaries, e.g., to monitor or act on individual cells accessed from capillaries. In this case, robots could initially travel through the circulation until they reach capillaries, at which point they attach to the capillary walls to perform their primary function. Robots in lung capillaries would have access to the oxygen not taken by red cells, as discussed in \sect{oxygen storage}. However, robots in capillaries of the systemic circulation would only consume oxygen from blood passing through capillaries, rather than during the entire circulation. 

Applying the model described in \sect{model} to robots positioned in systemic capillaries requires two modifications. First, there is no robot oxygen consumption in arteries or veins. Second, since these capillaries contain about $5\%$ of the total blood volume~\cite{freitas99},
the number density of robots in capillaries is about 20 times larger than when they are distributed throughout the blood volume.
With these modifications, \fig{capillary} illustrates how robot power and oxygen concentration vary in a capillary with typical length of $1\,\millimeter$.  Since there is no oxygen consumption in arteries, the concentration in blood entering the capillary does not depend on the number of robots. 
Oxygen concentration and robot power decrease through the capillary, so robots near the venous end of the capillary have less power. If necessary, robots near the arterial end of the capillary could reduce their power to leave more available for downstream robots, as a small-scale version of the mitigation strategy discussed in \sect{limiting power}. 

Comparing with \fig{oxygen}a shows that the oxygen concentration in blood after passing through a capillary is similar to that when robots are distributed throughout the blood and consume all oxygen reaching their surface. That is, in both cases, the robots extract about the same amount of oxygen from circulating blood, up to the time it passes through a capillary. Thus the average power per robot and their total dissipation are similar to the values in \tbl{power}.
However, this power is entirely generated in capillaries, giving a much larger power density in the blood as it passes through capillaries.
For example, $10^{10}$ robots heat capillaries at an average rate of about $30\,\kilowatt/\meter^3$. 
This is several times the resting tissue power demand used in this model (see \tbl{parameters}), and comparable to the power density for cells with high metabolic demands, such as heart or kidney cells~\cite{freitas99}.
Nevertheless, heat transport by blood and through tissue is rapid enough at these small scales that this increased power density does not result in much local temperature increase, even when robots are close enough to come into contact~\cite{hogg10}. This indicates that even when all robots operate in capillaries, the main heating issue is for the body as a whole, as discussed in \sect{heating}, rather than local heating in capillaries.

\fig{capillary} shows the average behavior in capillaries. However, the small blood volume in individual capillaries leads to considerable variation due to differences in capillary type~\cite{freitas99}, flows within a network of connected capillaries and the precise locations of robots in a capillary.
For instance, at the lower range of the numbers of robots considered here, each capillary will have only a few robots, on average, so that the actual number of robots will vary considerably among capillaries, including many capillaries with no robots.

\begin{figure}
\centering 
\includegraphics[width=\figwidth]{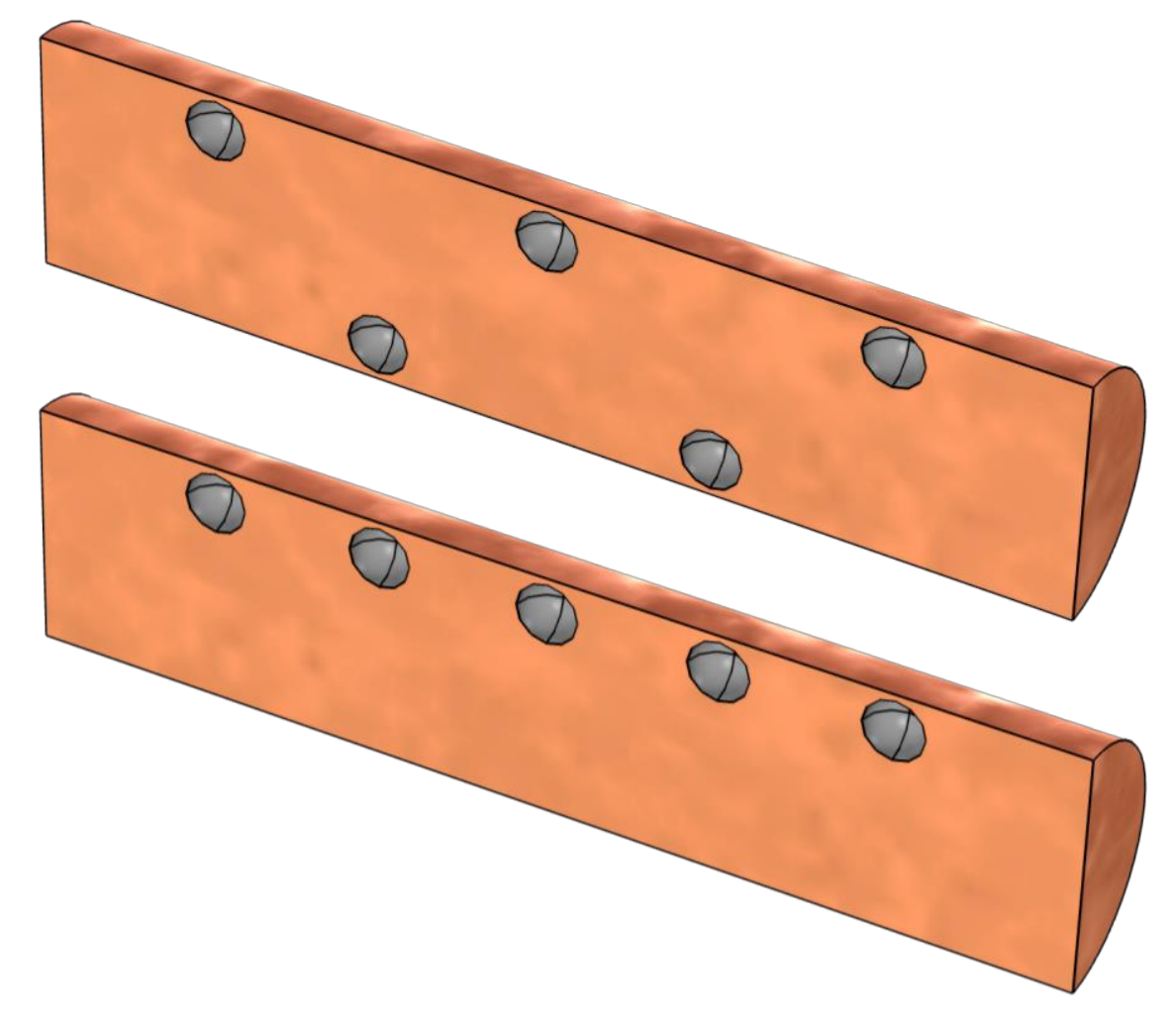}
\caption{Five robots anchored to the inner wall of an $8\,\micron$-diameter vessel with $5.5\,\micron$ spacing along the vessel. The vessel segment shown here is $34\,\micron$ long and is shown with a cross section through the middle of the vessels and robots. Successive robots positioned on opposite sides (top) and the same side (bottom) of the vessel.}\figlabel{robots in small vessel}
\end{figure}

\begin{figure}
\centering 
\includegraphics[width=\textwidth]{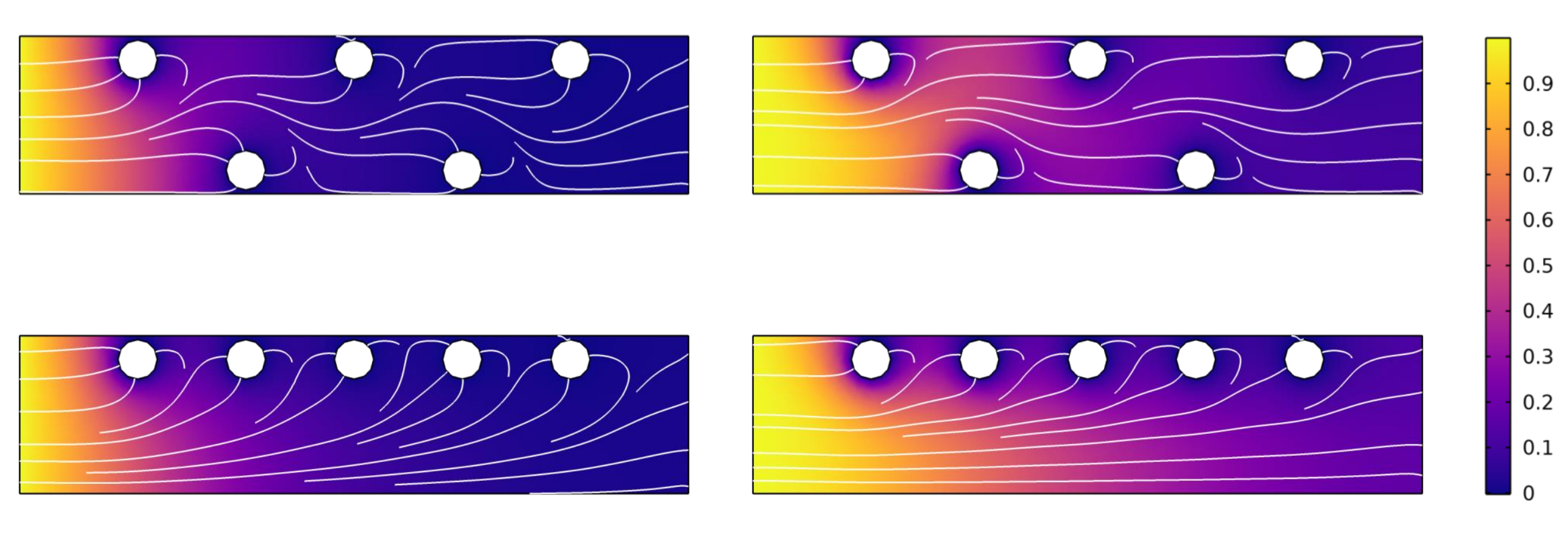}	
\caption{Relative oxygen concentration on a vertical slice through the center of the vessel and the robots, ranging from one at the inlet at the left to zero at the surface of each robot due to its absorption of all oxygen reaching it. The white curves are streamlines of the oxygen flux. The rows correspond to the robot positions shown in \fig{robots in small vessel}. The fluid flows from left to right in each diagram, with average speed $0.2\,\millimeter/\second$ and $1\,\millimeter/\second$ in the left and right columns, respectively.
The oxygen diffusion coefficient is in \tbl{parameters}.}\figlabel{absorbing robots}
\end{figure}

The circulation model considered in this paper does not evaluate variation in oxygen concentration at micron length scales. For example, it does not consider oxygen diffusion across the width of a vessel. Moreover, as discussed in \sectA{oxygen changes}, the model assumes robots are sufficiently far apart that they do not compete with neighboring robots for oxygen. From \tbl{scenarios}, this is reasonable even when all the robots are in capillaries, up to about $10^{11}$ robots or so. However, with $10^{12}$ robots in capillaries, the distance between neighboring robots is only somewhat larger than their diameter.
Thus, that many robots will have some competition for oxygen, reducing the oxygen they collect and the power they generate compared with the model's predictions. 

To illustrate the effect of competition among neighboring robots and diffusion across capillaries, \fig{robots in small vessel} shows two examples of robots in a small segment of a capillary with the average spacing corresponding to $10^{12}$ robots in the capillaries. These examples differ in the positioning of successive robots: alternating between opposites sides of the vessel or all on the same side.
\fig{absorbing robots} shows the resulting oxygen concentration for four scenarios: the two robot positions shown in \fig{robots in small vessel} and with two fluid flow speeds that are a typical range for capillaries~\cite{freitas99}. 
This figure shows large variations in concentration along the vessel, particularly near each of the robots. 
This contrasts with the monotonically decreasing concentration from the averaged circulation model shown in \fig{capillary}b.

The concentration variation across the vessel is particularly large at the higher flow speed, where oxygen does not have enough time to diffuse throughout the vessel cross section while the fluid passes the robots. Robots positioned on the same side of the vessel increase the difference in oxygen concentration between the two sides of the vessel compared to robots on alternate sides.
The differences in oxygen concentration around the robots leads to variation in the rate oxygen molecules reach the robot surface depending on robot position and fluid speed, as shown by the oxygen flux streamlines in \fig{absorbing robots}.

While the large-scale model shown in \fig{capillary} describes behavior averaged over capillaries, the oxygen concentration and robot power will vary considerably from this average in individual capillaries. This is due to variations in the number of robots in the small volume of blood in a capillary and the positions of the robots on the capillary walls. 
Additional variations could arise from features of the flow not included in this small-scale model of oxygen transport in capillaries. For instance, the model neglects oxygen consumption by nearby tissue cells and the possibility that some oxygen reaching nearby tissue diffuses back to the robots in the vessel. Moreover, the model assumes the oxygen saturation of passing blood cells remains in equilibrium with the concentration in the fluid. Accounting for tissue consumption, diffusion outside the vessel, and the kinetics of oxygen release from blood cells gives similar large concentration gradients near absorbing robots on a vessel wall~\cite{hogg10}. Additional changes to the oxygen concentration could arise from blood cells as they distort to move past robots anchored to the vessel wall.

An additional consideration for robots remaining in capillaries rather than moving with the blood is the increase in vascular resistance of the flow, particularly in the narrowest capillaries since vascular resistance of these vessels is inversely proportional to the fourth power of their diameter according to Poiseuille's law~\cite{happel83}.
For example, comparing the pressure drop for a vessel without robots to that of vessels with robots indicates the robots increase vascular resistance by about $20\%$ in both configurations shown in \fig{robots in small vessel}.
Robots on vessel walls may further increase the resistance by changing how blood cells distort as they move past the robots. As an extreme, robots could block passage of the cells.

If this increased resistance reduces the blood flow, robots and the tissue around the capillaries will not receive new oxygenated blood as rapidly as assumed by the circulation model. Alternatively, the body may compensate for the increased resistance by a corresponding increase in the blood pressure.
Avoiding injury from robot blockage or increased blood pressure will limit the number of robots that can be stationed in capillaries, and how long they can remain there. This could be an important limitation on applications that require longitudinal cell-specific data on many cells over an extended period of time.

An alternative way to collect information from the same cell over a period of time that avoids blocking capillary blood flow is using robots that move with the blood, as considered by the average circulation model of this paper. The robots collect data while passing through a capillary. By collecting sufficient data to uniquely identify the capillary and their location within it, post-processing could match data collected by different robots, at different times, from the same capillary, thereby reconstructing longitudinal observations with single-cell resolution. Instead of continuously monitoring cells, this would collect snapshots each time a robot passes through a capillary near the cell. 

On average, each of $n$ robots completes a circulation in $t=60\,\second$ and passes through about $1.25$ capillaries in the systemic flow (including a portion portal flows that pass through more than one capillary during a single circuit~\cite{feher17}). Thus, on average, robots pass through a given capillary at the rate $r=1.25 n/t/N$ where $N\approx 2 \times 10^{10}$ is the number of capillaries in the systemic circulation~\cite{freitas99}.
The average time between successive robots is $1/r$.
For example, with $n=10^{12}$ robots, a robot passes through a capillary about once a second, though with considerable variation around this average value.
If this is adequate temporal resolution for the task, and robots can collect data while moving with the flow instead of, e.g., requiring probes into the vessel wall, then collecting data as robots move  through capillaries is a viable alternative to robots attached to the wall. In either case, the robots would mainly be active, and consuming oxygen, during the time they are in capillaries, so \fig{capillary} describes their average power. This alternative places larger demands on robot data storage since they will need to collect not only the cell measurements of interest but also the data required provide the unique identifications. On the other hand, robots moving in capillary networks could also use their interactions with the flow and other robots to perform distributed microfluidic computation~\cite{prakash07} to provide useful information on the microcirculation.

\subsection{Anemic Patients}\sectlabel{anemia}

\begin{figure}
\centering 
\begin{tabular}{cc}
\includegraphics[width=2.5in]{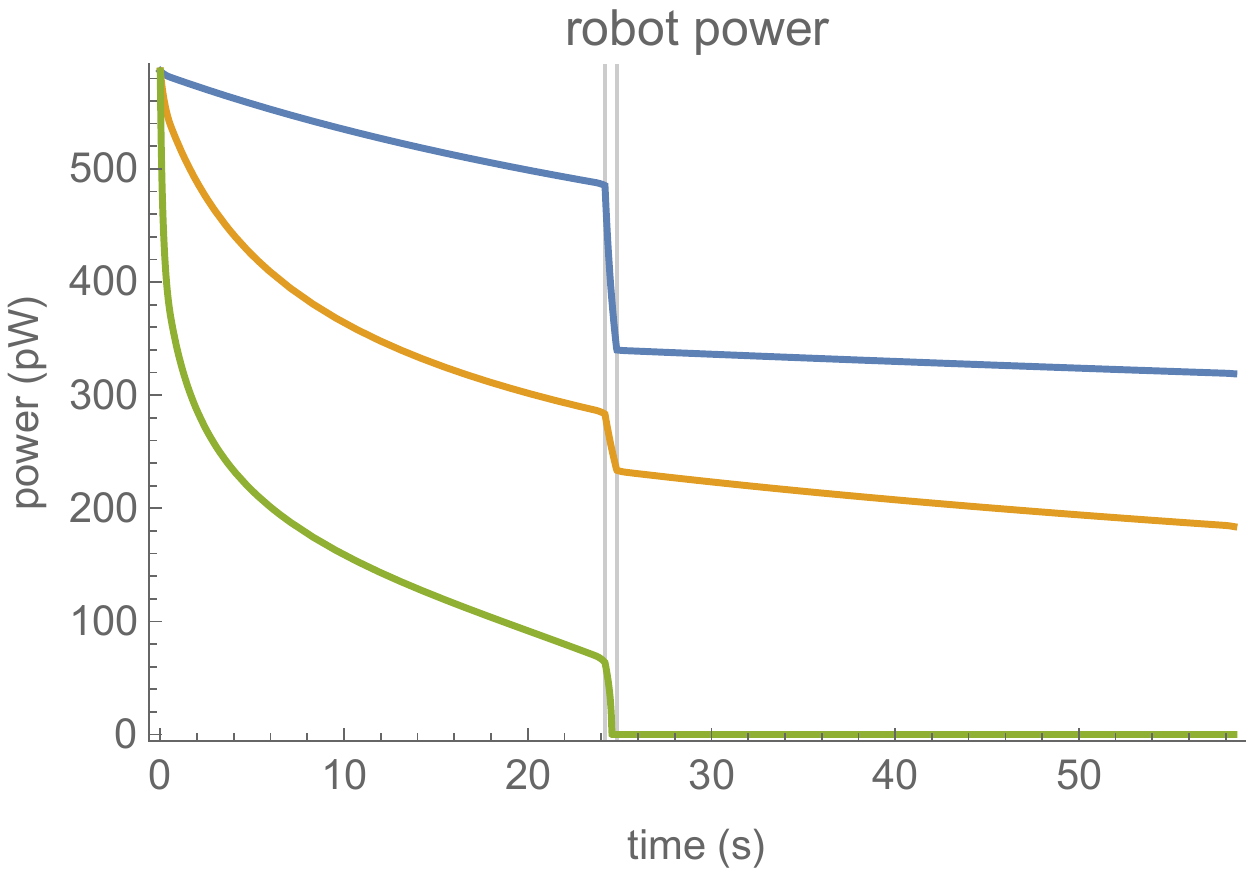}	&	\includegraphics[width=\mfigwidth]{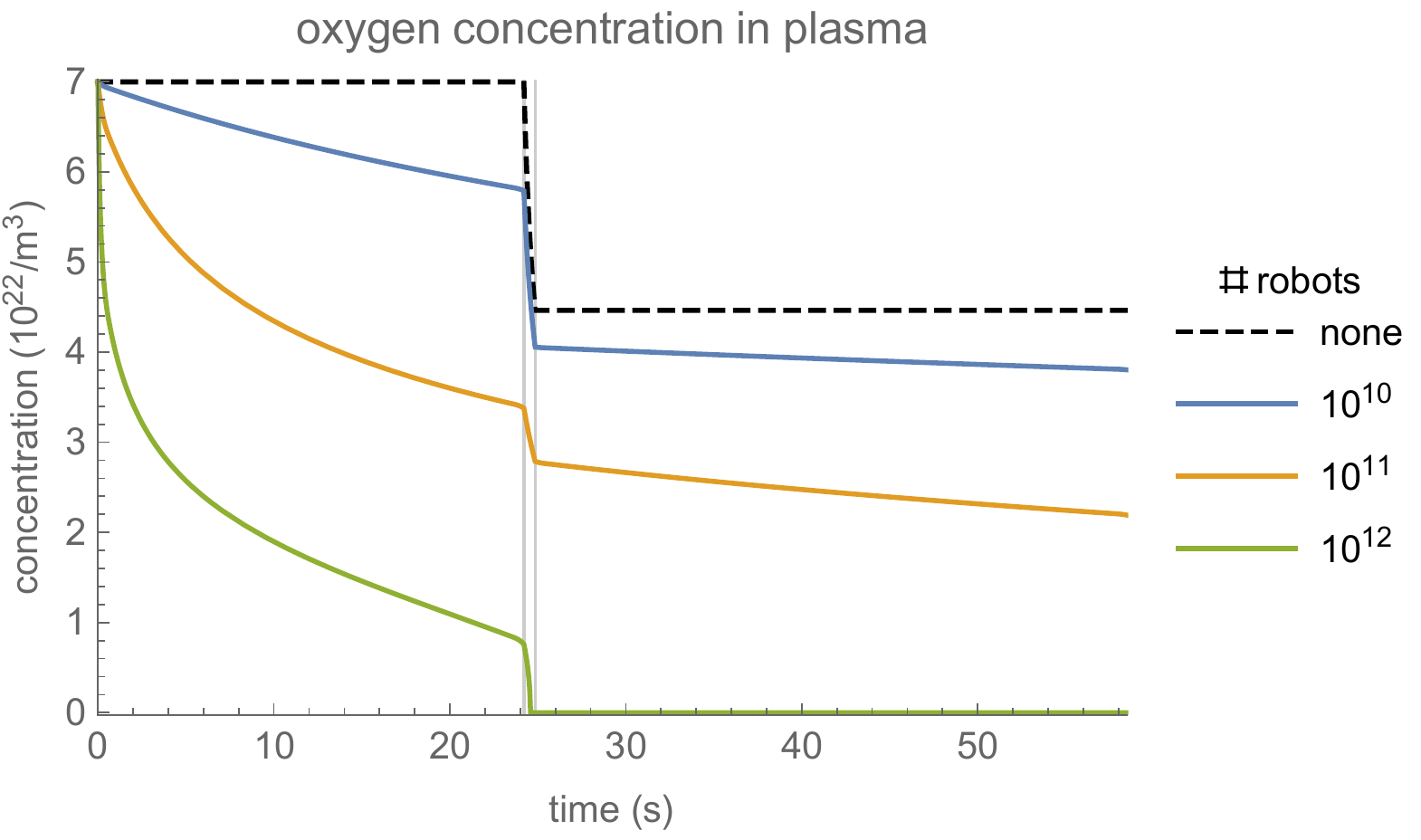} \\
(a) & (b) \\
\end{tabular}
\caption{Behavior with 25\% hematocrit. (a) Robot power. (b) Oxygen concentration in plasma.
The light vertical lines indicate when the blood is in a capillary.}\figlabel{Hct=25}
\end{figure}

Disease and injury could alter the amount of oxygen available to robots.
For example, patients with anemia, i.e., having lower than normal hematocrit, have less ability to replenish the oxygen robots remove from plasma than people with normal hematocrit.

\fig{power} shows power available to robots in an individual with normal hematocrit. This model can illustrate the effect of anemia by reducing the hematocrit parameter.
As an example, \fig{Hct=25} shows behavior with $25\%$ overall hematocrit. 
Even with this lowered hematocrit, the available oxygen from circulating cells provides significant power for all but the largest number of robots considered here. However, $10^{12}$ robots remove all oxygen before blood reaches capillaries. Thus for the lower range of robot number, anemia does not significantly change the situation for robots using chemical power. But the low oxygen reserve in cells with this reduced hematocrit significantly increases oxygen depletion with a large number of robots.

This example shows how the model discussed in this paper can evaluate the suitability of patients for applications of microscopic robots. Personalized versions of such models could help develop mission plans for the robots and provide a baseline to compare with robot measurements during early stages of a mission to identify and respond to deviations from the plan before they become harmful. This extends to microscopic robots the current use of computational models to help plan conventional medical procedures~\cite{taylor09}. 

As described in \sect{conventional sensors}, conventional monitoring may have reduced accuracy when large numbers of robots are consuming oxygen. The variation in response to large numbers of robots based on the patient's health status suggests a staged approach of gradually increasing the number of robots in the body. Another way to achieve a similar effect is by having robots initially limit their power use and gradually increase that limit to the level required for their mission. During this initial stage, the robots could monitor their effect on oxygen concentration and other body processes to determine whether full power would exceed safety limits prior to fully activating all the robots. This evaluation would allow determining patient-specific trade-offs in treatment options. For example, a treatment could use fewer robots or have them operate at lower power with the trade-off of longer treatment duration or less frequent communication with the robots. In particular, monitoring body function during a mission would allow adapting robot behavior and mission parameters to far more detailed measurements than would be available from conventional macroscopic sensors. Continual comparison with personalized versions of the model discussed here could aid in the interpretation of these measurements and anticipate the effect of increasing the number of robots or their power use.

\section{Mitigating the Effects of Robot Oxygen Consumption}\sectlabel{mitigation}

As described in \sect{power}, $10^{12}$ robots lead to significant oxygen reduction and heating when they consume oxygen as fast as possible. Thus the application of microscopic robots could be limited to fewer robots, a shorter mission duration or intermittent operation over a longer period of time. If these adjustments significantly degrade mission performance, altering robot design or behavior are other approaches to mitigate these effects.
This section discusses such approaches, which involve either enhancing the oxygen carried in the blood or robots consuming less of the oxygen available to them.

\subsection{Storing Oxygen}\sectlabel{oxygen storage}

Normally, oxygen collection in the lungs is perfusion limited~\cite{feher17},
i.e., oxygen in lung capillaries is more than sufficient to fully saturate blood plasma and red cells. 
Robots with storage tanks could collect some of this excess oxygen without reducing oxygen collected by red blood cells, thereby increasing the total circulating oxygen. 
This section discusses several ways stored oxygen could supplement the oxygen available to the robots,
with the caveat that 
the amount of additional oxygen depends on the health of the lungs: patients with limited lung capacity may not have sufficient additional available oxygen for robots to collect.

Robots could fill tanks from oxygen in vessels other than lung capillaries. 
However, this would reduce the oxygen available later in the circulation in the same way as robots immediately using all oxygen reaching their surfaces, as shown in \fig{oxygen}a. 
Thus, the discussion in this section focuses on robots obtaining oxygen in lung capillaries.

\subsubsection{oxygen storage tanks}\sectlabel{oxygen tanks}

\begin{figure}
\centering 
\includegraphics[width=\figwidth]{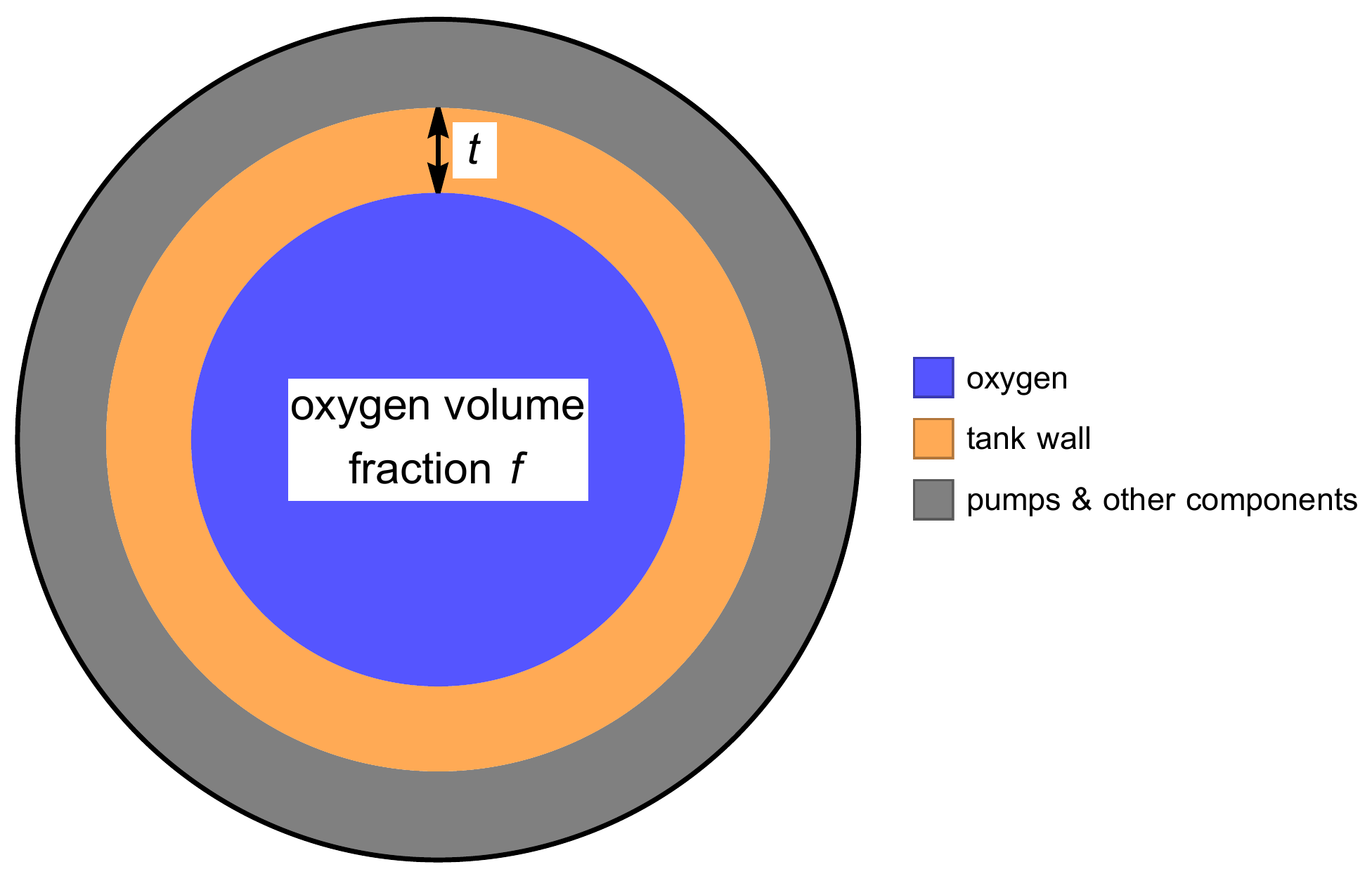} 
\caption{Cross section of an oxygen transport robot with volume fraction $f$ devoted to oxygen storage (blue) and storage tank wall thickness $t$. For clarity, the wall thickness is exaggerated compared to the other dimensions.}\figlabel{oxygen transport schematic}
\end{figure}

A robot stores oxygen in a pressure tank~\cite{freitas99}. 
\fig{oxygen transport schematic} illustrates a robot with a spherical oxygen tank with wall thickness $t$ and interior volume that is a  fraction $f$ of the robot's volume.

The amount of oxygen a robot can store is limited by three factors: the rate at which a robot absorbs oxygen, the time it spends in a lung capillary, and its storage capacity.

\paragraph{Absorption}
Oxygen absorption depends on how fast molecules diffuse to the robot's surface and the fraction of the surface that captures arriving oxygen with pumps.
 \eq{robot absorption} gives the rate oxygen diffuses to the robot surface. The corresponding flux to the surface is $\OxygenAbsorptionRate/(4\pi \rRobot^2)$. For a $1\,\micron$-radius robot $\OxygenAbsorptionRate = 1.8\times 10^9\,\molecule/\second$ in lung capillaries, corresponding to a flux of $1.4\times 10^{20}\,\molecule/\meter^2/\second$.

\begin{figure}
\centering 
\includegraphics[width=\figwidth]{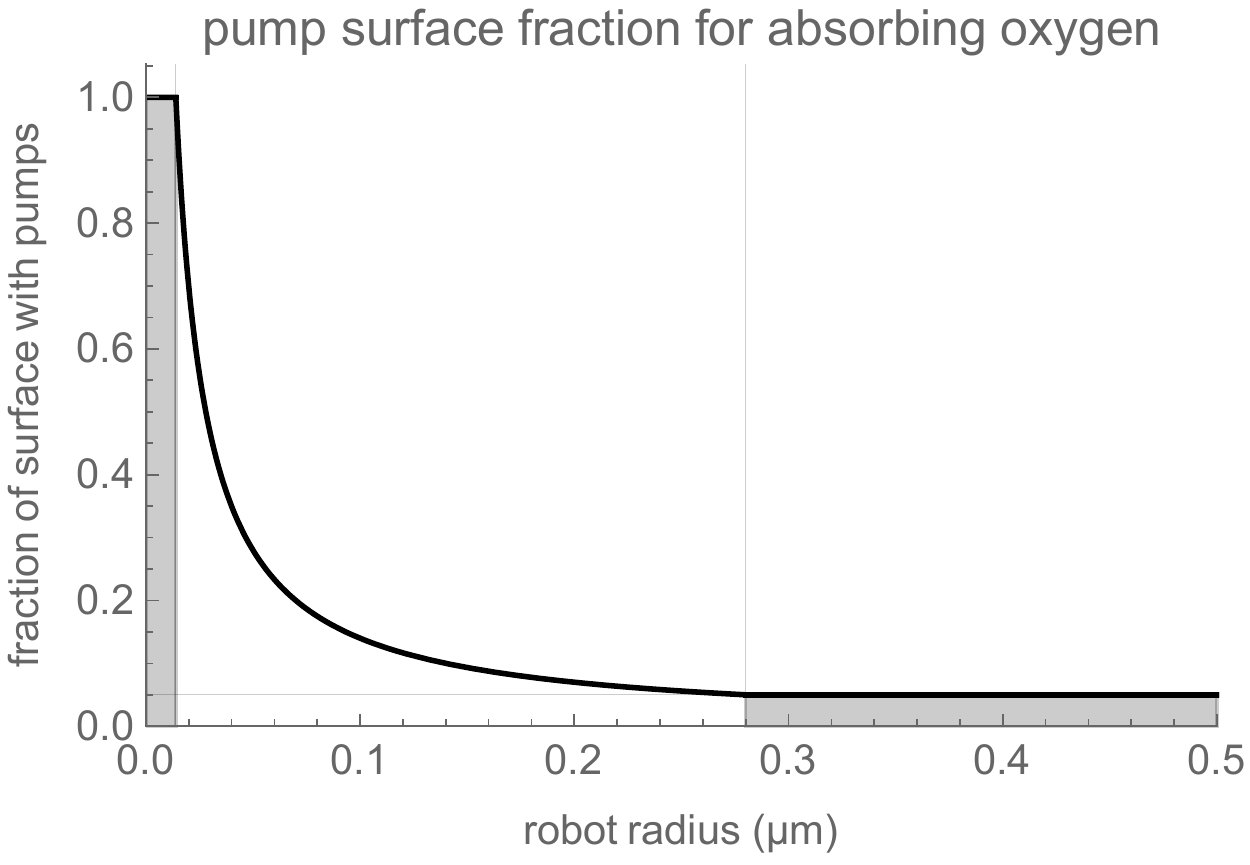} 
\caption{Fraction of robot surface covered with pumps to absorb oxygen in lung capillaries as a function of robot radius. In the shaded area on the left, molecules arrive too fast for pumps to collect, even if they cover the entire surface. In the shaded area on the right, surface coverage is $5\%$, which is sufficient for pumps to capture most of molecules reaching the robot surface. For intermediate sizes, the surface fraction is from \eq{pump surface}.}\figlabel{pump surface fraction}
\end{figure}

For maximum absorption, the pumps must have sufficient capacity to collect molecules as fast as they reach a pump. However, pumps have a maximum operating rate $\PumpOxygenAbsorptionRate = (4\pi \rRobot^2) s \PumpMaxFlux$ where $s$ is the fraction of the robot surface used by pumps and \PumpMaxFlux\ is their maximum pump capacity. For molecular sorting rotors, a plausible capacity is $ \PumpMaxFlux = 10^{22}\,\molecule/\meter^2/\second$~\cite{freitas99}.
Capturing all the molecules requires that $\PumpOxygenAbsorptionRate \ge \OxygenAbsorptionRate$. With \eq{robot absorption}, the minimum surface coverage is
\begin{equation}\eqlabel{pump surface}
s = \frac{\Doxygen c}{\PumpMaxFlux} \frac{1}{\rRobot}
\end{equation}
with $c$ the oxygen concentration in lung capillaries.

Another constraint on pump surface fraction arises from diffusion. When pumps do not cover the entire surface, an oxygen molecule diffusing to the robot's surface will not necessarily reach a pump. Nevertheless, once a molecule reaches any part of the robot's surface, it usually moves for a considerable time near the surface before diffusing away. 
This means that pumps that only cover a few percent of the surface can collect oxygen nearly as well as a completely absorbing surface~\cite{berg93}. This provides a lower limit on the fraction of the surface used by pumps.
An additional constraint arises from the multiple uses a robot has for its surface, e.g., for sensing and structural support. Thus pumps cannot occupy the entire surface.

These three constraints (pump capacity, diffusion rate to the robot surface and fraction of the surface available for pumps) combine to determine the rate robots can absorb oxygen. \fig{pump surface fraction} illustrates how these constraints determine the fraction of surface area needed for pumps to absorb most of the arriving oxygen when the robot is in a lung capillary as a function of robot size.
This example uses a minimum fraction of $5\%$, which allows capturing most arriving oxygen~\cite{berg93}. At small sizes, even if pumps cover the entire surface, they cannot absorb all the oxygen reaching the surface. However, even if the other components a robot needs on its surface limit pumps to no more than $25\%$ of the surface, \fig{pump surface fraction} shows that the limit due to pump capacity only applies to robots significantly smaller than used in the scenarios of this section.

\paragraph{Time} Blood passes through a lung capillary in about $0.75\,\second$~\cite{feher17}. We use this as the oxygen collection time for robots moving with the blood, although robots might have somewhat less time as they move with cells, due to the reduced hematocrit in small vessels (see \sectA{hematocrit}).
Robots could increase the filling time in several ways. Robots moving passively with the blood could use several circulations through the lungs to fill their tanks. Active robots could extend their filling time by sticking to capillary walls or selecting longer routes through the lung capillary network.

\paragraph{Capacity} The tank stores oxygen at high pressure. For a spherical tank of radius $r$ with a thin wall of thickness $t$, Laplace's law gives its maximum pressure as~\cite{freitas99}
\begin{equation}\eqlabel{Laplace's Law}
\pTankMax = \frac{2 t \sigma}{r}
\end{equation}
where $\sigma$ is the wall's failure strength. For a wall formed of covalently-bonded carbon, a conservative estimate of the failure strength is $\sigma = 10^{10}\,\pascal$, about $20\%$ of diamond's failure strength~\cite{freitas99}.
For example, a tank with radius $r=0.3\,\micron$ and wall thickness $t=5\,\nanometer$ has maximum pressure near $3000\,\mbox{atm}$.

The scenarios discussed in this section use tanks storing oxygen at about one-third of this maximum pressure.
At these storage pressures and body temperature, common gases such as oxygen deviate somewhat from the ideal gas law. To account for this deviation, we estimate the tank storage capacity using  the van der Waals equation of state for oxygen~\cite{freitas99}.

\subsubsection{diffusion-limited storage}\sectlabel{diffusion storage}

During a $0.75\,\second$ transit through a lung capillary~\cite{feher17}, 
 $1.3\times 10^9$ molecules diffuse to the surface of a $1\,\micron$-radius robot.  
For example, at body temperature and tank pressure of $1000\,\mbox{atm}$, the number density of oxygen molecules is $1.26\times10^{28}\,\molecule/\meter^3$~\cite{freitas99}, which is far larger than the concentration in blood (see \tbl{parameters}). Thus a tank with radius $0.3\,\micron$ could store all the collected molecules. This tank occupies about $3\%$ of the robot's volume, including a $5\,\nanometer$-thick wall.

From \eq{robot power}, a robot collecting oxygen as fast as it diffuses to its surface while passing through a lung capillary collects enough oxygen to provide about $7\,\picowatt$ for the duration of a $60\,\second$ circulation loop.
This is well below the power available from oxygen in the blood during the circulation with even as many as $10^{12}$ robots, except for  the last ten seconds or so of the loop (see \fig{power}). If the stored oxygen were only used during the last ten seconds of the circulation, it would provide about $50\,\picowatt$.
Thus this oxygen storage is not sufficient for robots requiring around $100\,\picowatt$ or so, and hence can not completely alleviate depletion by large numbers of robots consuming oxygen.
However, stored oxygen would be useful as a supplemental power source for brief intervals of reduced oxygen, such as when a robot is next to a white blood cell moving through a capillary, so the plasma around the robot is temporarily not replenished by nearby red blood cells.

\subsubsection{tank capacity-limited storage}\sectlabel{tank capacity storage}

As an illustration of oxygen storage, suppose robots collect enough oxygen to provide $100\,\picowatt$ for the $60\,\second$ duration of the average circulation loop, which corresponds to $1.8\times 10^{10}$ oxygen molecules.
Storing this much oxygen in a pressure tank at one-third its maximum pressure requires about one-third of the robot volume for a tank with $10\,\nanometer$ thick walls.
Filling this tank requires $10\,\second$ in a lung capillary, which is much longer than typical transit times. To fill its tank, a robot could remain in a lung capillary rather than flow with the blood, e.g., by anchoring itself to the capillary wall. Lung capillaries contain $70\,\milliliter$ of blood~\cite{feher17}. Thus if robots stay in capillaries long enough to fill their tanks, lung capillaries would hold about $15\%$ of the robots in only $1.2\%$ of the blood volume.

For $10^{12}$ robots, this scenario gives a nanocrit of $0.01$ in lung capillaries, about ten times larger than the overall nanocrit (see \tbl{scenarios}). The distance between robots in a capillary is correspondingly reduced, to about $10\,\micron$. This spacing is small enough that nearby robots compete for oxygen, thereby somewhat increasing the filling time beyond that determined here for isolated robots~\cite{hogg10}.

These estimates indicate the trade-offs required for robots to carry enough oxygen for $100\,\picowatt$ without consuming oxygen from the blood during the circulation.
In particular, this scenario requires a large fraction of robot volume for storage, significantly increases robot concentration in lung capillaries and requires robots have the capability to anchor themselves in capillaries.

\subsubsection{oxygen transport by specialized robots}\sectlabel{oxygen storage robots}

An alternative to oxygen tanks in the robots performing the medical mission is to use additional specialized oxygen transport robots~\cite{freitas98}. These robots collect oxygen while passing through the lungs and deliver it to the main mission robots in the systemic circulation.\footnote{This oxygen delivery contrasts with transport robots that also collect carbon dioxide for return to the lungs~\cite{freitas98}. Such robots require power to collect and compress molecules while in the systemic circulation, which robots that only deliver oxygen do not need.}  

Oxygen transport robots face the same limit on extracting oxygen from lung capillaries as discussed above, depending on their size and tank capacity. However, the transport robots can be optimized to address these limitations without compromising the main mission. 

For example, in the scenario of \sect{tank capacity storage}, robots cannot fill their tanks during a single transit through a lung capillary. Oxygen transport robots can fill their tanks in this time by exploiting the geometry of diffusion: from \eq{robot absorption} the rate molecules arrive at the robot is proportional to its radius \rRobot, while tank capacity is related to its volume, which is proportional to $\rRobot^3$ when tanks use a fixed fraction of the robot volume and a fixed storage pressure. Thus the time to fill a tank scales as $\rRobot^2$. This means that instead of a single large tank in one robot, using proportionally smaller tanks in many robots, with the same overall storage capacity, can collect the oxygen in a shorter time. 
Thus, for example, decreasing the robot size by a factor of two reduces the time required to fill its tank by a factor of four. Increasing the number of robots by a factor of eight gives the same total volume of robots and the amount of oxygen they can transport.

To quantify the design trade-offs for these robots, consider $n$ oxygen transport robots, each of radius $r$ with a fraction $f$ of its volume for oxygen storage, as shown in \fig{oxygen transport schematic}. Scaling the entire geometry with robot size means the thickness of the oxygen tank wall is proportional to $r$. 
From \eq{Laplace's Law}, this scaling has the same maximum pressure for the storage tank in robots of different sizes.
Specifically, the wall thickness in this example is $t = 20\,\nanometer \times (r/\rRobot)$ where $\rRobot = 1\,\micron$ is the radius of the main robots. The wall cannot be less than one atomic layer, so this scaling does not apply for arbitrarily small robots. E.g., with a minimum $t \ge 1\,\nanometer$ thickness, this relation applies for $\rRobot \ge 0.05\,\micron$. The transport robots considered here are larger than this size. 
As with the example of \sect{diffusion storage}, tanks store oxygen at one-third of their maximum pressure.

With this specification of the transport robots, we select the number of robots, their size and volume devoted to oxygen, i.e., values for the parameters $n$, $r$ and $f$, so the robots, in aggregate, carry $1.8\times 10^{10}$ oxygen molecules for each of the main robots, duplicating the scenario of \sect{tank capacity storage}. 
In addition to this requirement, we consider three constraints on the choice of these parameters for transport robots.

First, the robots should be small enough so the favorable diffusion scaling allows them to fill their tanks with a single transit of the lung capillaries. This includes setting the fraction of the surface used by pumps as shown in \fig{pump surface fraction} so the robots can collect the oxygen available to them in the capillary. 

Second, the volume of the robot outside its oxygen tank must be sufficient to contain its other components.
These additional components include pumps to collect oxygen and mechanisms to control the operation of the robot, e.g., to determine oxygen concentration in the fluid around the robot and when to collect or release oxygen.
For the volume used by the pumps, we assume each pump has size $d=10\,\nanometer$~\cite{freitas99} so the total volume of pumps is $4\pi r^2 s d$.
We take the controller and other components to require a fixed minimum volume $v=0.1\,\micron^3$ that was estimated for robots delivering oxygen to tissue~\cite{freitas98}. 
Thus the required volume for components other than the storage tank is
\begin{equation}\eqlabel{other components}
v + 4\pi r^2 s d
\end{equation}

The third constraint is on energy required by the transport robots. Their main energy use is to fill their tank. This consumes part of the received oxygen, thereby increasing the filling time. Comparing the pump energy required to compress oxygen~\cite{freitas99} with the rate robots receive oxygen in lung capillaries (see \sect{oxygen tanks}) indicates a robot will only need about $3\%$ of the oxygen it collects to operate the pumps. Thus filling the tank has a negligible effect on filling time.
Moreover, much of this pumping energy may be recovered with a
generator using the subsequent expansion of the gas when it is released to 
the lower partial pressure outside the robot~\cite{freitas98}.

Robots need energy for their operation after leaving the lung. This includes sensing when oxygen concentration is low and determining when to release oxygen, which are tasks that do not depend on robot size. 
Transport robots could obtain energy using oxygen from the blood in the same way as the main robots. However, that would decrease the oxygen available to the main robots, partially negating the use of transport robots to mitigate oxygen reduction in the blood. Instead, the transport robots could use their stored oxygen. For effective oxygen transport, this consumption must be a small fraction of the stored oxygen so a robot can deliver almost all its oxygen to the main robots. 

One measure of this power requirement is the average use over the $60\,\second$ circulation loop. 
For instance, $0.1\,\picowatt$ is an estimate of the power needed for computation by robots supplying oxygen to tissue~\cite{freitas98}.
After transport robots leave the lung, their main activity is determining when to release stored oxygen, so computation determines their energy demand. Thus $0.1\,\picowatt$ is a plausible estimate of their required power for continuous operation. In practice, transport robots would only need to check oxygen concentrations intermittently, thereby reducing their average power use below this value.
Thus, a reasonable requirement for a transport robot is that its tank can hold enough oxygen to provide at least $1\,\picowatt$, on average, during a circulation loop. This provides sufficient energy for the robot while using only a small portion of its stored oxygen.

\begin{figure}
\centering 
\includegraphics[width=\figwidth]{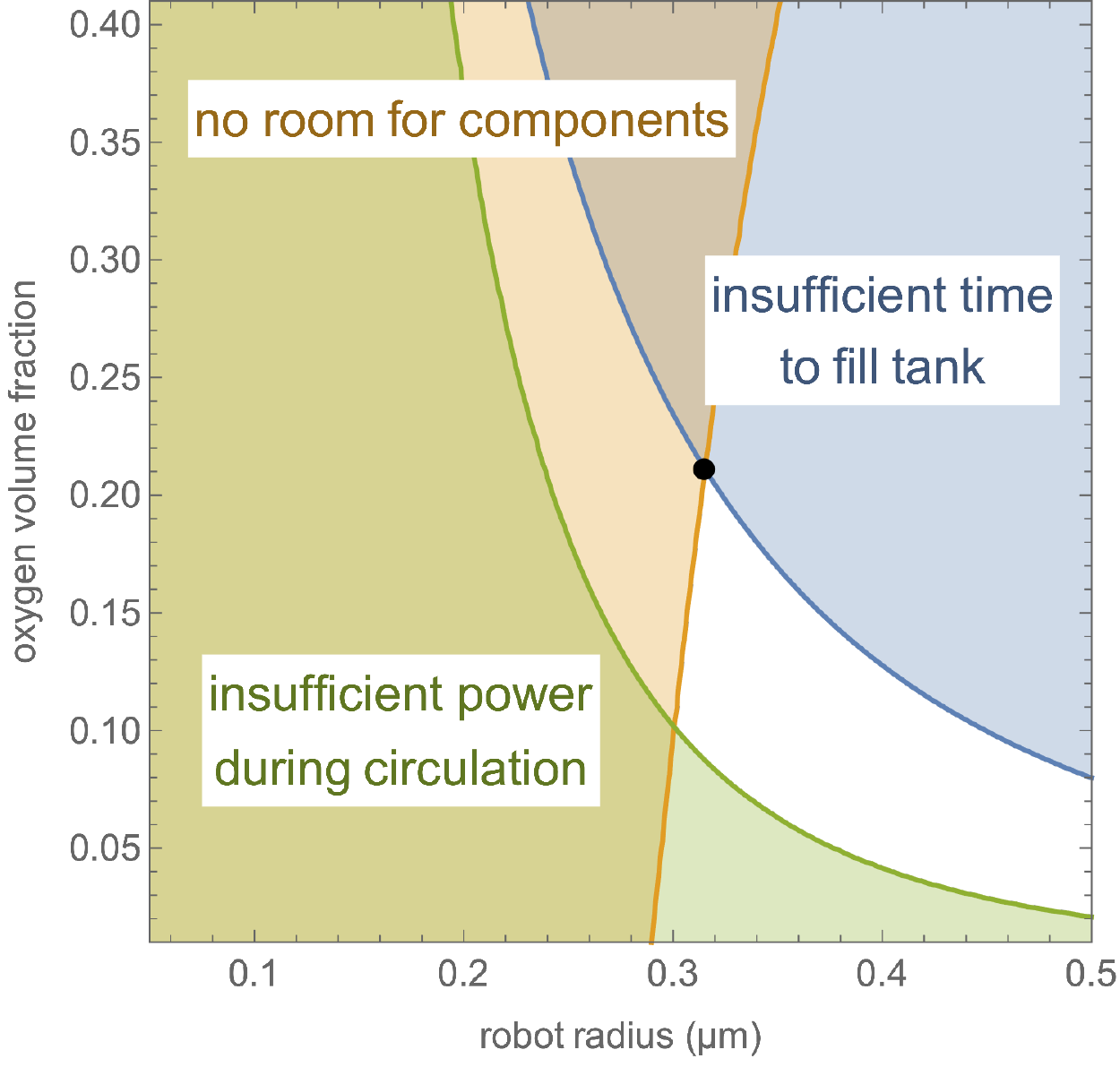}
\caption{Design constraints on oxygen transport robots. The shaded regions indicate combinations of robot radius and volume fraction for oxygen storage that violate one or more of the three constraints described in the text. The black point is the design choice satisfying the constraints with the smallest value of the total volume of the transport robots given in \eq{total volumes}.}\figlabel{oxygen transport robot constraints}
\end{figure}

\fig{oxygen transport robot constraints} shows the constraints on robot and tank size for the scenario of \sect{tank capacity storage}, namely when transport robots carry enough oxygen to provide each main robot with $100\,\picowatt$ for $60\,\second$.
The figure shows three constraints. At the upper right, diffusion is too slow to fill the tanks during a single lung capillary transit.
At the left, the robot is too small and the oxygen tank too large to leave enough room for other robot components. Finally, at the lower left, the oxygen tank is too small to both provide enough power to the transport robot during the circulation and deliver most of its oxygen to the main robots.

Robots satisfying all the constraints, i.e., the unshaded region of \fig{oxygen transport robot constraints}, are feasible designs. 
Selecting among these designs can optimize operational or production goals.
Operational goals include minimizing the nanocrit, i.e., the total volume of the transport robots and reducing their size to simplify passage through small vessels or slits of the spleen~\cite{freitas03}.
Production cost depends on the number of robots and the manufacturing cost of each robot. A proxy for manufacturing cost is the number of atoms required for their structure and mechanisms. 
This proxy is proportional to the volume of the robot other than the interior of the oxygen tank. Thus a proxy for production cost is the total volume of all the transport robots multiplied by $1-f$:
\begin{equation}\eqlabel{total volumes}
\begin{split}
V_{\textnormal{total}}  &= n \times \frac{4\pi}{3} r^3 \\
V_{\textnormal{production}}  &= (1-f) V_{\textnormal{total}} 
\end{split}
\end{equation}

\begin{table}
\centering
\begin{tabular}{cc}
property	& ratio \\ \hline
number of robots		& $43$ \\
total volume of robots		& $1.4$ \\
total manufactured volume of robots		& $1.1$ \\
\end{tabular}
\caption{Ratio of properties of optimal transport robots to those of the main robots, The transport robots have radius $r=0.32\,\micron$ and oxygen volume fraction $f=0.23$, corresponding to the black point in \fig{oxygen transport robot constraints}.
}\tbllabel{optimal transport}
\end{table}

An example of optimized parameters are those that minimize the total volume of the transport robots, $V_{\textnormal{total}} $. These parameters correspond to  the black point in \fig{oxygen transport robot constraints}, and also
minimize $V_{\textnormal{production}}$ in \eq{total volumes}.
Thus this choice of $r$ and $f$ provides both an operational benefit of minimizing the nanocrit and a production benefit of minimizing the volume proxy for manufacturing cost.
\tbl{optimal transport} describes the transport robots and how they compare to the main mission robots.\footnote{For comparison, the base design for gas transport to and from tissue~\cite{freitas98} is a robot with $0.5\,\micron$ radius and $50\%$ of the volume used for gas storage, divided equally among oxygen and carbon dioxide. It devotes about $0.1\,\micron^3$ to components other than the tanks.}
With these parameters, carrying the same amount of oxygen per main mission robot as in the  scenario of \sect{tank capacity storage} requires about 43 transport robots for each main robot. The main and transport robots together have a nanocrit about $2.4$ times larger than for the main robots alone.

\begin{figure}
\centering 
\begin{tabular}{c|c}
\includegraphics[width=\mfigwidth]{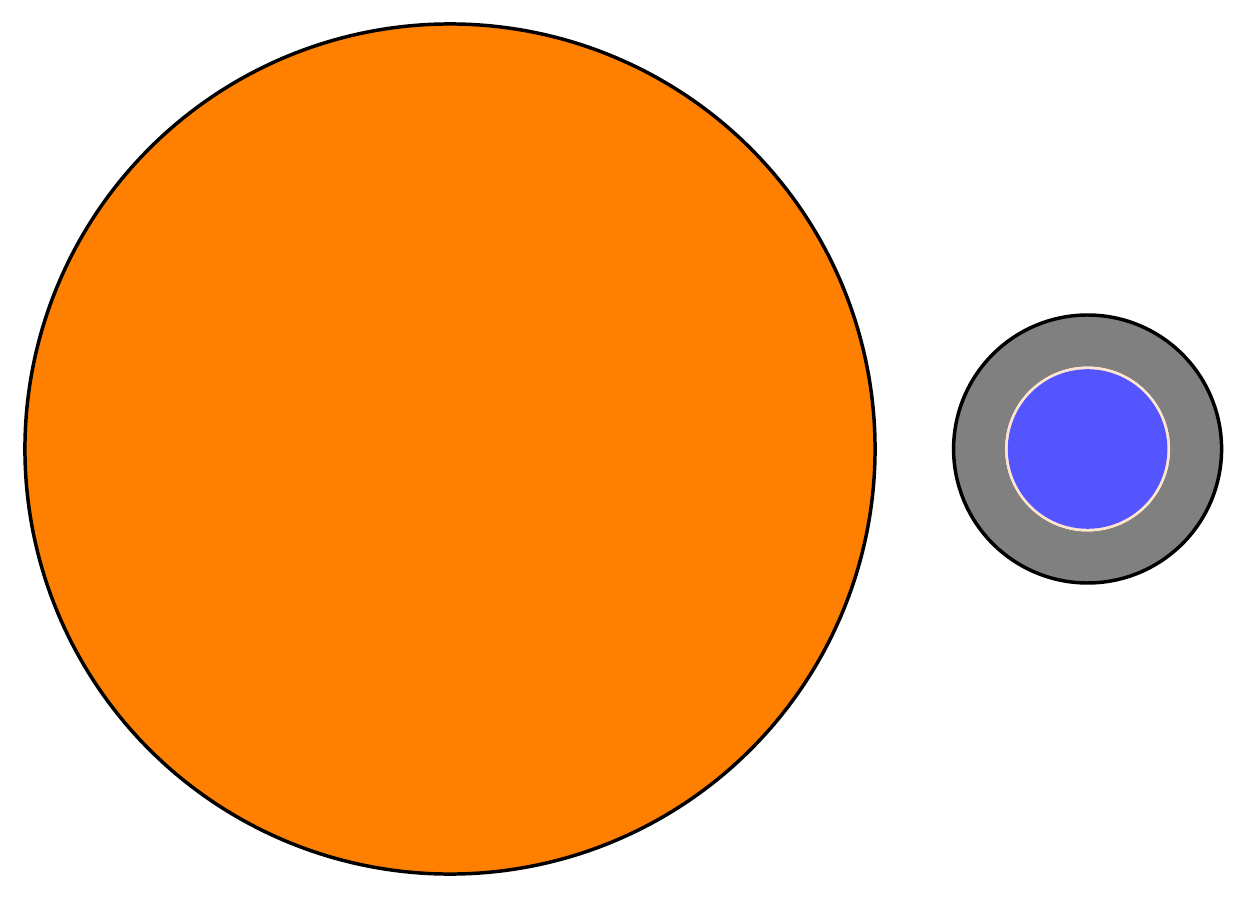}	&	\includegraphics[width=\mfigwidth]{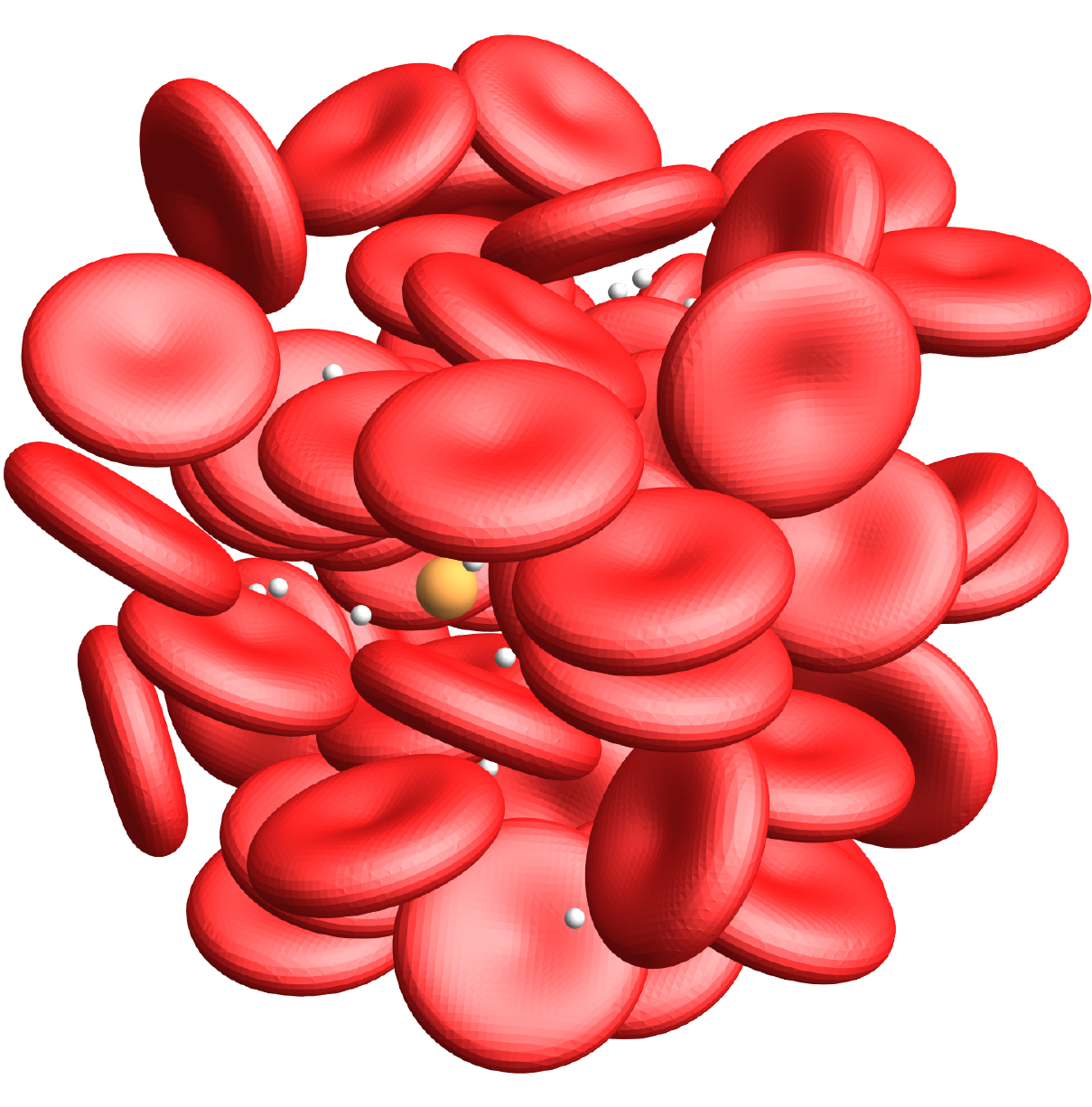} \\
(a) & (b) \\
\end{tabular}
\caption{(a) Cross sections of the main robot (orange, $1\,\micron$ radius) and a transport robot with parameters in \tbl{optimal transport} and oxygen storage volume indicated by the blue disk. (b) A main robot and nearby transport robots with blood cells.
The main robot, indicated in orange, is near the center. Transport robots are the smaller white spheres. 
}\figlabel{oxygen transport robot}
\end{figure}

\fig{oxygen transport robot} illustrates this optimized design by comparing the oxygen transport and main mission robots, and a typical placement of robots in a cube of blood about $20\,\micron$ across. 
For $10^{12}$ main robots, such a cube contains, on average, one of these robots (see  \tbl{scenarios}), and 43 transport robots (see  \tbl{optimal transport}), only a few of which are visible in the figure. 
The cells are shown as biconcave shapes~\cite{kuchel21} with about $10\%$ variation in volume around a typical red cell volume of $100\,\micron^3$~\cite{freitas99}.
This discussion of the number of robots mixed among cells applies to cells and robots uniformly distributed in a blood vessel whose diameter is larger than the size of the cube. This is an approximation to the actual distribution since smaller objects tend to concentrate toward the vessel wall~\cite{freitas99}.

Due to their larger numbers, the spacing between transport robots in capillaries is smaller than those for the main robots given in \tbl{scenarios}. However, the transport robots are also smaller, so their spacing measured in terms of their size is reduced to a lesser extent. For example, with  $10^{12}$ main robots, transport robots are separated by about 8 times their radius in straight capillaries used for the estimate in \tbl{scenarios}.
This suggests that neighboring transport robots compete to some extent for oxygen in lung capillaries, which will somewhat increase their filling time compared to the estimate assuming independent diffusion to each robot. 
However, lung capillaries have a complex network structure~\cite{weibel62}
which could alter the extent of this competition, especially due to the variation in paths and transit times. Extending studies of how cells move through these vessels~\cite{stauber17} to hard particles of size comparable to these robots could quantify this effect.

For the transport robots shown in \fig{oxygen transport robot}, the pump volume is a small fraction of the component volume in \eq{other components}. If pump size and surface coverage are larger than assumed here, an additional optimization is allowing for a smaller number of pumps than indicated by \fig{pump surface fraction}. This trades a longer filling time due to fewer pumps for additional volume for other robot components. This would give another parameter to optimize, in addition to robot radius and tank size shown in \fig{oxygen transport robot constraints}.

The specific optimal robot and tank sizes shown in \fig{oxygen transport robot} depend on the robot component size and power requirements (i.e., the second and third constraints in \fig{oxygen transport robot constraints}). This example illustrates how multiple design constraints combine to determine the choice of robot parameters. 
These constraints arise both from the robot's external environment (e.g., the diffusion rate of oxygen) and the robot's internal capabilities (e.g., the power required for its operation).
Better estimates of these parameters require more detailed design of the robot components. 
Depending on these values, there could be no feasible design, i.e., no unshaded region in \fig{oxygen transport robot constraints}.
That would indicate the robots' component volumes or power requirements are too large to provide this oxygen. In that case, oxygen transport robots could provide less than the $100\,\picowatt$ used in this scenario, or the main robots could use another mitigation strategy to avoid low oxygen concentration.

Using transport robots increases nanocrit. While likely not a significant issue for the scenarios of \tbl{scenarios}, large nanocrit from transport robots could alter blood flow~\cite{freitas99} or require a compensatory reduction in the number of main mission robots.
Moreover, increasing the number of robots collecting oxygen in the lungs will eventually extract all available oxygen. At this point, additional robots will not increase the total collected oxygen and would reduce the reserve available to the body by increasing blood perfusion in the lungs.

Transport robots carry oxygen from the lungs to the systemic circulation.
In the case of passive flow, transport robots release oxygen into the surrounding blood plasma when they detect low oxygen concentrations. Some of that released oxygen diffuses to nearby robots. The rest  remains in the blood plasma or diffuses into tissue or red cells. This means the blood carries some of the released oxygen back to the lung without providing robot power. 
This wasted oxygen is particularly significant when released in the veins, where it will not diffuse into tissue. This is a likely scenario since the lowest concentrations occur in veins (see \fig{oxygen}a). 
This diffusive oxygen delivery by specialized robots is not as effective as when robots carry their own oxygen, as described in \sect{tank capacity storage}. 

Robots could partially offset the limitation of diffusive transport by releasing oxygen only when they are close to a main mission robot, as determined, for example, via short range communication~\cite{freitas99,hogg12}. This proximity allows more of the released oxygen to reach the receiving robot's surface. If necessary to support high burst power, multiple transport robots could aggregate around a main one and simultaneously release oxygen. This will temporarily produce a high oxygen concentration in that region. However, as not all of that oxygen will reach the robot, the total release must be limited to avoid tissue damage due to excessive oxygen, i.e., hyperoxia, which is a particular concern in the brain~\cite{wilson19}. This safety limit due to released oxygen not reaching a main robot limits how rapidly a group of transport robots can deliver oxygen, thereby reducing their ability to support high burst power that robots could obtain if they carry their own oxygen.

For greater efficiency, oxygen transport robots could directly transfer oxygen to the main robots by docking with them. 
Direct transfer provides oxygen in the same way as an onboard tank. In this approach, transport robots act as external oxygen tanks for the main robots. Using separate robots to carry oxygen allows them to selectively provide power to robots that most need it. 
For example, this selectivity could allow one robot in a group to have a burst of high power for long range communication of a summary of data collected by the group. While this flexibility is a potential benefit, some amount of power variation within a group of robots can also be achieved if most robots limit their oxygen demands, as described in \sect{limiting power}.

Direct transfer requires more complicated robots than releasing oxygen into the blood and relying on diffusion.
In particular, docking requires locomotion and navigation on the part of the transport or main robots, or both, to find each other in the constantly changing fluid environment as the robots and cells move.
If the main robots need locomotion capability for their mission, they could also use that capability to reach oxygen transport robots. In that case, there is no need for the transport robots to also have locomotion capability, thereby simplifying their design and providing more room for them to carry oxygen.
An additional issue is if a transport robot delivers oxygen by completely emptying its tank while docked, direct oxygen delivery would produce a population of transport robots with a declining fraction of those with full tanks. So robots with full tanks will become harder to find and may need a communication protocol to identify transport robots that have available oxygen.

\subsubsection{oxygen reservoirs formed by specialized robots}

\fig{power} shows $10^{12}$ robots deplete oxygen toward the end of the circulation loop, i.e., in veins. 
One approach to mitigating this oxygen depletion is to deliver oxygen reservoirs to those vessels.
These reservoirs could be filled with oxygen prior to their placement in the body, and thereby supplement oxygen available via the lungs until the oxygen they carry is consumed.
Examples of such single-use devices include bubbles enclosed in biocompatible particles of sizes from about $100\,\nanometer$ to several microns~\cite{khan18} that have been used to enhance ultrasound imaging in the body. These particles can release oxygen gradually by diffusion or rapidly in specific areas by disrupting them with resonant ultrasound.
In addition, implanted biodegradable microscopic tanks that release oxygen by diffusion have been experimentally demonstrated in tissue scaffolds~\cite{cook15}. By relying on diffusion, these tanks require no power or control, but their rate of oxygen release decreases exponentially with time. The demonstrated time constants for this decrease is several hours, which could support robot missions of such durations, but the decreasing oxygen release could limit how well oxygen delivery from a reservoir matches robot demand. 
The development of more sophisticated oxygen delivery robots~\cite{freitas98} could provide a reservoir that supplements oxygen from the lungs in a more effective way.

A rechargeable reservoir with control over oxygen release could be formed from robots small enough to travel through the circulation and able to attach to vein walls. These would be the same type of circulating oxygen-carrier robots discussed in \sect{oxygen storage robots} but used as a reservoir fixed to vessel walls rather than traveling in the blood along with the main mission robots.

As an example of this scenario, consider a reservoir capable of supplying enough oxygen for $100\,\picowatt$ to $10^{12}$ robots during the last $20\,\second$ of their circulation. 
The robots in the reservoir could do so by releasing oxygen into the blood, but only a portion of that would diffuse to the robots, the rest would return to the lung unused. Instead, suppose circulating robots can find and dock with the reservoir robots to receive oxygen directly. 

Due to the uniform distribution of the main mission robots considered here, the last $20\,\second$ of the circulation contains about a third of these robots. To provide these robots with sufficient oxygen to generate $100\,\picowatt$, the reservoir would need to supply $10^{20}\,\molecule/\second$.
To maintain this rate for a single circulation time, the reservoir would need to start with $6\times 10^{21}$ oxygen molecules.
As an example, suppose the reservoir consists of $1\,\micron$-radius robots using $80\%$ of their volume to store oxygen in tanks with $20\,\nanometer$-thick walls and at one-third the tanks' maximum pressure. Such a robot could store $4.6\times 10^{10}$ molecules and the exterior volume of the tank, i.e., including the shell, would occupy $85\%$ of the robot's volume. If a reservoir robot uses $10\,\picowatt$ to support its operation, it would consume about $4\%$ of its stored oxygen during a $60\,\second$ operation time. Thus the reservoir would require $1.4\times 10^{11}$ robots to provide oxygen to the $10^{12}$ main mission robots.

A reservoir robot with these parameters would require $26\,\second$ to fill its tank in a lung capillary, compared to the typical $0.75\,\second$ transit time through a lung capillary. If the robots move passively with the circulation rather than concentrating in lung capillaries for the required time, and the robots do not use power while circulating, each would require $35$ circulations through the lung to fill its tank, which would take about half an hour.

Thus in this scenario, using a reservoir would provide oxygen to support $100\,\picowatt$ for robots during the last third of a circulation loop only once in every $35$ circulations. In the remaining circulations, the robots would have little power (see \fig{power}) during the last third of their circulation, or no power if they stop using power to avoid extremely low oxygen concentration and cell saturation toward the end of the circulation loop (see \fig{oxygen}).

\subsubsection{red blood cells as oxygen reservoirs}
Red blood cells are an oxygen reservoir in the blood. Robots, which are in the blood plasma, extract oxygen from nearby red cells indirectly: robots consume oxygen in the plasma, lowering the concentration in the plasma, which leads to partial replenishment with oxygen released from hemoglobin in the cells. Only some of the oxygen released from cells diffuses to nearby robots. 

The robots considered here have volume of a few cubic microns, which is considerably smaller than the typical $100\,\micron^3$ volume of a red blood cell. Thus, if robots are able to enter red blood cells, robots could extract oxygen directly from hemoglobin, e.g., by collecting oxygen-bound hemoglobin in a tank from which oxygen is continually pumped out. This process would be limited by the kinetics of oxygen release from hemoglobin~\cite{clark85} rather than the diffusion rate of oxygen in plasma. The robot could promote this oxygen release by altering the chemical environment of the hemoglobin in its tank, e.g., by adding carbon dioxide. 

This use of red blood cells as reservoirs could increase the rate robots obtain oxygen compared to operating in plasma. However, in cases where robots in plasma consume most of the oxygen by the end of the circulation loop, this would not provide more oxygen in total during the circulation. Instead, it would shorten the time until robots deplete the oxygen. Moreover, if robots inside cells reduce cell saturation below the equilibrium value with the surrounding plasma, then cells will absorb rather than release oxygen into the plasma. When passing through capillaries, this would lead to cells competing with tissue for oxygen, rather than replenishing oxygen in the plasma in response to tissue consumption. Evaluating the extent to which this occurs could be done by extending the model to include direct reduction in hemoglobin saturation within cells based on the robot consumption rate.

\subsection{Limiting Robot Power}\sectlabel{limiting power} 

\fig{power} shows a large variation in the power available to robots as they move through the circulation. Robots could mitigate their effect on oxygen concentration by limiting their power use, particularly early in the circulation loop, i.e., in arteries, where oxygen is plentiful. 
This section discusses some strategies robots could use for limiting power. 
Since $10^{12}$ robots consuming oxygen as fast as possible leads to significant oxygen reduction, we focus on this case for evaluating the consequences of limiting power use.

A fixed limit on robot oxygen consumption could be implemented in hardware by reducing the number or capacity of pumps on the robot surface, or fuel cells inside the robot. This would prevent the robot from using all oxygen reaching its surface when the concentration is high, i.e., shortly after leaving the lungs. The more flexible and targeted power limitation methods discussed in this section require robots to alter pump capacity based on sensor measurements of quantities such as oxygen concentration, location in the body or distance to vessel walls. Moreover, software-based limits would allow some robots to use all the oxygen when occasions arise that could benefit from high burst power.

An alternative method for limiting oxygen collection in large vessels is for the robots to travel in groups, analogous to the grouping of blood cells in some blood vessels~\cite{freitas99}.
Such groups could form from robots waiting in moderate-sized vessels, such as lung veins, for a group to accumulate, or robots could search for others to form a group while moving with the blood.
If the group is large and compact enough that some robots are completely surrounded by others, the robots at the surface of the group could share collected oxygen with those inside, or the robots could occasionally change positions so each robot spends part of the time at the surface of the group.
When vessels become too small for the group, the robots could disperse into smaller groups or individual robots, adjusting the size to match that of the vessels~\cite{fan22}.

Grouping reduces absorption per robot by exploiting the competition for oxygen among nearby robots, which is the opposite of the improved absorption by smaller robots discussed in \sect{oxygen storage robots}.
For example, suppose $n$ robots join to form a spherical group of radius $R$. To accommodate the volume of these robots, $R \approx \rRobot \sqrt[3]{n}$. From  \eq{robot absorption}, the rate this group collects oxygen is proportional to $R$. Thus the oxygen collected per robot in the group is reduced from that of isolated robots by a factor of $(R/\rRobot)/n \approx n^{-2/3}$. 
For example, 100 $1\,\micron$-radius robots could form a group with $R\approx 5\,\micron$ and then each robot would receive about $5\%$ of the oxygen that isolated robots collect.
For this method of power reduction, the groups need not all have the same size nor contain all the robots in large vessels. For example, if among 200 robots, 100 form a group while the others remain isolated, the total oxygen consumed by those robots would be about half what they would consume without any grouping.

\begin{figure}
\centering 
\includegraphics[width=\figwidth]{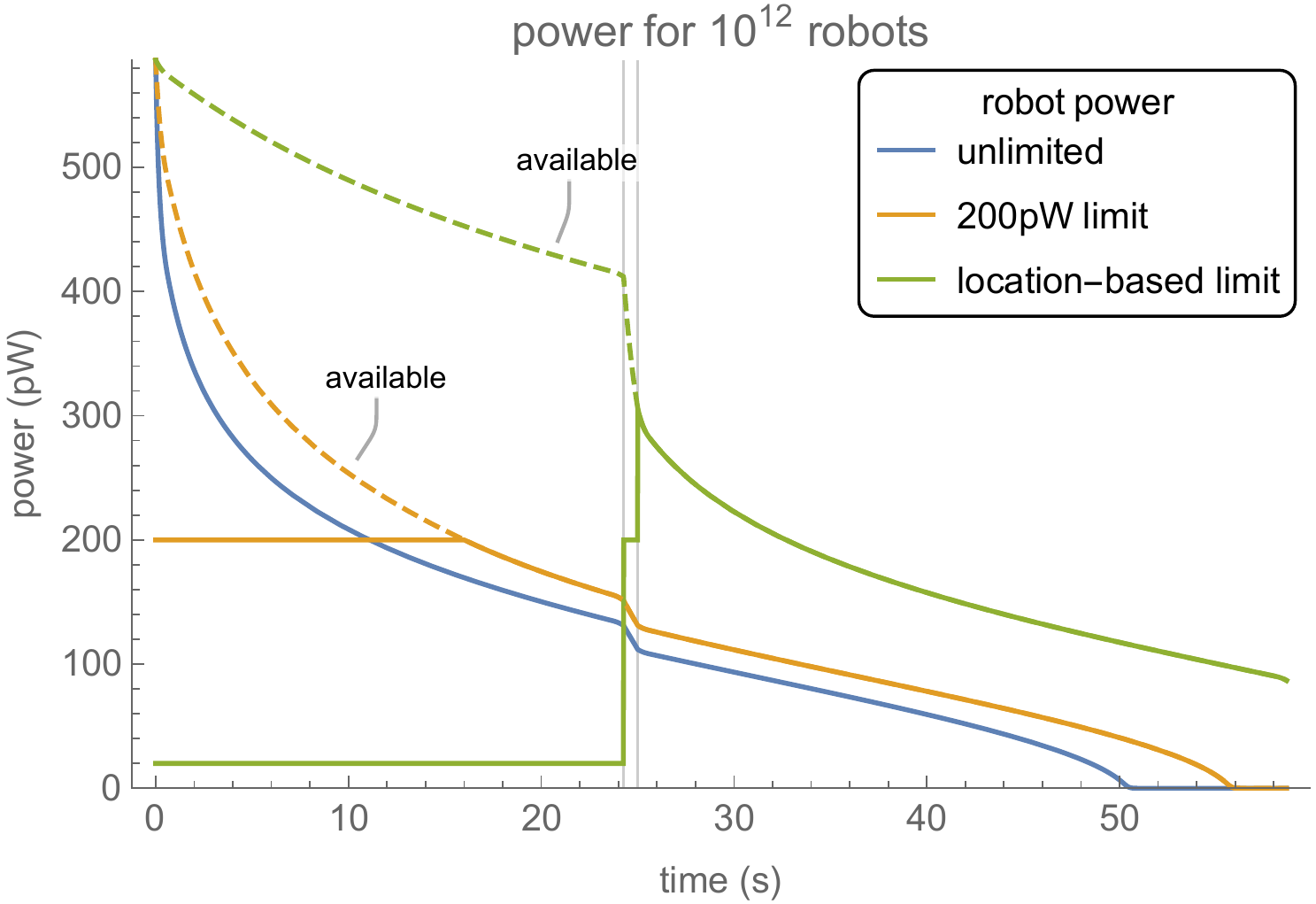} 
\caption{Power for $10^{12}$ robots during a circulation loop. The figure compares three cases: 1) unlimited power (the same as shown in \fig{power}), 2) robots limited to a maximum of $200\,\picowatt$, and 3) robots limited to $20\,\picowatt$ in arteries, $200\,\picowatt$ while in capillaries, and unlimited use in veins. For the two cases with limited power, the dashed curves show the power available to a robot if it consumed all oxygen reaching its surface (provided few, if any, other robots exceed their limits so oxygen is not significantly reduced by the robot switching to maximum power).
The light vertical lines indicate when the blood is in a capillary.}\figlabel{limited power}
\end{figure}

\begin{figure}
\centering 
\includegraphics[width=\figwidth]{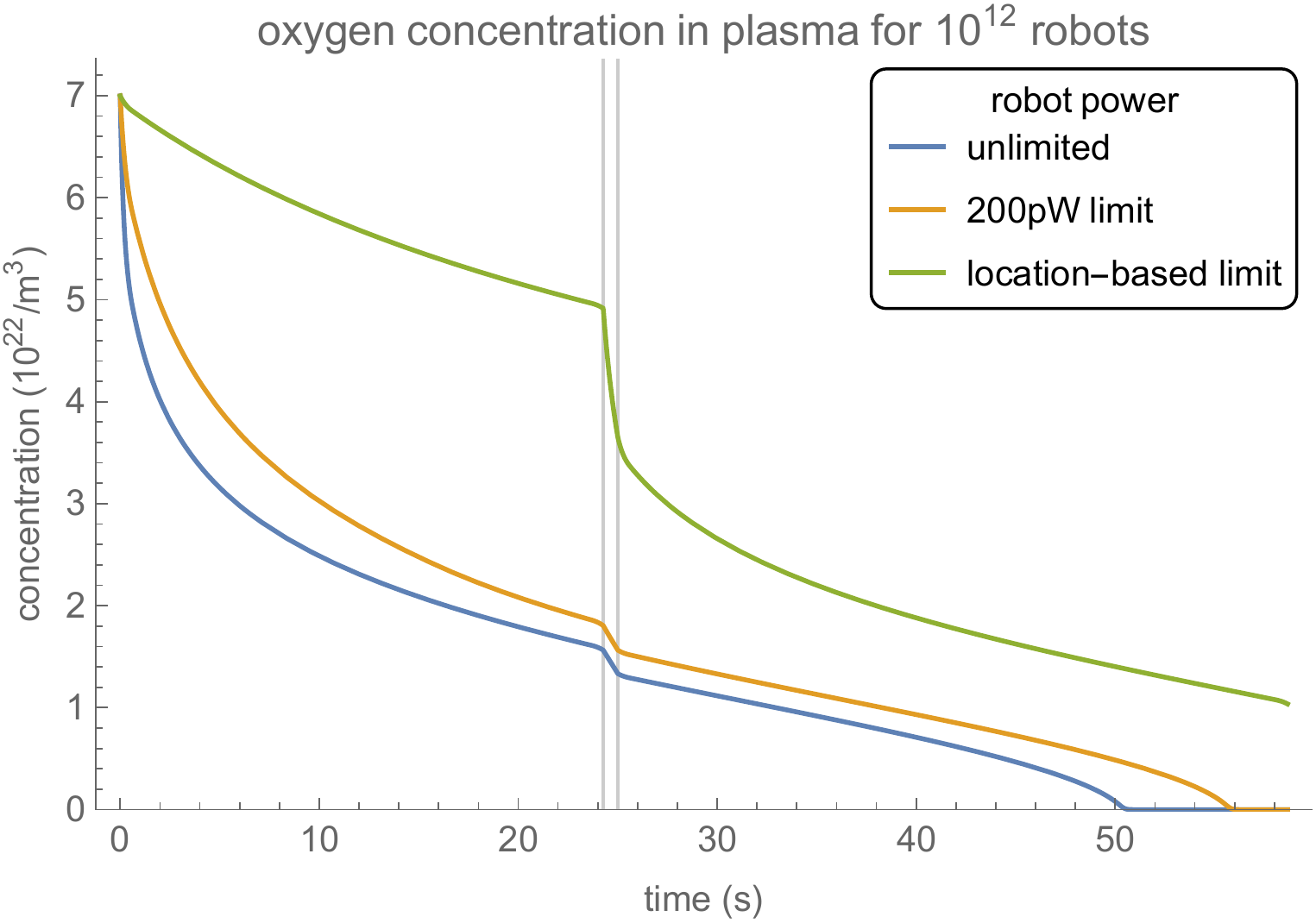} 
\caption{Oxygen concentration for $10^{12}$ robots. The figure compares the same three cases as shown in \fig{limited power}.
The light vertical lines indicate when the blood is in a capillary.}\figlabel{limited power oxygen}
\end{figure}

\subsubsection{Power Limits for All Robots}

The simplest approach to limiting robot oxygen consumption is for all of them to have a fixed maximum power generation rate.
 \fig{limited power} shows one example of this approach for $10^{12}$ robots where a $200\,\picowatt$ limit extends the range of a circuit where robots have power, but power still falls to zero before returning to the lung. To provide some power throughout the circulation loop with this many robots would require a somewhat lower power limit.
 
The dashed curves in \fig{limited power} indicate power available to a robot from the oxygen concentration in the plasma around that robot. This is the power a single robot would have if it switches from limited-power to using all available oxygen. If only a few robots make this choice, they will not significantly lower the oxygen. Thus the dashed curve indicates the power available to a few robots, e.g., for occasional burst activity, provided at most a tiny fraction of the robots make use of that additional power at any given time.
An example of such activity is the initial safety evaluation discussed in \sect{anemia}. Any robots detecting problems could use the high available power for burst communication of its measurements.

By limiting their power production when oxygen is plentiful, robots leave more oxygen for robots later in the circulation loop. In effect,  the robots use blood cells as external oxygen storage tanks to shift when and where robots utilize oxygen. This is conceptually similar to oxygen provided by additional transport robots discussed in \sect{oxygen storage} without the need for additional robots. On the other hand, specialized transport robots could deliver oxygen more effectively, especially if they use pumps for transfer rather than relying on diffusion.

Instead of a fixed power limit, robots could use a specified percentage of the oxygen reaching their surface. Unlike a fixed power limit, such as $200\,\picowatt$, a percentage limit would adjust to the decreasing oxygen concentration as robots move through a circulation loop. For example, $10^{12}$ robots using only $10\%$ of the oxygen would have the same effect on oxygen concentration as $10^{11}$ robots using all available power, as illustrated in \fig{oxygen}. The power available to each robot would be one-tenth that shown for $10^{11}$ robots in \fig{power}, i.e., the robots would have a few tens of picowatts. This would only be a worthwhile alternative to using $10^{11}$ robots if there is a benefit of having ten times as many robots, each with one-tenth the power. For instance, if the mission is to have at least one robot pass through every capillary to look for a target location, e.g., recognized by a rare pattern of chemicals, and this detection can be done with low power. In that case, a larger number of robots will complete the survey more rapidly than a smaller number of robots. Moreover, if the response to the detection requires much more power, that will be available to the few robots finding the target due to the higher oxygen concentration left by most robots using only a small fraction of the available power.

The above discussion of power limits supposes those limits are always in effect.
Another possibility is limits that apply occasionally so robots alternate between consuming much or all of the available oxygen and limiting their consumption. For example, robots could consume significant oxygen only every second or third circuit. If such duty cycling occurs independently among robots, it would reduce the number of active robots by the corresponding factor, i.e., 2 or 3 in this example. On the other hand, if robots synchronize their schedules, oxygen in the circulation would alternate between high and low levels. This could be beneficial if the harm from continuous moderately low oxygen is worse than switching between very low and normal concentrations.

\subsubsection{Power Limits Based on Location within a Circulation Loop}

A more sophisticated limitation method is for the limits to depend on the robot's location rather than using a single overall limit.
Such a strategy would be particularly useful if the main operation of a robot occurs when it is in a capillary, e.g., measuring properties of nearby tissues. In this case, robots could reduce power use while in arteries, to have more available for the short time they spend in capillaries. They could also defer some power consumption, e.g., analyzing measurements they collected in the capillary, until they reach vessels with more available oxygen. This could be in veins, where additional reduction in oxygen saturation of red cells would no longer affect oxygen available to tissues. 
This strategy would require some adjustment for portal flows, e.g., in the portal vein, where blood flows through another capillary, in the liver, before returning to the lungs. In that case, the robot could wait until it has passed a liver capillary before increasing its power use.

\fig{limited power} shows an example for $10^{12}$ robots using this method. With the parameters illustrated here, this approach provides power throughout the circulation loop. In particular, it provides higher power when robots are in the veins compared to robots either using all available power or limiting their use to at most $200\,\picowatt$.
\fig{limited power oxygen} shows the effect of this limit on oxygen concentration.

Implementing location-dependent limits on power requires that robots determine the type of vessel they are in. 
Robots could do so in a variety of ways, with trade-offs between complexity for robot processing and accuracy.
One approach is to use an onboard clock to measure the time since a robot left the high-oxygen environment of a lung capillary. They could use a fixed time, e.g., 30 seconds, to decide when they have likely reached a vein and can increase power use. This approach will not adjust for variation in circulation speed due to changes in heart rate, nor variations in circulation path lengths, but is a simple approach that avoids the need for robots to determine when they have passed through a capillary.

Alternatively, robots could measure oxygen concentration in the surrounding plasma. This is high in arteries, decreases as the robot passes through a capillary where tissue consumes oxygen, and is relatively low in veins. Concentration thresholds required to distinguish these types of vessels depends on the concentration of robots (see \fig{oxygen}a) and the power-consumption method of the robots (see \fig{limited power oxygen}). This variation could be predefined based on these choices. More challenging for estimating location from oxygen concentration is the variation in tissue use in different organs and at different times, and variation in hematocrit in small vessels. These variations limit the accuracy of using oxygen concentration to determine the type of vessel the robot is in.

Combining a variety of measures can identify vessel type more accurately. Most circuits through the body move through arteries of decreasing size, through a capillary, and then through veins of increasing size. Changes in pressure and how much it varies over the duration of a heartbeat distinguishes arteries from veins~\cite{freitas99}. For small vessels, changes in fluid flow near the robot allow estimating vessel size~\cite{hogg18}.

\subsubsection{Power Limits Based on Location in the Body}

In addition to adjusting power based on the type of vessel a robot is in, it could adjust its power based on its macroscopic location in the body. For example, robots could use information on which organ they are in~\cite{freitas99} to set their power limits. This would allow robots to adjust power generation to match organ-specific tasks. As an extreme case, if robots only need to be active in one organ, then 
a number of robots large enough to deplete oxygen if they all use power would instead consume much less oxygen and mainly affect that organ and tissue downstream from that organ. This would avoid depleting oxygen in veins throughout the body, but could lead to significant local oxygen reductions, particularly in the small veins leaving those organs before the blood reaches larger veins, where it mixes with blood from other organs. 

As seen in \fig{oxygen}, the most severe reduction in oxygen with $10^{12}$ robots occurs near the end of the circuit, i.e., after blood has mixed into large veins. For robots limiting power use to one or a few organs, that extreme reduction would not occur: instead, blood from other organs would partially restore oxygen in the vein. Thus organ-specific power use could tolerate larger numbers of robots than if all robots consume oxygen. A trade-off in this scenario is that only the portion of robots in the target organ at any given time would be actively performing their tasks while the majority of robots simply move passively with the blood. Alternatively, robots with locomotion could target the organs of interest, thereby providing sufficient concentration to perform their mission with a smaller total number of robots.

Another application for power limits based on location within the body is to adjust to organ-specific variations in tissue demand, beyond those compensated for by variation in tissue capillary density or changes in blood flow in response to those variations. That is, robots may need to reduce their oxygen consumption in tissue with higher than usual oxygen demand. At a local scale, robots could determine such limits from measuring oxygen concentration. However, by the time a robot reaches the tissue, and encounters the lowered oxygen concentration, it will be too late for limiting power in small arteries leading to that tissue and where oxygen concentration may still be relatively high. In this case, a robot could limit power more effectively using information on which organ it is entering and the overall tissue demand of that organ.

\subsubsection{Power Limits Based on Robot History}

Instead of a power limit on all robots, the limit could be selective by depending on the recent history of each robot.
The benefit of a history-dependent limit depends on the fraction of robots requiring significant power at any given time, as determined by their current configuration and their local environment. For example, if only 10\% of robots need significant power, then the effective number consuming oxygen is reduced by that fraction. If this reduction is, at least roughly, uniform throughout the body, the robot power will affect oxygen according to the effective number of active robots. 
For instance, if only 10\% of a trillion robots generate significant power and do so at their maximum possible rate, the effect on oxygen concentration will correspond to $10^{11}$ robots consuming oxygen as rapidly as they can.

One situation leading to history-dependence arises in robots with oxygen storage and considerable variation in power requirements, depending on where they travel in the body. Such a robot could limit its oxygen intake and supplement that intake with oxygen from its storage tank. When the tank is nearly empty, the robot could absorb more rapidly to fill it. This would be useful if neighboring robots share data and only one of them needs to expend significant power to process or communicate the data: the robot using burst power would deplete its tank and need to refill it.

Another case of history-dependent power is if robots only need high power when handling rare events. This could occur when a robot detects a specific chemical in the blood and needs to move toward its source~\cite{hogg06a}. Or if such a robot requires confirmation from other nearby robots before taking action, to reduce the number of false positives. In this situation, the detecting robot could increase its power use to communicate with nearby robots and evaluate their responses. In this case, the originating robot may need sufficient power to send a significant number of bits, e.g., a summary of the data it collected, whereas responding robots could just send a short response indicating whether their information is consistent with that from the detecting robot.

\begin{figure}
\centering 
\includegraphics[width=\figwidth]{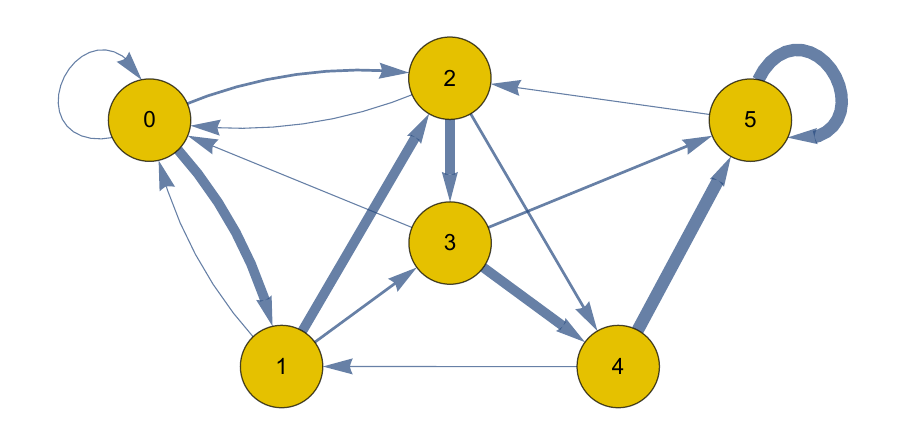}  
\caption{Markov model of amount of data stored by a robot with capacity to store data from $5$ capillaries and ability to transmit the data from up to $3$ capillaries each time it reaches the skin. The number in each node is the number of capillary measurements  the robot has currently stored. The thickness of the edges correspond to the transition probabilities. For this illustration, robots do not collect data from the lung or the skin.}\figlabel{Markov data collection}
\end{figure}

As a quantitative example of history-dependent power, consider robots that measure chemical concentrations in capillaries and communicate their readings when they are in range of external receivers on the skin. For example, receivers could emit a beacon signal that robots can detect when they are nearby.
Suppose the relevant data for this mission occurs in tissue other than the lung or skin. 
In this case, the relevant property of each circulation loop is whether the robot goes to the skin, and, if not, whether it flows through a portal system, thereby measuring two capillaries before returning to the lung for another loop through the circulation.
With typical resting perfusion rates~\cite{feher17}, 
$8\%$ of the blood flows to the skin and $20\%$ flows through the portal vein, which is the major circulation path passing through two capillaries. The rest of the blood flows through a single capillary that is not in the skin before returning to the heart.
Suppose a robot can store measurements from up to $5$ capillaries, and has time to transmit the data from up to $3$ measurements while near the skin.
 
Due to the mixing of blood in the heart, where a robot goes in the body during each circulation loop is independent of where it went previously. 
This leads to a Markov stochastic process for the amount of data stored by a robot with the transition graph shown in \fig{Markov data collection}. Each edge in the graph corresponds to the robot making a single circulation through the body. For instance, a robot with empty data storage is most likely to store data from one capillary during its next circulation. A robot could also go to the skin, in which case it does not collect data, or through a portal system, in which case it collects data from two capillaries. 

After multiple loops, the Markov process approaches its stationary distribution, in which about $70\%$ of the robots have filled data storage. Such robots are unable to collect any additional data until they have had an opportunity to transmit, and delete, some of their collected data during their next passage near the skin.
Moreover, most of the robots reaching the skin would have data on at least 3 capillaries, and hence have enough data to transmit at their maximum rate while near the skin.

Suppose the robots mainly require power during data collection. The $70\%$ of robots with full data storage then use only a minimal amount of power. In terms of power use, $10^{12}$ robots would be comparable to $3\times 10^{11}$ robots consuming oxygen as fast as they can. This would be similar to an overall limit of robots using $30\%$ of available power, but with history dependence so the power use is targeted to those robots that most need it, rather than a reduction for all robots.

\subsubsection{Limiting Oxygen Consumption Near Vessel Walls}

\begin{figure}
\centering 
\includegraphics[width=\figwidth]{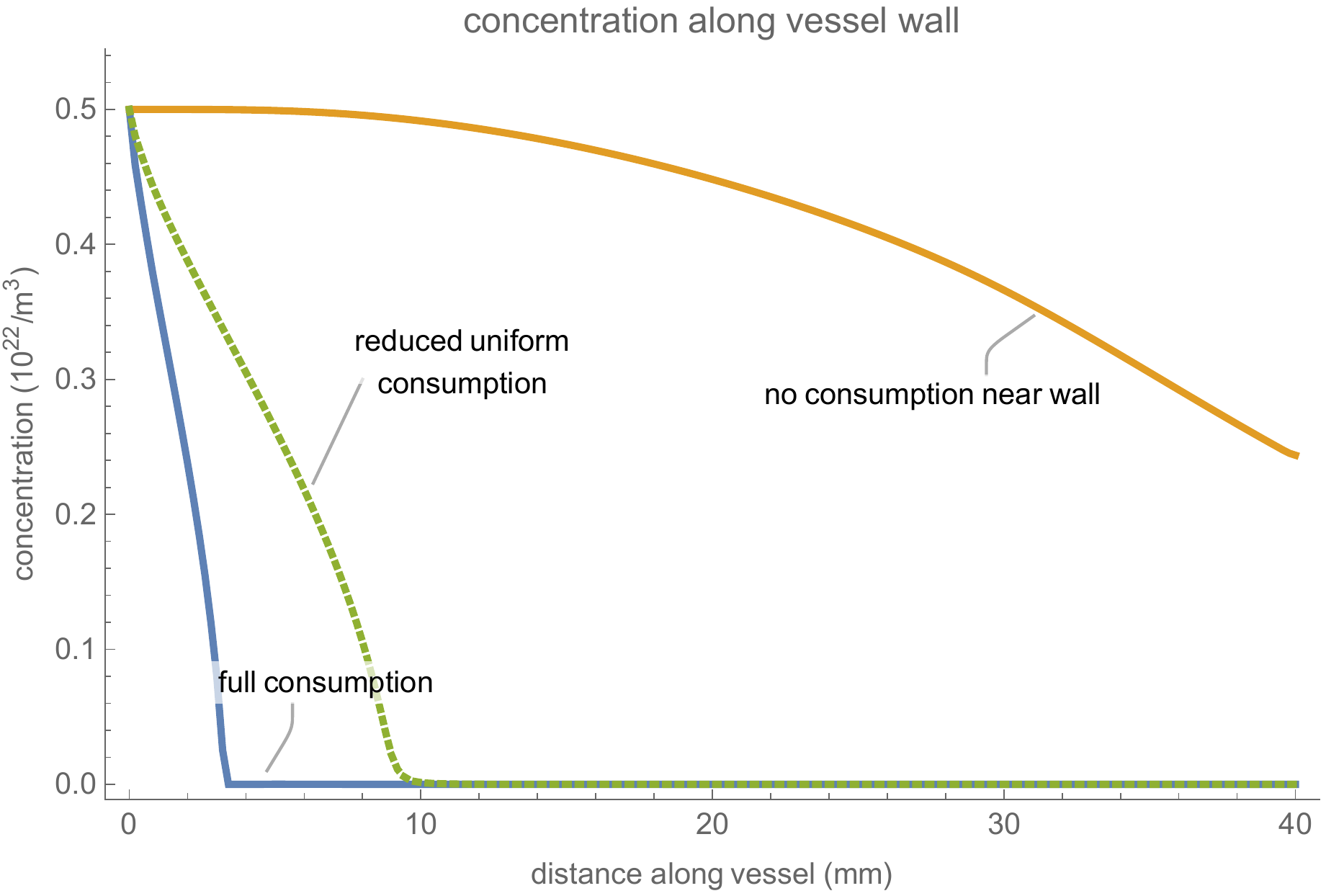}  
\caption{Oxygen concentration along the wall of a straight vessel of $2\,\millimeter$ diameter with $10^{12}$ robots employing different oxygen consumption methods: full consumption of all oxygen reaching the robot surface, no consumption by robots within $0.3\,\millimeter$ of the vessel wall, and all robots limiting their consumption to $50\%$ of available oxygen.}\figlabel{concentration along wall}
\end{figure}

The above location-based limits considered adjusting power consumption based on a robot's position in the circuit, corresponding to the time since the robot was last in the lung, and based on the type of tissue or organ the robot is in.
These approaches can increase the amount of oxygen remaining in the blood, particularly toward the end of the circulation. 

As described in \sect{typical circulation}, low oxygen concentrations in veins could affect cells on the vessel wall. Addressing this issue suggests another location-based limit, namely for robot power limits to apply only when a robot is close to the vessel wall. This could be useful in vessels whose diameter is at least a millimeter or so since, for vessels of that size, diffusion limits oxygen transport to a relatively small fraction of the vessel diameter during the time a robot passes through the vessel. Specifically, \fig{oxygen} shows the low oxygen concentration with $10^{12}$ robots occurs during the last 10 to 20 seconds of the circulation loop. The characteristic diffusion distance of oxygen during $t=20\,\second$ is $\sqrt{2\Doxygen t} = 0.3\,\millimeter$, with \Doxygen\ given in \tbl{parameters}.
Thus robots could avoid creating low concentration near vein walls by reducing their oxygen consumption when they are near the vessel wall, e.g., as determined by measuring fluid stresses on their surfaces~\cite{hogg18}. Alternatively, if robots have locomotion capability, they could move away from the vessel wall. In this case, the low oxygen concentration seen in \fig{oxygen} would occur in the central portion of the vessel only.

The effectiveness of limiting oxygen consumption based on distance to the vessel wall depends on how much mixing occurs during transport in moderate-sized veins, including the effect of merging vessels and cell motion.
For veins whose diameter is substantially larger than the size of cells, this could be evaluated by approximating the blood as a uniform fluid. For smaller vessels, this evaluation requires simulations including deformable cells, e.g., in vessels with diameters up to  hundred microns or so~\cite{bagchi07}.
The main advantage of limiting power use near vessel walls arises in vessels large enough that oxygen does not have time to diffuse across the vessel, and there is not significant mixing from the flow. In general, mixing is slow in laminar flow~\cite{squires05} found in such vessels.

As a quantitative example of this mitigation method, consider the flow in a $40\,\millimeter$-long portion of a vein with diameter of $2\,\millimeter$. Suppose oxygen is well-mixed through the blood as it enters this vessel segment, has a relatively low concentration of $0.5\times10^{22}/\meter^3$ and there are $10^{12}$ robots in the circulation. 
This example corresponds to the situation at about $45\,\second$ in \fig{oxygen}, and oxygen is fully depleted after a few more seconds of full consumption by the robots. 

The model developed here for robot power over a full circulation loop treats the flow as one-dimensional and gives the average concentration over the vessel cross section. 
For a more detailed view of oxygen concentration, this example evaluates oxygen transport in the vessel segment with parabolic, i.e., Poiseuille, flow profile and with average speed of $2.5\,\millimeter/\second$. Oxygen transport is a combination of convection with that flow, diffusion, consumption by robots and replenishment by red cells, as described in \sectA{oxygen changes}. This vessel is a vein, so there is no consumption by tissue.
In this case, the mixing due to the motion of blood cells is a minor addition to the diffusion of small molecules, such as oxygen, in the blood~\cite{freitas99}.
From \eq{Peclet} the Peclet number for this oxygen transport is $\Pec = 2500$, so oxygen mainly flows along the vessel with relatively little diffusion across it.
This behavior leads to considerable variation in oxygen concentration across the vessel: robots in the slowly moving blood near the walls deplete oxygen much more than robots traveling with the faster flow near the center of the vessel.

Robots could mitigate the rapid concentration decrease near the wall by reducing their oxygen consumption. As an example, \fig{concentration along wall} shows how the concentration near the vessel wall changes in three cases. In the first case, robots fully consume oxygen reaching their surface. In the second case, robots within $0.3\,\millimeter$ of the vessel wall do not consume oxygen, with that distance corresponding to oxygen's characteristic diffusion distance discussed above. This distance to the vessel wall accounts for about half the total cross sectional area of the vessel. Thus in this case, only about half the robots are consuming oxygen, i.e., those in the central portion of the vessel. This leads to the third case shown in \fig{concentration along wall} where only half the robots consume oxygen, or, equivalently, each robot consumes only half the oxygen reaching its surface, without regard to their position in the vessel. As seen in \fig{concentration along wall}, this overall reduction in power reduces the rate of oxygen depletion near the wall, but is much less effective than when robots limit power based on their distance to the wall.

\begin{figure}
\centering 
\includegraphics[width=\figwidth]{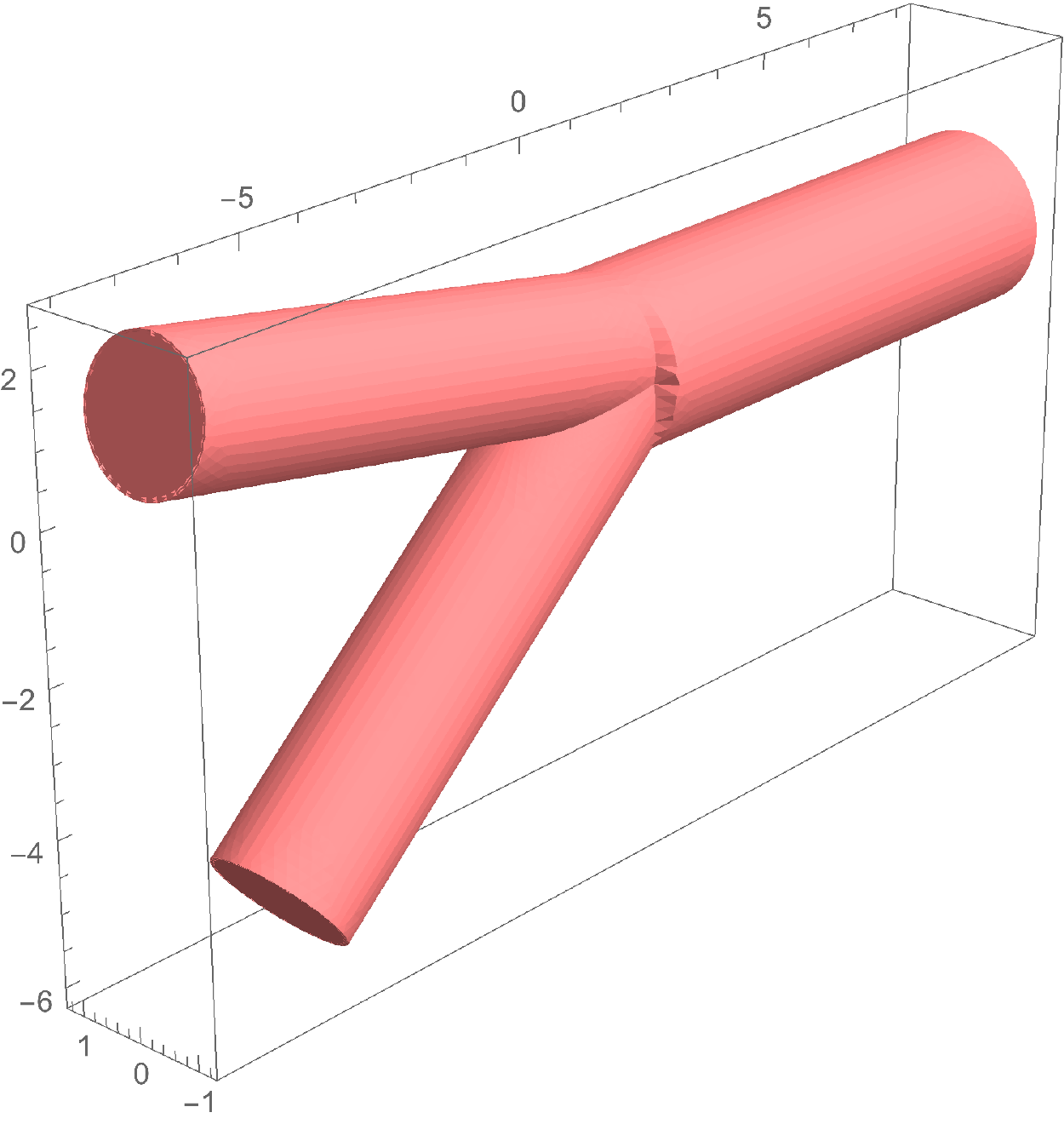} 
\caption{Example of merging vessels: the branches, each with diameter of $2\,\millimeter$, merge into a vessel with diameter of $2.5\,\millimeter$.
Numbers along the axes are positions in millimeters measured from an origin at the center of the merging vessels.}\figlabel{merging vessels}
\end{figure}

\begin{figure}
\centering 
\includegraphics[width=\figwidth]{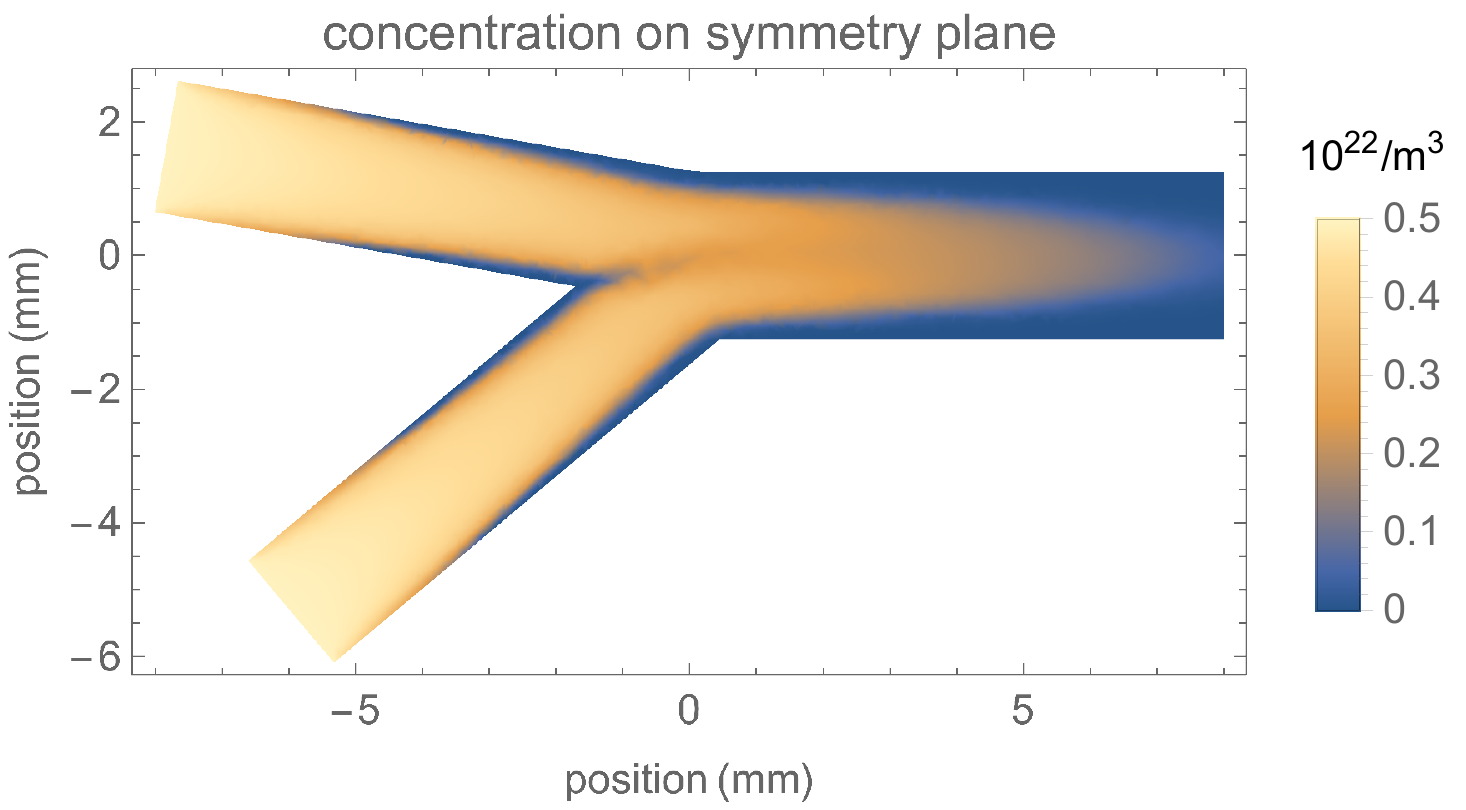} 
\caption{Oxygen concentration on a cross section through the merging vessels on the symmetry plane with $10^{12}$ robots in the bloodstream, each consuming all oxygen reaching its surface.}\figlabel{merging vessels concentration}
\end{figure}

\begin{figure}
\centering 
\includegraphics[width=\figwidth]{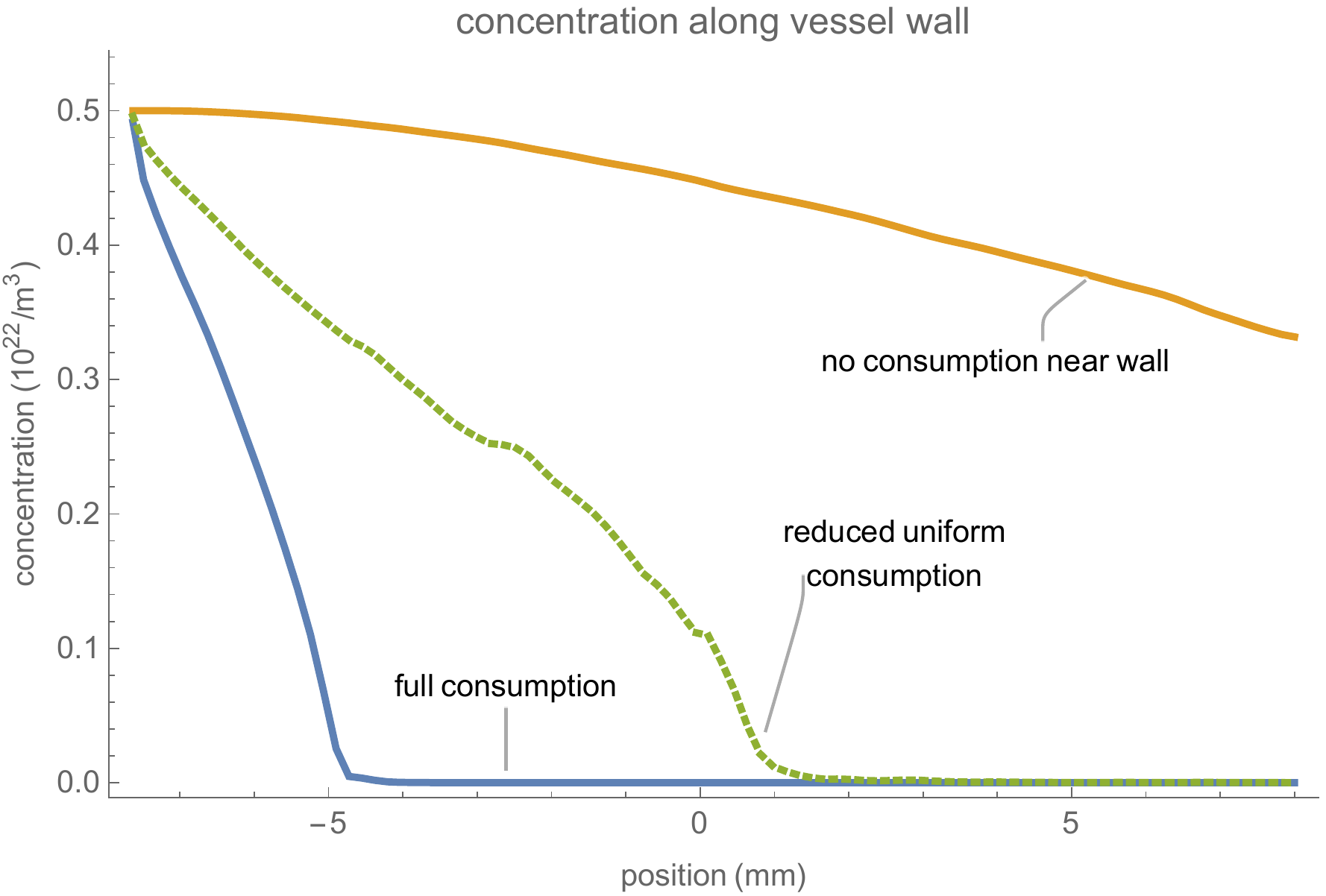} 
\caption{Oxygen concentration along the upper wall of  the merging vessels with $10^{12}$ robots employing different oxygen consumption methods: full consumption of all oxygen reaching the robot surface, no consumption by robots within $0.3\,\millimeter$ of the vessel wall, and reduced consumption corresponding to the fraction of robots at least $0.3\,\millimeter$ from the vessel wall, which is about 50\% in these vessels.}\figlabel{merging vessels concentration along wall}
\end{figure}

\begin{figure}
\centering 
\includegraphics[width=\widefigwidth]{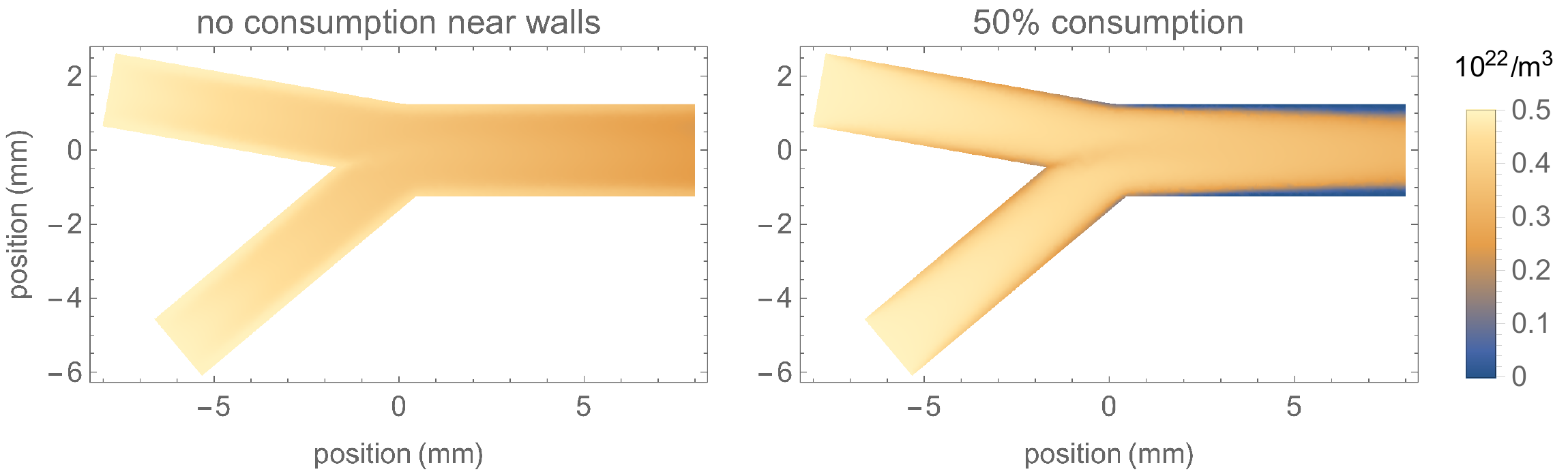} 
\caption{Oxygen concentration on a cross section through the merging vessels on the symmetry plane with $10^{12}$ robots in the bloodstream. Left: no consumption by robots within $0.3\,\millimeter$ of the vessel wall. Right: robots consume 50\% of oxygen reaching their surfaces.}\figlabel{merging vessels concentration 2}
\end{figure}

The example of \fig{concentration along wall} is for flow in a single straight vessel. The flow from the merging of veins of various sizes could somewhat increase the mixing and thereby reduce the benefit provided by robots limiting consumption near the wall. To evaluate this effect, \fig{merging vessels} is an example of merging vessels with asymmetric branching. As is typical of merging blood vessels, the total cross section of the two branches exceeds that of the main vessel, so flow speed in the branches is somewhat slower than in the main vessel.
To focus on the effect of the merging vessels, the example includes only $8\,\millimeter$ in each branch and the same distance in the main vessel. This is considerably shorter than the $40\,\millimeter$-long straight vessel discussed with \fig{concentration along wall}.

Accounting for the effect of the branches requires solving for the fluid flow through the vessels. As with the previous example, we suppose parabolic flow exits the main vessel with average speed $2.5\,\millimeter/\second$ and the inlet pressure at both branches is the same. The fluid flow with these boundary conditions is laminar.  \fig{merging vessels concentration} shows the resulting concentration distribution on a cross section through the vessels along their symmetry plane. The merging flows provides some mixing where the branches join. Nevertheless, the behavior of the concentration is similar to that seen in the straight-vessel example. Specifically, \fig{merging vessels concentration along wall} shows the oxygen concentration along the upper wall of the upper branch and main vessels for the same three cases as shown for the straight vessel in \fig{concentration along wall}.
\fig{merging vessels concentration 2} compares the concentration across the vessels for the two cases where oxygen consumption is limited.

Even when robots near the vessel wall do not consume oxygen, the concentration near the wall eventually gets very small. Thus, in the context of these examples, the main benefit of this mitigation is extending the range of the circulation loop by a few centimeters compared to when robots consume all oxygen or limit consumption to $50\%$. This increase in range could be especially beneficial in a vein just before it merges with other veins that contain blood from shorter circuits and hence have more remaining oxygen than the model estimates for an average circulation loop in \fig{oxygen}. In that case, avoiding fully depleted oxygen for a few additional centimeters could be sufficient to avoid extremely low concentrations near the walls of any veins, without requiring all robots in those vessels to reduce their power generation. Alternatively, achieving the same increase in range with a power limit on all robots would require a much larger reduction in power than the $50\%$ reduction that corresponds to robots near the wall consuming no oxygen while the rest of the robots consume at their maximum rate. 

The one-dimensional flow model used here assumes oxygen and cells are well mixed over the vessel cross section over the distance and time scales of interest here. In small vessels, diffusion is fast enough to provide this mixing. In larger vessels, mixing depends on the nature of the flow, including enhanced mixing due to motion of the cells moving in the blood, changing vessel diameters and mixing flows from branches.
Those fluid behaviors are not sufficient to fully mix oxygen in millimeter-sized vessels. Thus another consequence of limiting consumption near walls is increasing the uniformity of the concentration across the vessel. In this case, the robot behavior extends the accuracy of the model's assumption of uniform mixing to millimeter length scales. This is an example of robots adjusting their behavior to improve the accuracy of a simple global model of their behavior. This simplification arising from robot behavior can be useful when applying such models to evaluate global behaviors and select among different control protocols for the robots.

\subsection{Positioning Robots in Capillaries}

The small volume of capillaries leads to relatively large variation in robot numbers as described in \sect{capillaries}. Those capillaries that contain significantly more robots than average will have less power. Robots could reduce this variation by moving away from particularly close neighbors, including moving to other capillaries that have fewer robots.

Within a single vessel, robots in close proximity compete for oxygen. Such robots could improve oxygen distribution using a small-scale version of the power limiting strategies discussed in \sect{limiting power}. 
In addition, the flow of oxygen to robots depends on their positions along the vessel wall, as shown by the streamlines in \fig{absorbing robots}. This position dependence provides an additional mitigating strategy for small groups of nearby robots: improving power distribution by deliberately adjusting their positions relative to their neighbors.

\fig{absorbing robots} suggests there are competing effects on the oxygen delivered to robots. On the one hand, a robot directly upstream of another absorbs much of the oxygen that would otherwise go to the downstream robot. However, placing all robots on one side of the vessel allows more oxygen to reach downstream on the other side of the vessel, which can then diffuse across the vessel to downstream robots. The relative importance of these effects depends on the ratio of convective to diffusive transport of the oxygen, i.e., the Peclet number of the oxygen transport~\cite{squires05}.

\begin{figure}
\centering 
\includegraphics[width=\mfigwidth]{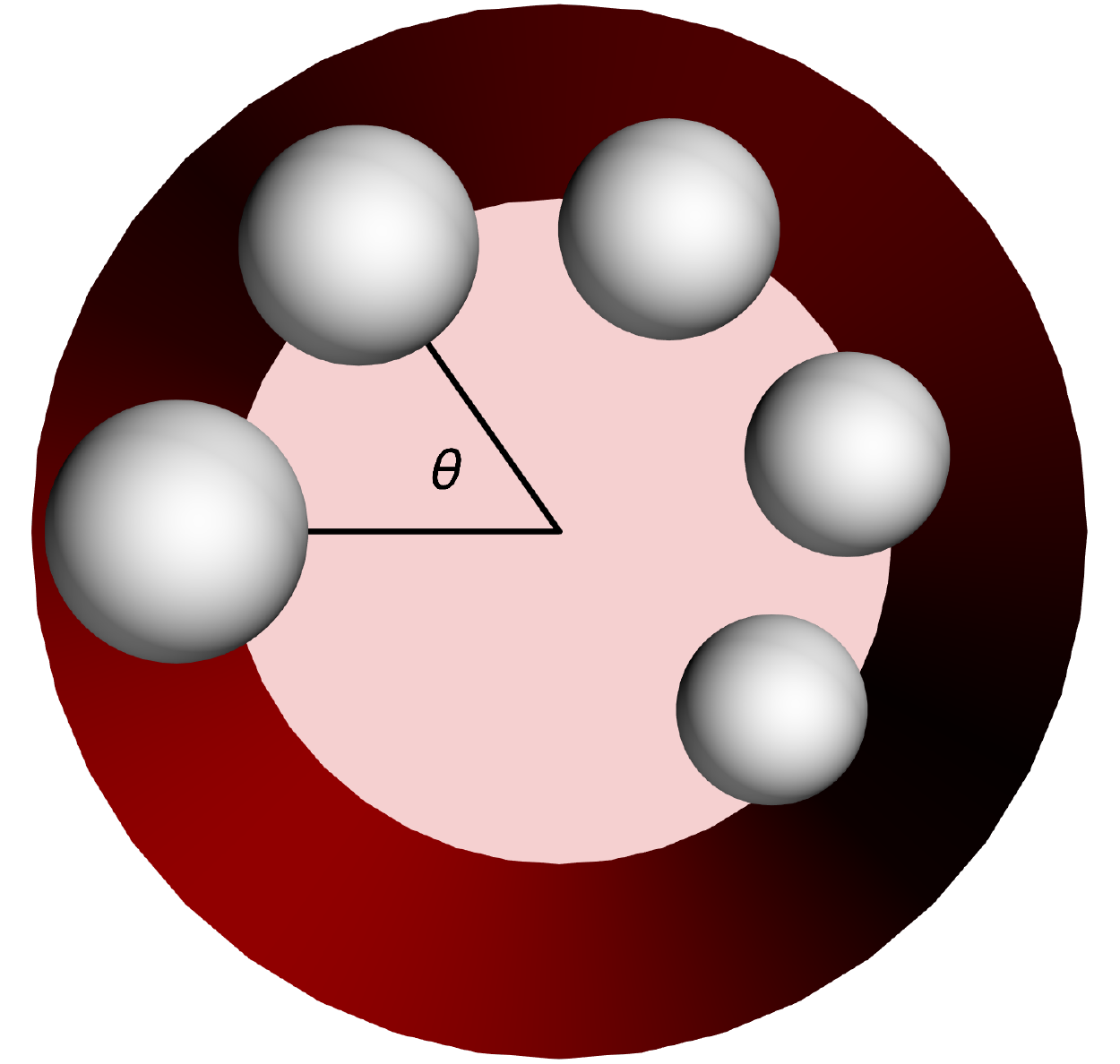} 
\caption{Example positioning of 5 robots in a vessel with $8\,\micron$ diameter viewed from the vessel inlet. Successive robots are offset around the wall by the angle $\theta$ and spaced along the vessel by $5.5\,\micron$. The arrangements shown in the top and bottom of \fig{robots in small vessel} correspond to $\theta=180^\circ$ and $0^\circ$, respectively.}\figlabel{5 robots in vessel}
\end{figure}

\begin{figure}
\centering 
\includegraphics[width=\figwidth]{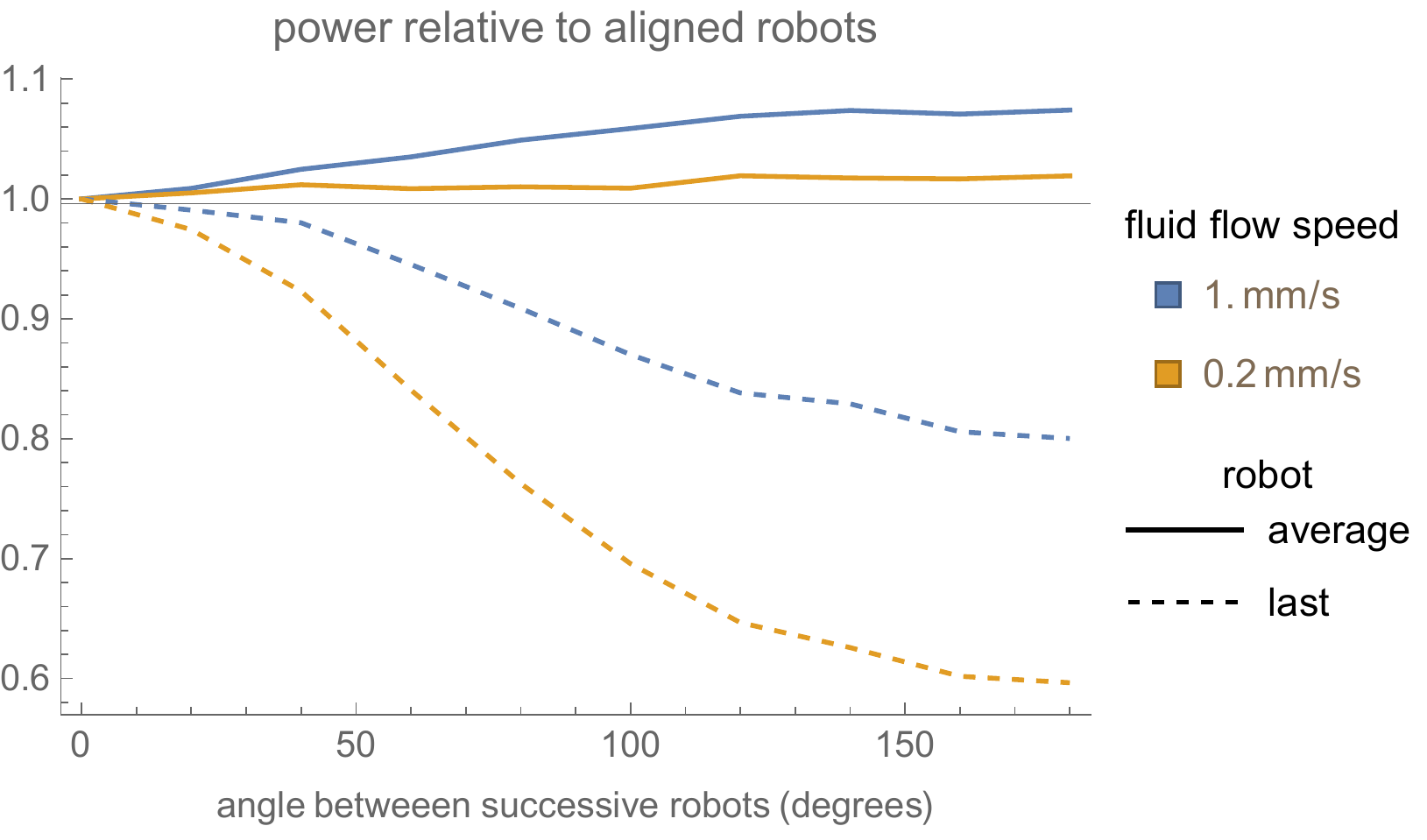} 
\caption{Relative power for 5 robots in a vessel as a function of angle $\theta$ with the geometry of \fig{5 robots in vessel}. Each curve shows the power relative to the situation with all robots aligned along the vessel wall, i.e., with $\theta=0$. The solid curves show the average power for the 5 robots with average fluid speed $1\,\millimeter/s$ and $0.2\,\millimeter/s$ for the upper and lower curves, respectively. The dashed curves show the power available to the fifth robot, i.e., the robot farthest downstream in this group, for these fluid speeds.
}\figlabel{relative power for 5 robots in vessel}
\end{figure}

To illustrate the potential of position adjustment, consider 5 robots positioned along a vessel wall with neighbors offset by an angle $\theta$, as shown in \fig{5 robots in vessel}. This arrangement is analogous to the angular spacing of leaves around the stem of a plant, where angles related to the golden ratio can minimize the extent to which higher leaves cast shadows on lower ones~\cite{strauss19}. However, unlike the direct path of light, diffusion allows some oxygen to reach robots that are directly downstream of others on the vessel wall.

\fig{relative power for 5 robots in vessel} shows the offset angle has different effects on the average power for these robots than it does for the last robot, which receives the least oxygen.
Specifically, offsetting neighboring robots by $180^\circ$ increases the average power by a few percent, with larger effect for faster moving fluid. This corresponds to the top arrangement in \fig{robots in small vessel}.
On the other hand, positioning all robots on the same side of the vessel (i.e., $\theta = 0$) increases power for the last robot even though it is directly downstream of all the other robots in the group. This occurs because robots on one side of the vessel allow more oxygen to reach downstream along the other side of the vessel and then diffuse to the last robot. As seen with the streamlines in \fig{absorbing robots}, this effect is larger for slower moving fluid. 
The robots could select among these options based on the relative importance of maximizing power for the group as a whole compared to ensuring all robots have at least a minimum amount of power, and depending on the size of the vessel and the speed of the fluid.

\begin{figure}
\centering 
\includegraphics[width=\textwidth]{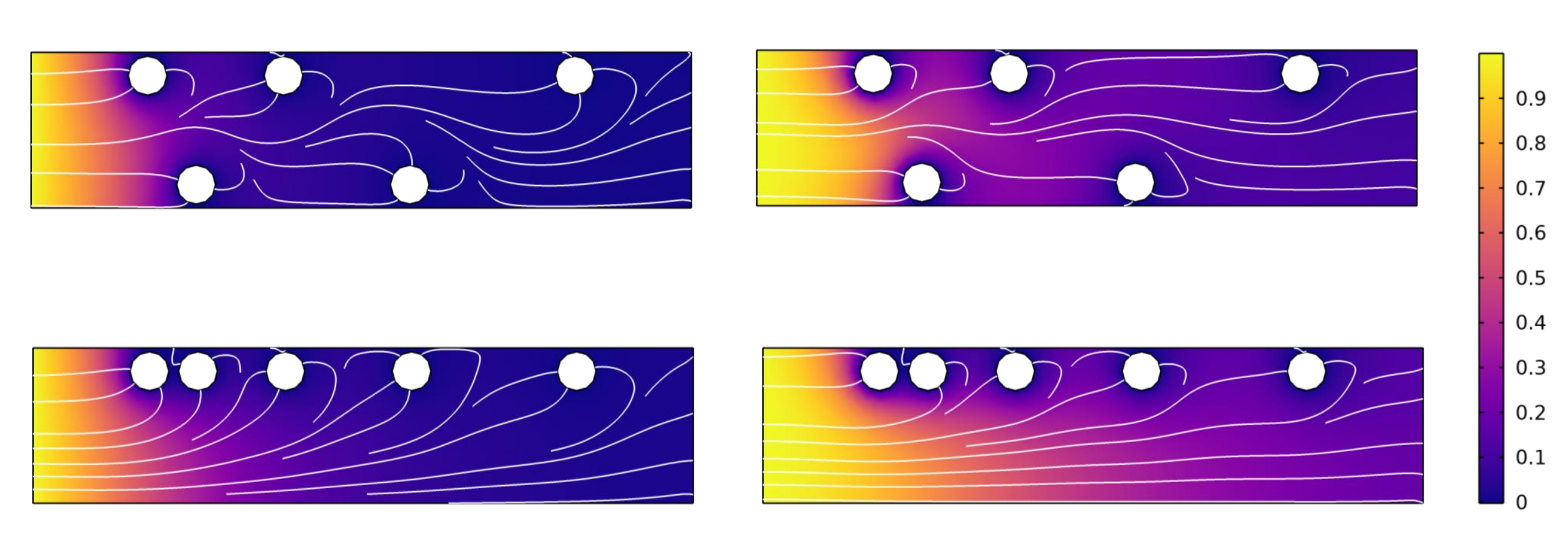}	
\caption{Relative oxygen concentration for nonuniformly-spaced robots on a vertical slice through the center of the vessel and the robots. The white curves are streamlines of the oxygen flux. The rows correspond to the robot positions shown in \fig{robots in small vessel}. The fluid flows from left to right in each diagram, with average speed $0.2\,\millimeter/\second$ and $1\,\millimeter/\second$ in the left and right columns, respectively.
}\figlabel{nonuniform robots}
\end{figure}

Robots could use nonuniform positions along the vessel to provide more power to downstream robots by having upstream robots closer to each other than downstream ones. As an example, in addition to the angular separation around the vessel discussed above, suppose the robot positions along the direction of the vessel increase quadratically, starting with a difference of $2.5\rRobot$ for the first two robots and occupying the same total distance along the vessel as the uniformly spaced robots discussed above. In the case of zero offset angle between neighbors (i.e., $\theta=0$), the distance between facing surfaces of the first two robots is $0.5\rRobot$.
\fig{nonuniform robots} shows the oxygen concentration for nonuniformly spaced robots when successive robots are on opposite sides and the same side of the vessel. The larger distance between the last robot and the others allows the last robot to collect oxygen from a larger portion of the fluid than when robots are uniformly spaced (\fig{absorbing robots}).

\begin{figure}
\centering 
\includegraphics[width=\figwidth]{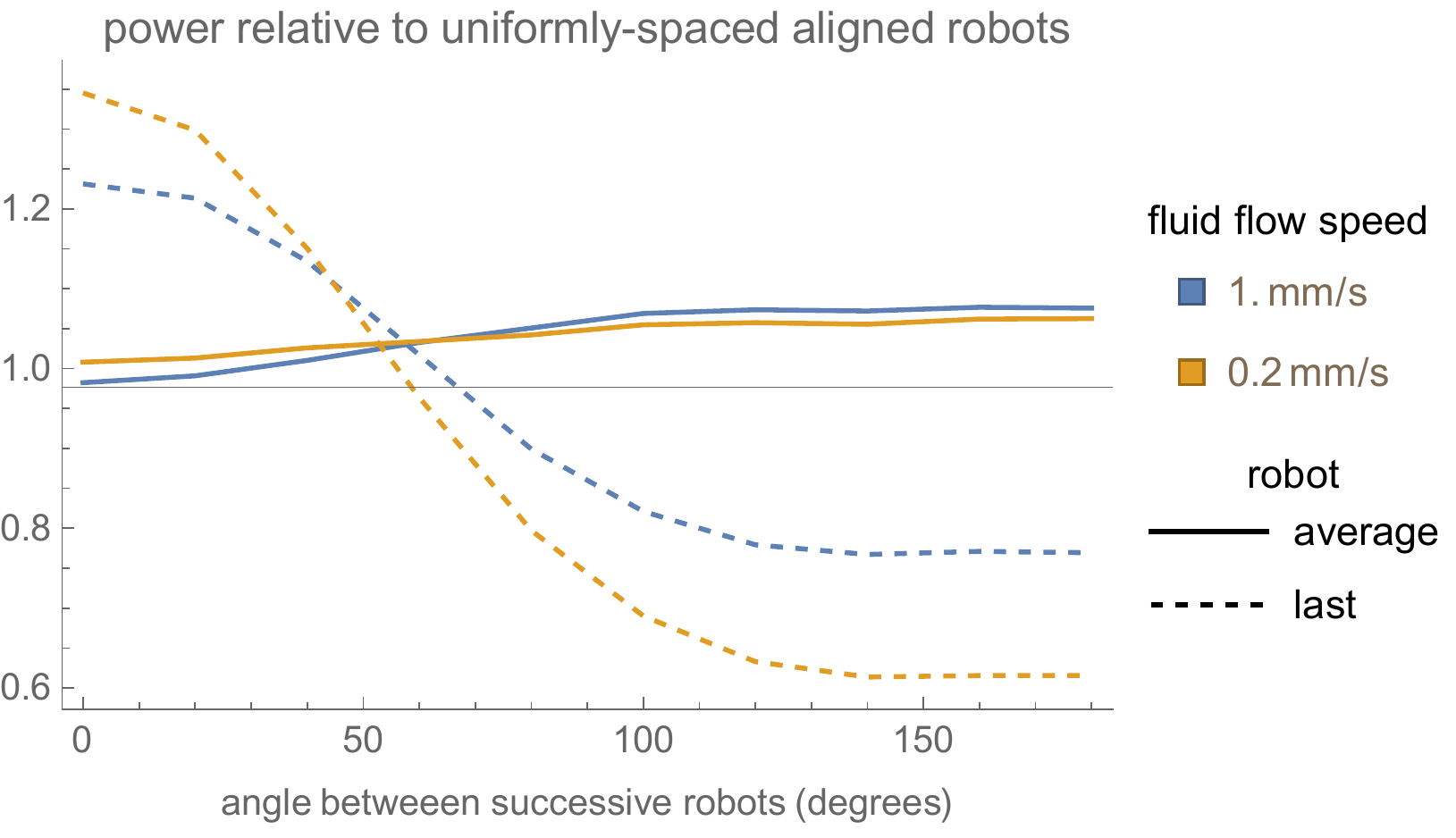} 
\caption{Relative power for 5 robots in a vessel as a function of angle $\theta$ with nonuniform spacing along the vessel. Each curve shows the power relative to the situation of uniformly-spaced robots aligned along the vessel wall, i.e., the same normalization used in \fig{relative power for 5 robots in vessel}. The solid curves show the average power for the 5 robots with average fluid speed $1\,\millimeter/s$ and $0.2\,\millimeter/s$ for the upper and lower curves, respectively. The dashed curves show the power available to the fifth robot for these fluid speeds.
}\figlabel{relative power for nonuniform robots in vessel}
\end{figure}

\fig{relative power for nonuniform robots in vessel} shows the average power for the robots and for the last robot as a function of the offset angle $\theta$. The power is relative to the same situation as used in \fig{relative power for 5 robots in vessel}.
The nonuniform spacing has little effect on the average power but aligned, nonuniformly spaced robots provide $20\%$ more power to the last robot than uniformly spaced robots.

The choice of robot positions on the vessel wall alters the power available to downstream robots in a manner similar to that of robots limiting their power use, as discussed in \sect{limiting power}, but without requiring the robots to continually monitor and adjust their power consumption. Instead, robots attaching to a vessel wall for extended operation could determine their positions during their setup and then avoid devoting computation to maintaining power limits while they perform their tasks while attached to the vessel wall. On the other hand, actively adjusted power limits provide more flexibility in distributing power to downstream robots. Robots could adopt a hybrid approach of positioning themselves to best achieve their goals and having upstream robots limit their power consumption when downstream robots indicate they require more power.

As described in  \sect{capillaries}, robots moving passively with the blood and attaching to capillary walls can lead to significant variations in the number of robots in each capillary. Thus, in addition to adjusting their position in individual capillaries, robots could alter how they divide among nearby capillaries in a network of vessels. For example, robots could preferentially position themselves in some of the capillaries in a network of vessels while leaving others with few or no robots. In that case the increased vascular resistance due to robots would tend to direct blood cells away from branches of the network containing many robots, thereby contributing to the heterogeneity of paths that cells take through a network of small vessels~\cite{stauber17}.
Provided the open paths interleave closely with the vessels blocked by robots, the flow through the open paths could be sufficient to support the surrounding tissue with less disruption to the overall flow than if robots were positioned in all the capillaries.
The possibility of using just a portion of capillaries to perfuse tissue arises because resting tissue can contain more capillaries than required for adequate perfusion~\cite{feher17}, though with considerable variation among locations in the body~\cite{augustin17}.

\subsection{Managing Heat Generation}

\tbl{power} shows large numbers of robots add significant heat to the body. This applies to any method robots use to generate power~\cite{freitas99}.

In terms of the chemical power discussed here, mitigating oxygen depletion by transporting more oxygen (\sect{oxygen storage}) does not address issues arising from heating. Instead, providing more oxygen allows robots to increase their power use, and hence produce more heat.
Limiting power production can both reduce oxygen depletion (\sect{limiting power}) and reduce the heat generated by the robots. Thus limiting power is the better approach if the large number of robots leads to significant problems due to both oxygen depletion and heating. 

Another method to mitigate heating is modifying the location-based power limits discussed in \sect{limiting power} to change where the heating occurs. In particular, robots could shift heat production to regions of the body that readily dissipate the heat. Such regions might be detected by having a lower temperature than the core of the body, such as near the skin.
This approach to heat mitigation is particularly suitable for missions where high power demands occur near the skin, e.g., for robots communicating information to external receivers when they are near the skin, such as described with the Markov process example of \fig{Markov data collection}. An  alternative communication strategy of network message passing through hubs located throughout the body would distribute power use and heating throughout the body. With enough robots that heating becomes an issue, heat dissipation could constrain the choice of communication method, in addition to constraints on transmission rates, latency or reliability.

\section{Discussion}\sectlabel{discussion}

The circulation model developed here provides a quantitative assessment of chemical power available to large numbers of microscopic robots in the circulation. 
The simplifying assumptions used in this model suggest directions for extending the model.

The model uses a fixed relationship between hemoglobin binding and oxygen concentration in the plasma around the cell. However, hemoglobin binding changes in response to other chemicals. In particular, elevated \CarbonDioxide\ increases the release of oxygen by altering the cell saturation equilibrium given in \eq{equilibrium saturation}. Carbon dioxide is a byproduct of full oxidation of glucose. If fuel cells in the robots complete that oxidation, the robots will increase \CarbonDioxide. Accounting for this effect is particularly relevant for a modest number of robots: few enough that cells return to the lung with relatively high saturations but large enough to noticeably decrease oxygen concentration. Since cells release more oxygen in response to higher \CarbonDioxide, they will provide those robots with somewhat more power than estimated from the model used here.
Alternatively, robots might use a simpler fuel cell that only partially oxidizes glucose. This can produce a similar amount of power as full oxidation because power is limited by availability of oxygen rather than glucose~\cite{hogg10}, but would produce partially oxidized glucose rather than \CarbonDioxide\ as a byproduct, and thus would not have the same effect on hemoglobin binding.

The oxygen concentration used at the start of the circulation loop assumes robots do not limit red blood cell oxygenation in lung capillaries. This is reasonable for the number of robots considered here when they move passively with the blood flow. However, if those robots remain in lung capillaries long enough to fill oxygen tanks (see \sect{oxygen storage}), the high concentration of robots in lung capillaries could reduce oxygen that reaches red cells. Adjusting for this effect requires evaluating the competition between cells and nearby robots~\cite{hogg10} and accounting for the kinetics of oxygen binding to hemoglobin~\cite{clark85}. The result of this analysis would replace the initial plasma oxygen concentration and cell saturation used by the model.

The model considered here follows a robot moving through an average circulation loop. According to this model, a large number of robots consume almost all the oxygen toward the end of the circuit, i.e., as the blood returns to the heart in large veins. 
However, the actual circulation consists of paths with a wide distribution of circulation times~\cite{morris57}. Blood from these paths mixes in large veins before returning to the heart.
Merging flows from circulation paths of different lengths result in somewhat higher oxygen concentration in large veins than suggested by the average-circuit model, and the lowest concentrations will occur in long circulation loops just before they merge with blood from shorter circuits. Extending the model to include various circulation times is necessary to determine the minimal oxygen concentration in the circulation due to robot consumption and where it occurs.

Another extension to this model is to consider robots operating outside blood vessels. This requires extending robot oxygen consumption, discussed in \sectA{robot oxygen}, to also occur in the tissue around capillaries. Tissue has less available oxygen than the blood~\cite{krogh19,popel89}. Moreover, oxygen in tissue, supplied by diffusion from nearby vessels, takes longer to replenish from red blood cells than when robots consume oxygen from plasma within vessels. Thus robots operating in tissue have less available power and could lead to more significant reduction in oxygen available to tissue than robots in blood vessels. The extent of these effects depends on the number of robots that leave vessels to operate in tissue and how long they remain. In addition, variation in how rapidly oxygen diffuses from capillaries and in tissue power demands in different parts of the body are not included in the average circulation model considered here.

Robots using glucose fuel cells decrease oxygen in the body, add carbon dioxide and produce heat.
These changes could induce compensatory responses by the body similar to those from higher tissue demand and heat production during exercise, such as more rapid breathing and faster circulation speed.

How blood flow changes in response to robot oxygen use depends on how closely robot consumption mimics the signals from increased tissue demand. 
For instance, exercising increases metabolic waste products in active muscles while robot consumption does not produce these waste products.
One way to estimate this response is to experimentally reduce oxygen directly rather than as an indirect effect of increasing tissue demand~\cite{chapler86}.  
The regulation of blood flow through small vessels~\cite{popel05} could also be important in determining this response.
If such studies identify additional changes needed to induce physiological responses to decreased oxygen, microscopic robot applications could apply these signals, either directly by the robots or as an additional intervention in conjunction with the robots.

The body may respond to robot consumption by increasing blood flow to tissues. However, this would not bring additional oxygen to the same extent as the normal response to higher tissue demand. This is because the additional blood would itself contain oxygen-consuming robots. Thus, at the higher ranges of robot numbers considered here and for long-term use, robots may need to supplement the body's response, for example by extracting additional oxygen from the lungs into storage tanks. The body's response and any supplementation by robots could be included in the model through changes to the circulation parameters, particularly the blood flow rate. More generally, the robots could alter a variety of physiological properties. 
It will be important to include these effects to create more accurate models~\cite{laubenbacher21} to guide the development and use of these robots.

The fuel cells considered here use glucose as the fuel. This fuel is far more abundant than oxygen~\cite{hogg10}, leading to the focus of this paper on how robots affect the concentration of oxygen. However, full oxidation of glucose is a complicated series of reactions~\cite{chaudhuri03}. Other fuels, such as hydrogen, have simpler oxidation chemistry though are not normally available in the circulation. If such fuels were added to the blood, e.g., via the lungs~\cite{cole21} or injection into the blood, robots could use these fuels instead of glucose. The distribution of these fuels in vessels could have significantly different concentration profiles than that of glucose. In particular, fuel rather than oxygen could be the limiting chemical is some parts of the body, requiring extending the model considered here to evaluate the concentrations of both the fuel and oxygen. In addition, a robot could contain fuel cells that react different fuels, allowing it to select among fuels depending on their availability. This capability leads to an additional mitigation strategy for a swarm, whereby robots limit their power production from each fuel based on how that affects available power to other robots in light of how those fuel concentrations vary in the body.

As described in \sect{oxygen}, the model discussed in this paper focuses on steady-state oxygen concentration arising from robot consumption.
In addition to a steady-state reduction in oxygen throughout the blood, the timing of robot activities may affect the body's response.
For instance, robots could alternate between high-power active operation and low-power intervals where they passively monitor for cellular changes or wait for instructions. If many robots alternate at the same time, this could lead to large changes in oxygen concentration.
Due to delays between changes in oxygen consumption and the body's response, such variation can lead to instabilities in respiration over periods of tens of seconds~\cite{francis00}.

Instabilities arising from the timing of robot oxygen consumption contrast with those that may arise even if robots use oxygen at a steady rate.
For example, large numbers of robots in the blood may alter vascular resistance, which can lead to instability in blood pressure~\cite{keener09}.
Another instability could occur during extended missions, e.g., for health monitoring or treatment evaluation over days or weeks.
Since glucose and oxygen are continually present in the body, their use for chemical power could support such missions.
In such cases, longer-term body responses could become important.
One example is if continual, repeated collisions between robots and blood cells substantially reduce the lifetime of the cells, even though each individual collision is unlikely to significantly damage a cell~\cite{freitas99}.
Due to delays in production of new blood cells, such reduction in cell lifetime can lead to instability in blood cell production~\cite{keener09}.
The resulting oscillations in hematocrit could affect oxygen available to tissue and robots.

Thus it will be important to evaluate temporal variation in physiology, at a variety of time scales, from the use of large numbers of robots. Extending the steady-state model discussed here to include variation in flow could help estimate such effects.

\section{Conclusion}\sectlabel{conclusion}

This paper evaluates the effect of large numbers of microscopic robots using chemical power as they move with the blood through a circulation loop with average transit time.
Normally, blood absorbs oxygen in the lungs and holds that oxygen until reaching capillaries, where some of that oxygen diffuses into surrounding tissue while the rest returns to the lungs. Robots, on the other hand, could consume oxygen continually, not just during the short time while the blood passes through capillaries. This leads to a continually decreasing oxygen concentration throughout the circulation. 

This analysis indicates that tens of billions of robots can each generate hundreds of picowatts with relatively little effect on oxygen levels or additional heating. As discussed in \sect{consequences}, this many robots are unlikely to lead to detrimental consequences beyond local effects if and where robots aggregate~\cite{hogg10}.
On the other hand, a trillion robots significantly reduce oxygen, especially in the veins, and produce appreciable heating. If such a large number of robots is necessary for an application, methods to reduce these effects include providing additional oxygen in the blood or having robots limit their power consumption.  \sect{mitigation} applied the circulation model developed in this paper to evaluate these mitigation strategies.
As described in \sect{anemia}, the model of robot performance and their systemic effects can be adjusted to patient-specific parameters. More broadly, such models could help develop individualized mission plans for microscopic robots.

Chemical power is one of several methods to power microscopic robots~\cite{freitas99}. 
Acoustic power provided by transducers located on the skin is another option. In that case, the main variation in available power is due to distance from the skin and macroscopic acoustic properties, particularly attenuation, in the tissue between the robot and the skin. Swarm behaviors can reduce the robots' effect on attenuation~\cite{hogg22}. For example, the high attenuation in lung significantly reduces acoustic power available to robots in that organ. Thus acoustic and chemical power are somewhat complementary: chemical power is limited by oxygen concentration, which depends on the time since a robot last passed through the lung, whereas acoustic is limited by attenuation and distance from the skin.
These power methods also differ in their implementation: acoustic gives mechanical power whereas fuel cells give electrical power, which needs to convert to mechanical to move robot components, e.g., vibrating surfaces for communication or moving external structures for locomotion. Thus selection among these or other power sources will depend on the nature of the robots' tasks, where in the body they require the most power, and feasibility of design and fabrication of the power supply within the robots. Large numbers of robots can significantly alter the availability of alternate power sources in different ways. Thus accounting for the effects of swarm behaviors is important in choosing how to power large numbers of microscopic robots. The results of this paper indicate some of these effects for chemical power.

\section*{Acknowledgements}

I thank Robert Freitas, David Lister, Ralph Merkle, James Ryley and Jeff Semprebon for helpful comments.

\newpage
\appendix

\numberwithin{equation}{section}
\numberwithin{figure}{section}
\numberwithin{table}{section}

\begin{center}
{\huge \textbf{Appendix}}
\end{center}

This appendix describes the model of robot power in a typical circulation loop. This description consists of three parts: 1) a model of flow in a circulation loop and variation in hematocrit through that loop that treats cells and plasma as separate compartments, 2) how oxygen transfer from blood cells to plasma affects the concentration in these vessel compartments, and 3) the rates at which blood cells release oxygen in response to consumption by robots and tissue.

\section{Vessel Circuit Model} 
\sectlabel{vessel structure}

The circuit we consider starts as blood leaves a lung capillary, where its oxygen concentration is set by that from the air in the lung. After leaving the lung, oxygen concentration decreases due to consumption by robots and tissue.  This circuit continues through the body and back to the lung. The circuit ends as the blood is about to enter a lung capillary.
The time to complete a circuit depends on where the blood goes, e.g., shorter times for transit through the head than through the feet. We consider an average circulation time of about one minute, i.e., the time required for the resting heart rate, $5\,\liter/\minute$, to pump the entire blood volume, \bloodVolume.

\subsection{Hematocrit and Vessel Size}\sectlabel{hematocrit}

The concentration of glucose in blood is much larger than that of oxygen~\cite{freitas99}, so oxygen is the chemical that limits the power available to robots. Blood carries most of its oxygen bound to hemoglobin in red blood cells. 
The number of red cells in blood is commonly expressed by the fraction of the blood volume occupied by cells, i.e., the hematocrit, which is typically around $45\%$~\cite{freitas99}. This is the average value over the entire blood volume. 

In small vessels the fluid flow tends to push cells toward the center of the vessel where they move faster than the average speed of the flow. This Fahraeus effect decreases the hematocrit in small vessels. 
While there is considerable variation among such vessels, on average the hematocrit \hematocrit\ in a vessel of diameter \dVessel\ (measured in microns) is approximately~\cite{pries92}
\begin{equation}\eqlabel{hematocrit}
\frac{\hematocrit}{ \hematocritFull } =
	\hematocritFull  + (1-\hematocritFull) \left( 1 + 1.7 e^{-0.35 \dVessel} - 0.6 e^{-0.01 \dVessel} \right)  
\end{equation}
where \hematocritFull\ is the hematocrit of the entire blood volume. 
For example, when overall hematocrit is $\hematocritFull = 0.45$, \eq{hematocrit} gives $\hematocrit = 0.34$ in a capillary with diameter $d=8\,\micron$.
The circulation model uses  \eq{hematocrit} to determine hematocrit, thereby treating hematocrit as only depending on vessel diameter.

\subsection{Parts of a Circulation Loop}\sectlabel{segments}

Oxygen concentration in the plasma arises from the balance of consumption, by robots and tissue, and replenishment, from red blood cells. Replenishment depends on the number of cells in the plasma, i.e., the hematocrit. 

From \eq{hematocrit}, hematocrit only deviates significantly from \hematocritFull\ in vessels whose diameters are less than about a millimeter. Thus it is not necessary to distinguish diameters larger than this to determine hematocrit.
Instead, we only need to explicitly account for vessel diameter during the portion of the circuit through small vessels. 
This observation allows grouping the transport through large-diameter vessels and the heart, which all have the same hematocrit, to 
produce a circuit with the following parts:
\begin{enumerate}
\item a sequence of small veins of increasing diameters starting from the end of a lung capillary
\item large pulmonary veins, the heart, large arteries \label{large 1}
\item a sequence of small arteries of decreasing diameters to the start of a body capillary
\item a body capillary
\item a sequence of small veins of increasing diameters from the end of a body capillary
\item large veins, the heart, large pulmonary arteries \label{large 2}
\item a sequence of small arteries of decreasing diameters to the start of a lung capillary
\end{enumerate}

Passage through the lung capillary at the end of the circuit contributes to the total circulation time but is not explicitly included in the model. Instead, the concentration at the start of the circuit, i.e., just after passing through a lung capillary, is a boundary condition because the transit time through lung capillaries is more than sufficient to saturate red cells with oxygen~\cite{feher17}. 
Moreover, the large spacing between robots in capillaries, even at the largest number of robots considered here (\tbl{scenarios}), means that robots do not significantly alter the oxygen available in lung capillaries.

The change in oxygen concentration in each part of the circuit depends on its transit time and hematocrit. \eq{hematocrit} gives hematocrit equal to  \hematocritFull\ in the large-vessel parts of the circuit, i.e., circuit parts \ref{large 1} and \ref{large 2}. 
We set the transit time for these parts so the total circuit time equals one minute when combined with the estimates of transit times through small vessels discussed below.

For the capillary part of the circuit, we use typical capillary diameter and transit time.  \eq{hematocrit} gives the hematocrit in the body capillary based on its typical diameter. 

For the remaining parts of the circuit, consisting of small branching vessels, we estimate hematocrit from the vessel diameter, \dVessel, via \eq{hematocrit}, and transit time from vessel geometry: diameter, \dVessel, length, \lVessel, and number of such vessel segments, \NVessel. There is considerable variation in these values. For the purpose of this model, we use average values, analogous to using the average relation between vessel diameter and hematocrit in \eq{hematocrit}.

We relate these geometric parameters to transit time in the vessels of a given type (i.e., artery, capillary or vein) and diameter. In aggregate, small vessels have larger cross section than large vessels, which leads to slower speeds in those vessels~\cite{labarbera90}. Nevertheless, the short length of the small vessels more than offsets their slower flow speed, so blood spends most of the circuit time in large vessels. The aggregate cross section of vessels with diameter $d$ is $\pi  (\dVessel/2)^2 \NVessel$.
The entire blood volume passes through this aggregate cross section, when neglecting the relatively small portion of the flow through portal systems, so that
\begin{equation}\eqlabel{flow conservation}
\NVessel \;  \frac{\pi}{4}  \dVessel^2 \; \vVessel = \frac{1}{T} \bloodVolume
\end{equation}
where \vVessel\ is the average flow speed in the vessels and $T$ is average transit time for the blood volume \bloodVolume, i.e, about one minute.
This relation gives \vVessel\ in terms of \dVessel\ and \NVessel. This velocity determines the segment's transit time as $\tVessel = \lVessel / \vVessel$.

\subsection{Geometry of Branching Vessels}\sectlabel{branching geometry}

As described above, we use geometric parameters of small branching vessels to quantify those parts of the circuit. For evaluating robot power in an average circuit, it is sufficient to use representative average branch geometry.
One approach to quantifying vessel geometry uses scaling laws that relate flow, transit time, morphology (e.g., vessel diameter and length) and topology (i.e, connectivity)~\cite{razavi18}. 
A second approach uses empirical measurements of vessel geometry. 
We apply this approach using the branching of vessels in the lung for which data on complete vascular trees is available~\cite{weibel62,singhal73,huang96,townsley12}. Although the structure of capillary networks in the lung differs to some extent from that in other parts of the body~\cite{townsley12}, we take the branching of small arteries and veins in the lung as representative of such vessels for a typical circuit through the body.

Specifically, we use the measured geometry of arterial and venous trees in the lungs~\cite{huang96}, which indicates that most of the flow is through vessels of successive branch orders. Flow that skips a few branch orders corresponds to blood that reaches capillaries through fewer branchings than blood that goes through all orders. The model of average flow neglects this variability and instead focuses on the main flow through successive branching orders. This gives a sequence of vessel lengths, diameters and number of branches at each branching order for both arterial and venous trees~\cite[Tables 2 and 5]{huang96}.
Doubling the number of branches to account for flow through both lungs, \eq{flow conservation} determines transit speed, and hence transit time, for each level of branching.

\subsection{Transit Through a Circulation Loop}

Large arteries branch into successively smaller vessels until they reach capillaries, and then merge into increasingly large veins. 
For circulation through a sequence of vessels $i=1,2,\ldots$, the total length is $\sum_i \lVessel_i$, and total passage time is $\sum_i \lVessel_i/\vVessel_i$, which equals the typical total transit time $T$. 

As described above, for small vessels we use geometric parameters of branching in lungs. Subtracting the transit time through small vessels from the total time $T$ gives the time in the circuit parts corresponding to large vessels, i.e., circuit parts \ref{large 1} and \ref{large 2} in \sectA{segments}.  To partition this time between these two large-vessel parts, i.e., arteries and veins to and from a capillary, respectively, we take the time in veins to be $1.5$ times larger that in arteries, corresponding to somewhat slower flow speed in the veins.

\begin{figure}
\centering 
\includegraphics[width=\figwidth]{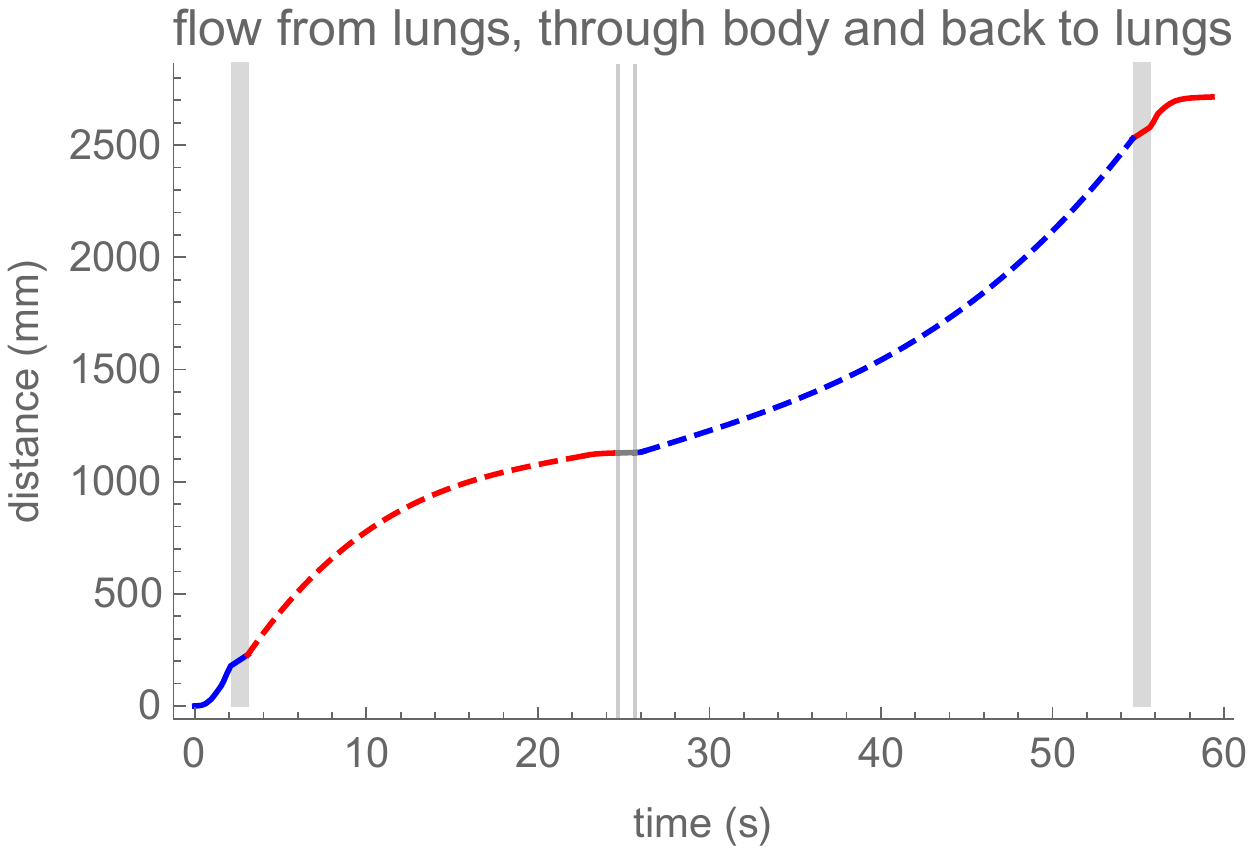}
\caption{A circulation loop starting and ending in lung capillaries.  The shaded vertical bars indicate passage through the heart, and the vertical lines near the center indicate passage through a capillary. The dashed sections are interpolations for large vessels to and from the small-vessel branches into and out of a body capillary. Red and blue portions of the curve correspond to arteries and veins, respectively.}\figlabel{circuit}
\end{figure}

\begin{figure}
\centering 
\includegraphics[width=\figwidth]{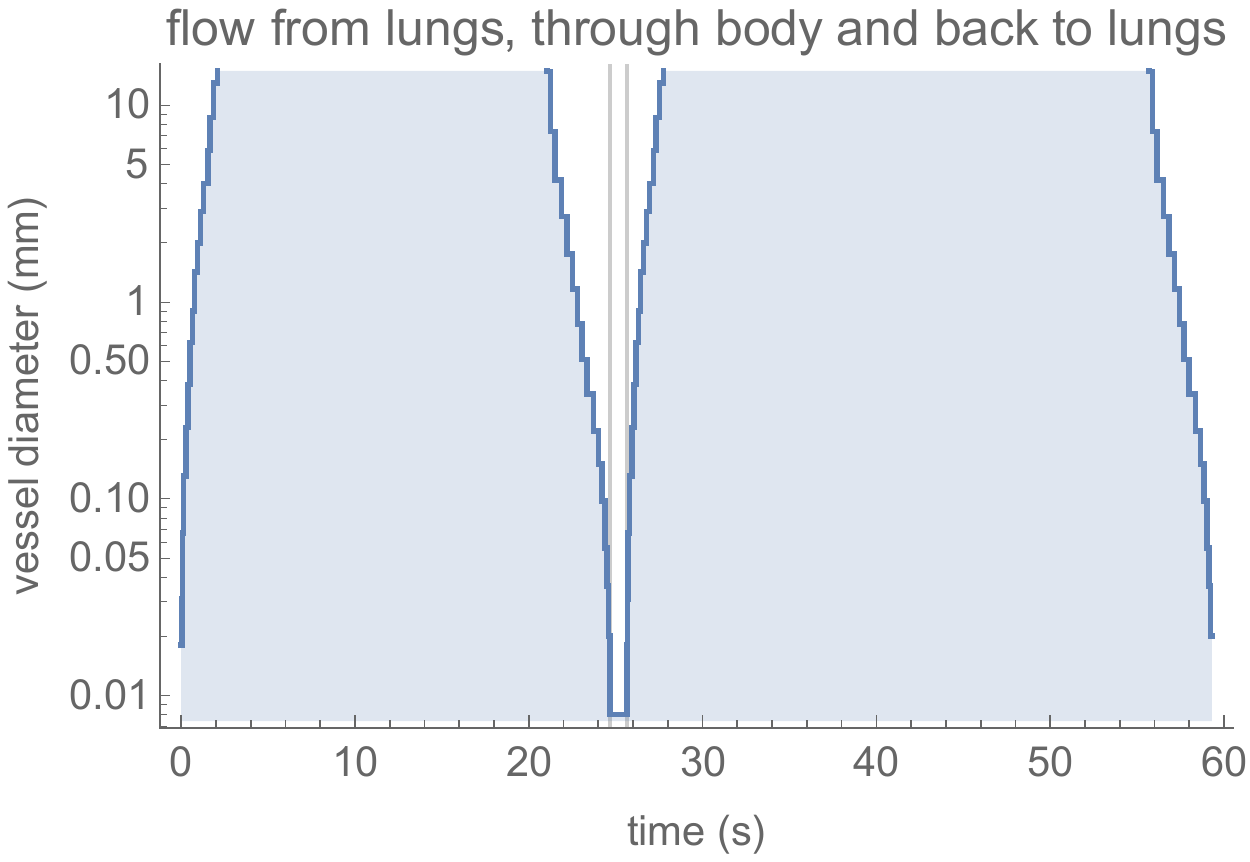}
\caption{Diameters of vessels, on a log scale, as a function of time through the vessel circuit, highlighting the diameters of small vessels, where hematocrit deviates from its overall value.}\figlabel{diameters}
\end{figure}

Combining these vessel properties, \fig{circuit} illustrates the circuit model. For the sake of illustration, the diagram shows the transit through each side of the heart as taking $1\,\second$ to transverse $50\,\millimeter$.
Since hematocrit is the same in the heart and in large vessels, the precise time spent in the heart has no effect on the model results.
The diagram schematically illustrates transit through large vessels as an interpolation (dashed curves) between the heart and the branching through small vessels to and from a capillary. This interpolation matches typical average flow speed of blood leaving and entering the heart. Specifically, flow speed in the aorta averaged over a heart beat is around $110\,\millimeter/\second$ and speed in the vena cava is around $135\,\millimeter/\second$~\cite{gabe69}.
This interpolation illustrates the flow through large vessels. However, the model of oxygen consumption only depends on the transit time through those vessels, not the specific shape of the dashed curves in the figure.

\fig{diameters} shows the vessel diameters for the vessel circuit. For small vessels, the diameters and transit times are the values described in \sectA{branching geometry}. The diameters for large vessels are not shown in the figure, and are not relevant for this model: such vessels are large enough that hematocrit equals the overall value \hematocritFull\ (see \eq{hematocrit}).

\subsection{Flow of Plasma and Cells}

\begin{figure}
\centering
\begin{tabular}{cc}
\includegraphics[width=\mfigwidth]{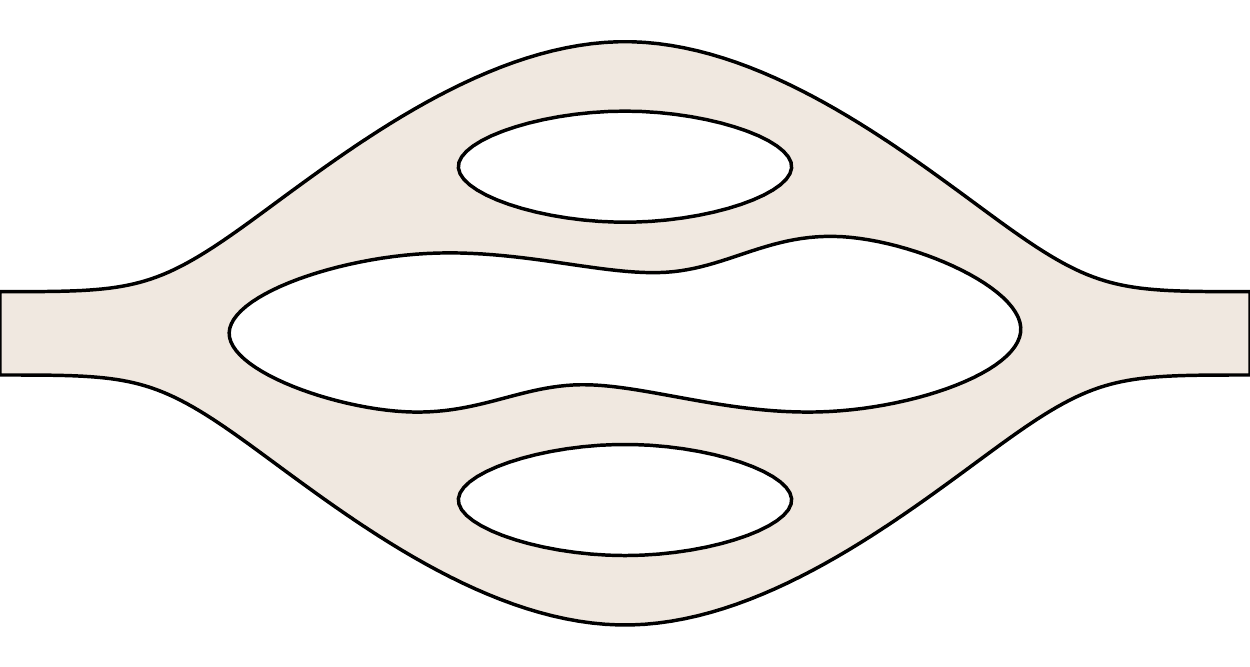}	&	\includegraphics[width=\mfigwidth]{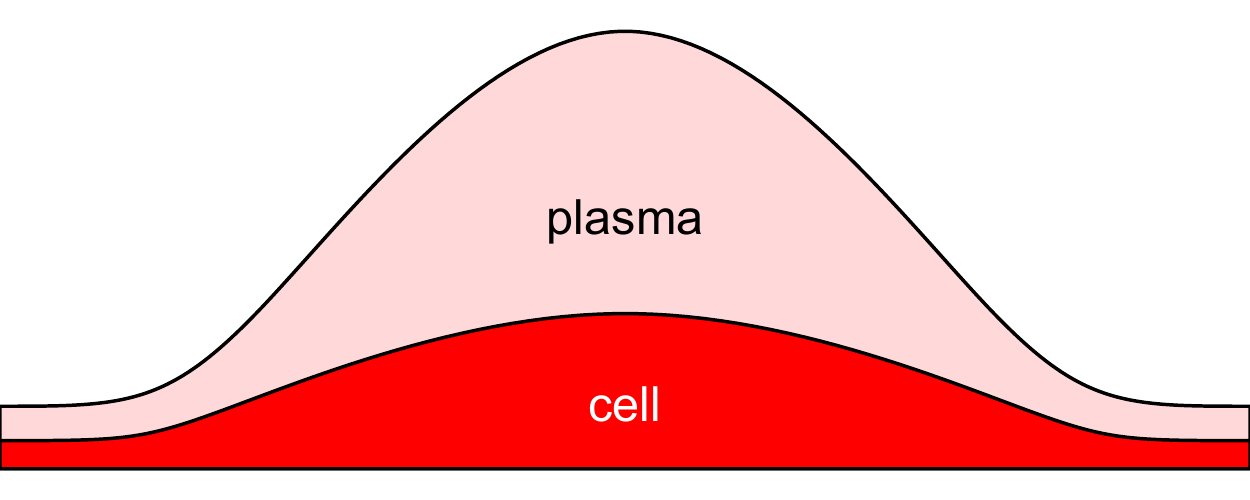} \\
(a) & (b) \\
\end{tabular}
\caption{(a) Schematic vessel branching. (b) Aggregated vessel cross section, with larger cross section corresponding to smaller vessels. The fraction of volume occupied by cells, i.e., hematocrit, is smaller in the smaller vessels.}\figlabel{branching}
\end{figure}

\fig{branching}a is a schematic of vessel branching. Smaller vessels have larger total cross section and lower hematocrit than larger vessels. The vessel properties relevant to robot oxygen consumption are the flow speed and hematocrit, not the detailed vessel branching. This allows using the simplified aggregated model of the flow illustrated in \fig{branching}b. 
In this aggregated model, the circulation consists of a single vessel with varying cross section and hematocrit.

To evaluate robot oxygen consumption over a minute or so, we consider a one-dimensional model of vessel flow and average over the variation in speed due to heart contractions. This gives flow speed $v(x)$ depending on the location $x$ along the aggregated vessel, but not on time or how close the fluid is to the vessel wall. With this time averaging, the total cross section $A(x)$ is independent of time.

Consider the flow in a vessel with cross section area $A(x)$ and flow speed $\vVessel(x)$ at position $x$. Fluid flows through the cross section at $x$ at a rate $\density \vVessel(x) A(x)$ where \density\ is the fluid density. This rate is constant throughout the vessel for incompressible fluid, as given in \eq{flow conservation}. Thus speed is inversely proportional to cross section area:
\begin{equation}\eqlabel{velocity}
\vVessel(x) = \vVessel_0 A_0/A(x)
\end{equation}
where $\vVessel_0$ and $A_0$ are the speed and cross section at an arbitrarily specified position along the circuit.

The total cross section $A(x)$ varies with position, as does hematocrit: larger cross sections correspond to the aggregation of smaller vessels, which have lower hematocrit as described in \sectA{hematocrit}. Thus the fraction of the total cross section corresponding to plasma and cells varies in the aggregate vessel, as illustrated in \fig{branching}b.
This observation leads to a model of aggregated vessels with two compartments, plasma and cells, whose relative cross sections change with the changing hematocrit in the vessels.

Evaluating how much cells replenish oxygen in blood plasma requires specifying the average speed of cells and plasma, \vCell\ and \vPlasma, respectively. Due to changing hematocrit, these speeds are not the same and vary with the changing cross section of the vessel.
One relation among these speeds is the average speed \vVessel\ in terms of hematocrit $h$:
\begin{equation}\eqlabel{average flow}
\vVessel = (1- \hematocrit) \vPlasma + \hematocrit \vCell
\end{equation}
In this expression, \vVessel\ is the average speed in the vessel, taken from the parameters of the circuit in \fig{circuit}. This relation assumes the addition of robots to the blood does not noticeably alter the speed, which is reasonable with the small fraction of the volume occupied by robots (\tbl{scenarios})~\cite{freitas99}.
This expression splits the blood volume between plasma and cells, ignoring the tiny volume occupied by robots. 

Another relation among these speeds arises from conservation of flow. 
To see this, consider a small volume of blood with hematocrit \hematocrit\ containing plasma and cells in a vessel with cross section area $A$. In a small time $\Delta t$, the volume of plasma and cells moving across that area are $(1- \hematocrit)A \vPlasma$ and $\hematocrit A \vCell$, respectively. 
Over a circulation time, e.g., about a minute, neither plasma nor cell volumes change significantly. Thus the flow rates of plasma and cells must be the same throughout the circuit, i.e., equal to some constants \alphaPlasma\ and \alphaCell. That is, $(1- \hematocrit)A \vPlasma = \alphaPlasma$ and $\hematocrit A \vCell = \alphaCell$.
These relations imply that the ratio of cell to plasma speed is independent of the total cross section:
\begin{equation}\eqlabel{alpha ratio}
\frac{\vCell}{\vPlasma} = \frac{1-\hematocrit}{\hematocrit} \frac{\alphaCell}{\alphaPlasma}
\end{equation}
The Fahraeus effect does not apply to large vessels, where cells and plasma move together with the average flow of the blood, i.e, $\vCell = \vPlasma$ and hematocrit is \hematocritFull. For this case, \eq{alpha ratio} gives $\alphaCell/ \alphaPlasma = \hematocritFull/(1-\hematocritFull)$.
Thus \eq{alpha ratio} becomes
\begin{equation}\eqlabel{vCell/vPlasma}
\frac{\vCell}{\vPlasma} = \frac{1-\hematocrit}{\hematocrit} {\Bigg /} \frac{1-\hematocritFull}{\hematocritFull}
\end{equation}
Combined with \eq{hematocrit}, this gives the velocity ratio as a function of vessel diameter.
For example, with the range of hematocrits used here for large and small vessels, \eq{vCell/vPlasma} gives a ratio around $1.7$ in small vessels, which matches reported values~\cite{barnard68}.

The vessel diameters illustrated in \fig{diameters} and \eq{hematocrit} give the variation in $\hematocrit(x)$ in the aggregated vessels. This value, combined with \eq{average flow} and \eqbare{vCell/vPlasma} gives the speeds $\vPlasma(x)$ and $\vCell(x)$ as a function of position in the circuit.

\section{Oxygen Concentration in Vessels} 
\sectlabel{concentration changes}

The aggregate vessel model shown in \fig{branching}b consists of two compartments, plasma and cells. The cross sections of these compartments vary along the length of the aggregate vessel. Robots are a small portion of the blood and not treated as a separate compartment for the discussion of this section. These compartments exchange oxygen with each other, and with robots and tissue.  This section describes how oxygen concentration changes in vessels with variable cross section: first for a single vessel, then for two such vessels treated as two compartments exchanging oxygen. These behaviors determine how concentration changes in a volume of fluid moving with the flow in one of these vessels. This discussion differs from behavior in vessels of a fixed cross section due to the changing cross section and fraction of the total cross section occupied by the two compartments.

\subsection{Concentration in a Vessel with Changing Cross Section}

A chemical in the fluid moves by convection with the fluid's motion as well as diffusion. For the scales and flow speeds considered here, diffusion is a minor contribution to changing concentration. In particular, the Peclet number characterizes the relative importance of convection and diffusion~\cite{squires05}:
\begin{equation}\eqlabel{Peclet}
\Pec = \frac{v d}{ \Doxygen}
\end{equation}
where $v$ is the flow speed, $d$ a characteristic distance and $\Doxygen$ the diffusion coefficient. For flow through a vessel of diameter $d$, \Pec\ roughly corresponds to the number of vessel diameters required for diffusion to spread the chemical across the vessel. For motion along the vessel, the distance at which $\Pec \approx 1$, i.e., $d = \Doxygen/v$, is the distance at which diffusion and convection have about the same effect on mass transport in a moving fluid. At significantly longer distances, convection is the dominant effect.

The distance over which diffusion is important is largest in the vessels with slowest flow, i.e., the capillaries.
Capillary flow speeds are around $1\,\millimeter/\second$ for which $\Doxygen/v = 2\,\micron$ (see \tbl{parameters}). This distance, comparable to the size of the robots, is considerably smaller than a typical capillary length, i.e., a millimeter, and the length of the full circulation loop (e.g., as indicated in \fig{circuit}) relevant for evaluating systemic effects of robot oxygen consumption. 
Moreover, the typical distance between neighboring robots is larger than $\Doxygen/v$ in the scenarios considered here (see \tbl{scenarios}).  Thus neighboring robots do not directly compete with each other for oxygen. This is unlike the case of closely spaced robots, such as aggregates on vessel walls, where robots significantly reduce oxygen available to their neighbors~\cite{hogg10}. 
In light of these observations, we neglect diffusive transport in modeling the change in oxygen on the scale of the circulation loop.

For convective flow, the chemical flux at position $x$ along the vessel is $J(x) = v(x) c(x)$ where $c(x)$ is the chemical's concentration. 
In a small section of vessel between $x$ and $x+\Delta x$, in time $\Delta t$, $J(x)A(x) \Delta t$ molecules enter that section of vessel, and $J(x+\Delta x)A(x+\Delta x)\Delta t$ leave it. In addition, reactions, such as release of oxygen by cells or consumption by tissue,  change the concentration at a rate $R$. Combining these contributions to concentration change gives
\begin{displaymath}
\frac{\partial c}{\partial t} = -\frac{1}{A} \frac{\partial}{\partial x}(J A) + R
\end{displaymath}
From \eq{velocity}, at position $x$, $J A = v_0 A_0 c$  so the rate of change of concentration is
\begin{equation}\eqlabel{concentration change}
\frac{\partial c}{\partial t} = -\frac{A_0}{A} v_0 \frac{\partial c}{\partial x} + R
	= -v \frac{\partial c}{\partial x} + R
\end{equation}
with $v$ given by \eq{velocity}.

As described in \sect{robot oxygen}, the time constant for robot oxygen consumption is less than a minute. Proposed applications for the robots involve operation over at least tens of minutes, corresponding to many circulations. Thus a reasonable simplification is to focus on the steady-state concentration in the vessels, so \eq{concentration change} becomes
\begin{equation}\eqlabel{concentration}
v \frac{\partial c}{\partial x} = R
\end{equation}
For transient operations, e.g., for a few seconds after robots start consuming oxygen, robots will have more oxygen than indicated by the steady-state analysis considered here.

\subsection{Concentration in Flowing Compartments that Exchange Oxygen}

As illustrated in \fig{branching}, the aggregated vessel consists of two main compartments: plasma and cells. These compartments have separate flow speeds, indicated by \eq{vCell/vPlasma}. Each compartment acts as a separate vessel in terms of flow, but they can exchange oxygen. Generalizing \eq{concentration} to account for this exchange gives the behavior of the concentrations, $c_1$ and $c_2$, in the two compartments:
\begin{equation}\eqlabel{concentration 2}
\begin{split}
v_1 \frac{\partial c_1}{\partial x} &= R_1 + \Rfrom \\
v_2 \frac{\partial c_2}{\partial x} &= R_2 - \Rto 
\end{split}
\end{equation}
where $\Rfrom$ is the change of concentration in compartment 1 due to chemicals from compartment 2, $\Rto$ is the decrease of concentration in compartment 2 from chemicals that move to compartment 1, and $R_i$ is the rate concentration changes in compartment $i$ due to production of chemical in that compartment (with a negative value for chemical consumed).

The two compartments share the total cross section of the aggregated vessel model, $A(x)$. With hematocrit $h(x)$, the cross sections of the two compartments are $A_1 = (1-h(x)) A(x)$ and $A_2 = h(x) A(x)$. These expressions assume robots occupy a negligible fraction of the vessel volume so there is no need to subtract the fraction of volume occupied by robots. Conservation of flow means $v_1(x) A_1(x)$ and $v_2(x) A_2(x)$ are independent of $x$.

$\Rfrom$ and $\Rto$ are rates of concentration change in the two compartments from oxygen moving from compartment 2 to compartment 1. These rates must account for different volumes, i.e., a given amount of oxygen makes a larger change to concentration in a smaller volume. The number of molecules in volume element $i$, extending from $x$ to $x+\Delta x$, is $A_i(x) \Delta x c_i$. Thus the rate molecules move from compartment 2 to compartment 1 is both $\Rto A_2 \Delta x$ and $\Rfrom A_1 \Delta x$. These must be the same, so
\begin{equation}
\Rto = \frac{1-h}{h}\Rfrom
\end{equation}
For example, when $h$ is small, the transfer of a given amount of oxygen has a much larger effect on the concentration in compartment 2 than it does on that of compartment 1.

\subsection{Concentration in a Volume Moving with the Fluid}\sectlabel{concentration moving}

Consider a robot moving in a small volume of fluid in compartment 2 of the two-compartment vessel model. \eq{concentration 2} describes the steady-state concentration in the two compartments. These relations determine how the concentration in that fluid volume changes as it moves through a circuit, such as illustrated in \fig{circuit}. 
In time $\Delta t$, the fluid volume moves from position $x$ to $x + v_2 \Delta t$. Thus the time rate of change of concentration in the volume element of compartment $i$, due to motion with the speed $v_2$ is $dc_i/dt = v_2 \partial c_i/\partial x$. Thus \eq{concentration 2} gives
\begin{equation}\eqlabel{concentration moving}
\begin{split}
\frac{d c_1}{d t} &= \left( R_1 + \Rfrom \right) \frac{v_2}{v_1}\\
\frac{d c_2}{d t} &= R_2 - \Rto 
\end{split}
\end{equation}
for the rate that concentration changes in the two compartments from the viewpoint of a volume moving with the flow in compartment 2.

The circulation shown in \fig{circuit} starts with blood leaving lung capillaries. Thus the initial condition for \eq{concentration moving} is the concentration in the lung.

\section{Changing Oxygen Concentration} 
\sectlabel{oxygen changes}

For this study, we assume robots move with blood cells rather than plasma. 
Specifically, this assumes the flow pushes robots away from vessel walls in a manner similar to the behavior of blood cells~\cite{whitmore67}. This is reasonable in capillaries where cells move through single-file: robots are too large to fit in the gap between cells and the vessel wall, so robots move between successive blood cells, and hence at similar speed. In somewhat larger vessels, if robots are pushed closer to the vessel wall than the cells, robots would move somewhat more slowly. The difference in speeds between cells and the average flow rate is largest when hematocrit differs most from its overall value, i.e., in small vessels. Even then, the difference is relatively minor due to the short time (a few seconds out of the one minute circulation) robots spend in those vessels. Thus the model results are not very sensitive to the accuracy of this assumption.
With this assumption, robots travel with the speed of compartment 2 discussed in \sect{concentration moving}. 

Completing the model of concentration change requires specifying the reaction rates appearing in \eq{concentration moving}.
During the circulation, robots and tissue consume oxygen from the plasma, and red cells replenish it. 
\fig{oxygen flows} illustrates the processes that change oxygen concentration.
In terms of \eq{concentration moving}, $R_1 = \Rrobot + \Rtissue$ is the rate, per unit volume, that robots and tissue remove oxygen from the plasma. There is no consumption within cells so $R_2=0$. The rates \Rfrom\ and \Rto\ are the transfer rates from cells to plasma that maintains the equilibrium between red cells and plasma given by \eq{equilibrium saturation}.
The remainder of this section quantifies these processes.

\subsection{Blood Cells}

Oxygen removed from blood plasma is partially replaced by oxygen released from nearby red blood cells. The time scale for this process is less than $100\,\millisecond$~\cite{clark85}, which is much shorter than the one minute circulation time considered here, and even the one second capillary transit time during which tissue extracts oxygen from the blood. Thus a reasonable approximation is that oxygen bound inside red cells is in equilibrium with the concentration in the surrounding plasma.

Oxygen bound in red cells is characterized by the hemoglobin saturation $S$: the fraction of hemoglobin capacity
in a cell which has bound oxygen~\cite{mauroy07}. The oxygen
concentration in the cell is $\CoxygenMax S$, where $\CoxygenMax$ is
the concentration in the cell when all the hemoglobin has bound
oxygen. 

Quantitatively, the equilibrium
saturation, conventionally expressed in terms of the equivalent
partial pressure $\pressure$ of \Oxygen\ in the fluid around the
cell, is described by the Hill equation~\cite{popel89,goldman08}:
\begin{equation}\eqlabel{equilibrium saturation}
\Sequib(a) = \frac{a^\nHill}{1+a^\nHill}
\end{equation}
where $a=\pressure/\Phalf$ is the partial pressure ratio, $\Phalf$
is the partial pressure at which half the hemoglobin is bound to
oxygen and $\nHill$ characterizes the steepness of the change from
low to high saturation. The saturation ranges
from near $1$ in the lungs to around $1/3$ in 
tissues. 

Henry's Law relates the partial pressure to the oxygen
concentration in the plasma around the cell: $p = \Hoxygen \Coxygen$ with the proportionality
constant $\Hoxygen$ depending on the temperature.
Thus, $ a = (\Hoxygen/\Phalf) \Coxygen = \Coxygen/C_{\textnormal{half}}$ where $C_{\textnormal{half}}=\Phalf/\Hoxygen$, which is about $2.2\times10^{22}\,\molecule/\meter^3$, using the values from \tbl{parameters}. This is comparable to lower range of oxygen concentration in tissue.

This model does not consider deviations from \eq{equilibrium saturation}, which mainly occur at low saturations~\cite{goldman08,popel89}, and variations in its parameters with changing blood chemistry, such as pH and carbon dioxide concentration.

\subsection{Robots}\sectlabel{robot oxygen}

Consider a small volume $\Delta V$ of blood with oxygen concentration $c$ in the plasma. This volume contains $\nDensityRobot \Delta V$ robots, where \nDensityRobot\  is the robot number density in the blood. With each robot absorbing at the rate given by \eq{robot absorption}, the total rate robots remove oxygen per unit volume of plasma is 
\begin{equation}\eqlabel{robot consumption}
\Rrobot = \rateRobot c
\end{equation}
with rate constant $\rateRobot =  4\pi \Doxygen \rRobot \nDensityRobot/(1-\hematocrit)$.

This absorption rate assumes each robot draws oxygen independently from a fluid with concentration $c$, that is robots are sufficiently far apart that they do not compete with their neighbors. Such competition would reduce the local oxygen concentration and hence robot absorption rate~\cite{hogg10}. This competition is insignificant for robots separated by at least about ten times their size~\cite{berg93}, which is the case for the numbers of robots considered here (see \tbl{scenarios}).

The time constant for robots removing oxygen from plasma is $\tRobot = 1/\rateRobot$.
From \tbl{parameters},  $\tRobot \approx 10\,\second$ for $10^{10}$ robots. \tRobot\ is correspondingly smaller for the scenarios with larger numbers of robots described in \tbl{scenarios}.
Thus concentration changes due to robots will reach steady-state within a single circulation time.

\subsection{Tissue}

Tissue extracts oxygen from blood passing through capillaries. In terms of the aggregated vessel model, consumption by tissue occurs over a short distance around the peak in total cross section shown in \fig{branching}b, which corresponds to capillaries. 

A simple model of tissue oxygen consumption is using a cylindrical region of tissue around each capillary~\cite{krogh19,popel89}. The radius of this tissue cylinder, \radiusTissue, corresponds to a typical maximum distance of tissue supplied from a capillary, which is a few cell diameters.

In this model, the volume of tissue receiving oxygen from a length $\Delta x$ of capillary, with radius $\radiusCapillary$ is $\pi (\radiusTissue^2-\radiusCapillary^2)\Delta x$. So the ratio of tissue to vessel volume is $(\radiusTissue/\radiusCapillary)^2-1$.

A consistency check on this model is that multiplying the total capillary volume by the tissue to capillary volume ratio should equal the total body volume.
The total capillary volume is about $4\times 10^5\,\millimeter^3$~\cite{freitas99}.
From \tbl{parameters}, the ratio of tissue to capillary volume is about 100, giving corresponding tissue volume of about $0.04\,\meter^3$, which is comparable to body volume.

Models of oxygen use and power generation in tissues can account for tissue
structure~\cite{popel89}. A simpler approach~\cite{mcguire01}, adopted in this paper,
treats the tissue surrounding the vessel as homogeneous and
metabolizing oxygen at the rate that produces power
\begin{equation}\eqlabel{tissue reaction rate}
\PowerTissue = \PowerTissueMax \frac{\Coxygen}{\ChalfReactionTissue
+ \Coxygen}
\end{equation}
where $\PowerTissueMax$ is the nominal demanded power density (i.e, power per unit volume)
of the tissue and $\ChalfReactionTissue$ is the concentration of
$\Oxygen$ giving half the maximum reaction rate.
When oxygen concentration is substantially larger than $\ChalfReactionTissue$, tissue power is nearly independent of oxygen concentration. In that case, tissue metabolic demand rather than available oxygen limit tissue power.

Dividing \PowerTissue\ by the reaction energy per oxygen molecule consumed gives the rate of oxygen consumption per unit volume of tissue. Oxidizing a single glucose molecule consumes six oxygen molecules. Thus the energy per oxygen molecule is $\reactionEnergy/6$, with \reactionEnergy\ the energy from oxidizing one glucose molecule, given in \tbl{parameters}.

For the resting tissue power demand considered here, oxygen concentration in tissue is nearly constant as a function of distance from the capillary~\cite{krogh19}, even with additional consumption from robots~\cite{hogg10}. Thus, for evaluating tissue oxygen consumption from capillaries, we consider the oxygen concentration in the tissue to be the same as that of the plasma in the capillary. That is, in \eq{tissue reaction rate} we set \Coxygen\ equal to the oxygen concentration in the capillary surrounded by the tissue.
In this case, \PowerTissue\ is constant within the tissue cylinder and total tissue consumption is \PowerTissue\ multiplied by the tissue volume around the capillary. Tissue takes oxygen from plasma in the capillary, so the rate it reduces concentration in the plasma is enhanced by the ratio of tissue to capillary volume given above, and the fraction of the vessel volume that is plasma, i.e., $1-\hematocrit$. Combining these factors, the rate tissue reduces oxygen concentration in the plasma is
\begin{equation}
\Rtissue = \frac{\PowerTissue}{\reactionEnergy/6} \left(  \left( \frac{\radiusTissue}{\radiusCapillary} \right)^2 - 1 \right) \frac{1}{1-\hematocrit}
\end{equation}
in the capillaries, and zero elsewhere.

\section{Model Parameters}\sectlabel{parameters}

\begin{table}
\centering
\begin{tabular}{lcc}
\hline \multicolumn{3}{c}{\bf geometry} \\
capillary radius    &   $\radiusCapillary$ & $4\,\micron$   \\
tissue cylinder radius   & $\radiusTissue$ & $40\,\micron$ \\ 
robot radius	& \rRobot	& $1\,\micron$\\
%
\hline \multicolumn{3}{c}{\bf fluid} \\
density		& $\density$	&$10^3 \,\kilogram/\meter^3$	\\
hematocrit    &   $\hematocritFull$ & $45\%$   \\ 
blood volume	& \bloodVolume 	& $5.4\,\liter$\\ 
\hline \multicolumn{3}{c}{\bf tissue} \\
power demand    & $\PowerTissueMax$ & $4 \,\kilowatt/\meter^3$ \\
\Oxygen\ concentration for half power   & $\ChalfReactionTissue $ & $10^{21}\, \molecule/\meter^3 $ \\
\hline \multicolumn{3}{c}{\bf power} \\
reaction energy from one glucose molecule  & $\reactionEnergy$ &
$4\times10^{-18}\,\joule$\\
robot fuel cell efficiency & $\fuelCellEfficiency$ &
$50\%$\\
%
\hline \multicolumn{3}{c}{\bf red blood cells} \\ 
partial pressure for 50\% $\Oxygen$ saturation   & $\Phalf$ &
$3500\,\pascal$ \\ 
$\Oxygen$ saturation exponent    & $\nHill$ & $2.7$ \\
maximum $\Oxygen$ concentration in cell  & $\CoxygenMax$ & $10^{25}\,\molecule/\meter^3 $ \\ 
\hline \multicolumn{3}{c}{\bf oxygen} \\ 
$\Oxygen$ diffusion coefficient & $\Doxygen$ & $2 \times 10^{-9}\,\meter^2/\second$ \\
$\Oxygen$ concentration in lung  &  $\CoxygenLung$ & $7 \times 10^{22}\,\molecule/\meter^3$ \\ 
\Oxygen\ partial pressure to concentration ratio & $\Hoxygen$ & $1.6\times10^{-19}\,\pascal/(\molecule/\meter^3)$ \\
\end{tabular}
\caption{Parameters for fluids, oxygen and microscopic robots considered here. Sources for these values are described in the text.
}\tbllabel{parameters}
\end{table}

\tbl{parameters} gives the parameter values used to evaluate robot power in the circulation. 

For tissue, $\ChalfReactionTissue$ is from Ref.~\cite{mcguire01}, and the blood cell parameters are from Refs.~\cite{clark85} and~\cite{popel89}.
The oxygen concentration in the lung corresponds to arterial concentration~\cite{freitas99}. Concentrations of glucose 
in blood plasma are in the millimolar range (about $10^{24}\,\molecule/\meter^3$), far larger than the oxygen concentrations~\cite{freitas99}.

For evaluating the rate oxygen diffuses to the robot surface (e.g., in \eq{robot absorption}), it is convenient to express oxygen concentration in terms of molecules per unit volume, as shown in \fig{oxygen}. 
By contrast, macroscopic studies usually express concentrations in more readily measurable quantities.
These include moles of chemical per liter of fluid (i.e.,
molar, M) and grams of chemical per cubic centimeter. Discussions of
gases dissolved in blood often specify concentration indirectly via
the corresponding partial pressure of the gas under standard
conditions. 
As an example of these units, oxygen concentration $\Coxygen = 10^{22}
\,\molecule/\meter^3$ corresponds to a $17 \,\mbox{$\mu$M}$
solution, $0.53\,\mbox{$\mu$g}/\mbox{cm}^3$ and to a partial
pressure of $1600\,\pascal$ or $12\,\mbox{mmHg}$. 
This concentration corresponds to
$0.037\mbox{cm}^3\Oxygen/100\mbox{cm}^3\,\mbox{tissue}$ with
oxygen volume measured at standard temperature and pressure.

Tissue power demands vary considerably, depending on the tissue type
and overall activity level. We set  \PowerTissueMax\ to a typical resting power demand~\cite{freitas99}. 
For comparison, peak metabolic rate in human tissue can be as high as $200\,\kilowatt/\meter^3$~\cite{mcguire01}.

Conventional fuel cell efficiencies are around $50\%$~\cite{freitas99}. While the efficiency of the fuel cells required for micron-size robots remains to be determined, for definiteness we use fuel cell efficiency $\fuelCellEfficiency = 50\%$.


\end{document}